\documentclass[lettersize,journal]{IEEEtran}
\usepackage{bbding}
\usepackage{pifont}
\usepackage{amsmath,amsfonts}
\usepackage{algorithm}
\usepackage{array}
\usepackage[caption=false,font=normalsize,labelfont=sf,textfont=sf]{subfig}
\usepackage{textcomp}
\usepackage{stfloats}
\usepackage{url}
\usepackage{verbatim}
\usepackage{graphicx}
\usepackage{cite}
\usepackage{booktabs}
\usepackage{balance}
\usepackage{amssymb}
\usepackage{bm}
\usepackage{dsfont}
\usepackage{multirow}
\usepackage{setspace}
\usepackage{algpseudocode}
\usepackage{makecell}
\usepackage{indentfirst}
\usepackage{arydshln}
\usepackage{color}
\usepackage[edges]{forest}
\definecolor{hiddendraw}{RGB}{205, 44, 36}
\usepackage{pifont}
\usepackage{amsthm}
\usepackage{booktabs}
\usepackage{siunitx}
\usepackage{multirow}
\usepackage{graphicx} 
\usepackage{environ}
\usepackage{tikz}
\usepackage{ragged2e}
\usepackage{url}
\renewcommand{\raggedright}{\leftskip=0pt \rightskip=0pt plus 0cm}
\begin{document}

\title{A Survey of Visual Transformers}

\author{Yang Liu, Yao Zhang, Yixin Wang, Feng Hou, Jin Yuan, \\ Jiang Tian, Yang Zhang, Zhongchao Shi, Jianping Fan, Zhiqiang He

\thanks{This work was done at AI Lab, Lenovo Research. (\textit{Corresponding authors: Z. He; Z. Shi; Y. Zhang.})}
\thanks{Y. Liu, Y. Zhang, Y. Wang, F. Hou, and Z. He are with Institute of Computing Technology, Chinese Academy of Sciences, Beijing, 100000, China and also with University of Chinese Academy of Sciences, Beijing, 100000, China (e-mail: liuyang20c@mails.ucas.ac.cn).}
\thanks{J. Yuan is with Southeast University, Nanjing, 214135, China.}
\thanks{J. Tian, Y. Zhang, Z. Shi, and J. Fan are with AI Lab, Lenovo Research, Beijing, 100000, China. Y. Zhang and Z. He are also with Lenovo Ltd., Beijing, 100000, China (e-mail: $\{$hezq; shizc2; zhangyang20$\}$@lenovo.com).}
}

\markboth{This paper has been accepted by  IEEE Transactions on Neural Networks and Learning Systems for publication}%
{Liu \MakeLowercase{\textit{et al.}}: A Survey of Visual Transformers}
\maketitle

\begin{abstract}
Transformer, an attention-based encoder-decoder model, has already revolutionized the field of natural language processing (NLP). Inspired by such significant achievements, some pioneering works have recently been done on employing Transformer-liked architectures in the computer vision (CV) field, which have demonstrated their effectiveness on three fundamental CV tasks (classification, detection, and segmentation) as well as multiple sensory data stream (images, point clouds, and vision-language data). Because of their competitive modeling capabilities, the visual Transformers have achieved impressive performance improvements over multiple benchmarks as compared with modern Convolution Neural Networks (CNNs). In this survey, we have reviewed over one hundred of different visual Transformers comprehensively according to three fundamental CV tasks and different data stream types, where a taxonomy is proposed to organize the representative methods according to their motivations, structures, and application scenarios. Because of their differences on training settings and dedicated vision tasks, we have also evaluated and compared all these existing visual Transformers under different configurations. Furthermore, we have revealed a series of essential but unexploited aspects that may empower such visual Transformers to stand out from numerous architectures, e.g., slack high-level semantic embeddings to bridge the gap between the visual Transformers and the sequential ones. Finally, two promising research directions are suggested for future investment. We will continue to update the latest articles and their released source codes at \textit{\url{https://github.com/liuyang-ict/awesome-visual-transformers}}.

\end{abstract}


\section{Introduction}\label{sec:01}
\begin{figure*}[!htbp]
    \centering
    \includegraphics[width=6.5in]{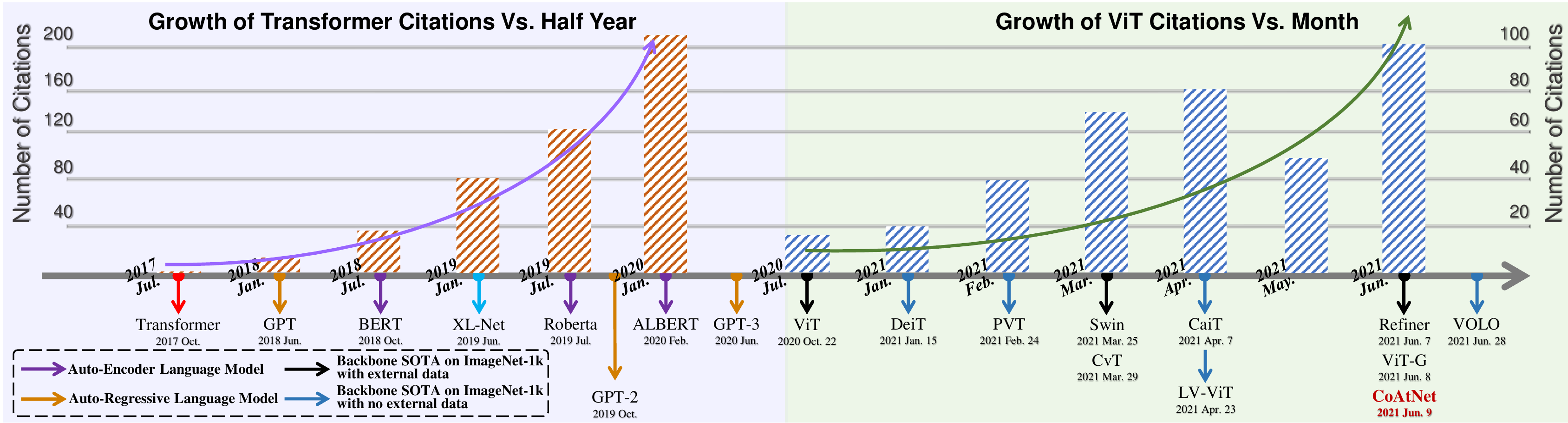}
    \caption{Odyssey of Transformer application \& Growth of both Transformer~\cite{2017Attention} and ViT~\cite{dosovitskiy2021an} citations according to Google Scholar. (Upper Left) Growth of Transformer citations in multiple conference publication including: NIPS, ACL, ICML, IJCAI, ICLR, and ICASSP. (Upper Right) Growth of ViT citations in Arxiv publications. (Bottom Left) Odyssey of language model~\cite{2017Attention,devlin2018bert,radford2018improving,yang2019xlnet,radford2019language,brown2020language,liu2019roberta,lan2020albert}. (Bottom Right) Odyssey of visual Transformer backbone where the black~\cite{dosovitskiy2021an,liu2021swin,wu2021cvt,zhou2021refiner,zhai2021scaling,dai2021coatnet} is the SoTA with external data and the blue~\cite{touvron2021training,wang2021pyramid,touvron2021going,jiang2021token,yuan2021volo} refers to the SoTA without external data (best viewed in color).}
    \vspace{-0.2cm}
    \label{fig:01}
\end{figure*}

\IEEEPARstart{T}{ransformer}~\cite{2017Attention}, which adopts an attention-based structure, has first demonstrated its tremendous effects on the tasks of sequence modeling and machine translation. As illustrated in Fig.~\ref{fig:01}, Transformers have gradually emerged as the predominant deep learning models for many NLP tasks. The most recent dominant models are the self-supervised Transformers, which are pre-trained over sufficient datasets and then fine-tuned over a small sample set for a given downstream task~\cite{radford2018improving,radford2019language,brown2020language,devlin2018bert,liu2019roberta,lan2020albert,yang2019xlnet,otter2020survey}. The Generative Pre-trained Transformer (GPT) families~\cite{radford2018improving,radford2019language,brown2020language} leverage the Transformer decoders to enable auto-regressive language modeling, while the Bidirectional Encoder Representations from Transformers (BERT)~\cite{devlin2018bert} and its variants~\cite{liu2019roberta,lan2020albert} serve as auto-encoder language models built on the Transformer encoders.  

In the CV field, prior to the visual Transformers, Convolution Neural Networks (CNNs) have emerged as a dominant paradigm~\cite{krizhevsky2012imagenet,he2016deep,tan2019efficientnet}. Inspired by the great success of such self-attention mechanisms for the NLP tasks~\cite{2017Attention,galassi2020attention}, some CNN-based models attempted to capture the long-range dependencies through adding a self-attention layer at either spatial level~\cite{wang2018non,huang2019ccnet,cao2019gcnet} or channel level~\cite{hu2018squeeze,woo2018cbam,qilong2020eca}, while others try to replace the traditional convolutions entirely with the global~\cite{parmar2018image} or local self-attention blocks~\cite{hu2018relation,hu2019local,bello2019attention,ramachandran2019stand,zhao2020exploring,zheng2020global,vaswani2021scaling}. Although Ramachandr et al. have demonstrated the efficiency of self-attention block~\cite{ramachandran2019stand} without the help from CNNs, such pure attention model is still inferior to the State-of-The-Art (SoTA) CNN models on the prevailing benchmarks.

With the grateful achievements of linguistic Transformers and the rapid development of visual attention-based models, numerous recent works have migrated the Transformers to the CV tasks, and some comparable results have been achieved. Cordonnier et al.~\cite{cordonnier2020on} theoretically demonstrated the equivalence between multi-head self-attention and CNNs, and they designed a pure Transformer by using patch downsampling and quadratic position encoding to verify their theoretical conclusion. Dosovitskiy et al.~\cite{dosovitskiy2021an} further extended such a pure Transformer for large-scale pre-training, which has achieved SoTA performance over many benchmarks. Additionally, the visual Transformers have also obtained great performances for other CV tasks, such as detection~\cite{carion2020end}, segmentation~\cite{wang2021max,cheng2021per}, tracking~\cite{chen2021transformer}, generation~\cite{jiang2021transgan}, and enhancement~\cite{chen2021pre}. 

As illustrated in Fig.~\ref{fig:01}, following the pioneer works~\cite{dosovitskiy2021an,carion2020end}, hundreds of Transformer-based models have been proposed for various vision applications within the last year. Thus, a systematic literature survey is strongly desired to identify, categorize, and evaluate these existing visual Transformers. Considering that the readers may come from different areas, we review all these visual Transformers according to three fundamental CV tasks (i.e., classification, detection, and segmentation) and different types of data streams (i.e., images, point clouds, multi-stream data). As illustrated in Fig.~\ref{fig:03}, this survey categorizes all these existing methods into multiple groups according to their dedicated vision tasks, data stream types, motivations, and structural characteristics.

Before us, several reviews on the Transformers have been published, where Tay et al.~\cite{tay2020efficient} reviewed the efficiency of the linguistic Transformers, Khan et al.~\cite{khan2021transformers} and Han et al.~\cite{han2022survey} summarized the early visual Transformers and attention-based models. The most recent review of the  Transformers is introduced by Lin et al., which provides a systematic review of various Transformers, but they only mention vision applications sketchily~\cite{lin2021survey}. Distinctively, this paper aims to provide more comprehensive review of the most-recently visual Transformers and categorize them systematically: 
\begin{itemize}
  
  \item[(1)] \textit{Comprehensiveness \& Readability.} This paper comprehensively reviews over a hundred visual Transformers according to their applications on three fundamental CV tasks (i.e., classification, detection, and segmentation) and different types of data streams (i.e., image, point clouds, multi-stream data). We select more representative methods with detailed descriptions and analyses, but introduce other related works briefly. In addition to analysing each model independently, we also build their internal connections from certain senses such as progressive, contrastive, and multi-view analysis.
  
  \item[(2)] \textit{Intuitive Comparison.} As these existing visual Transformers follow different training schemes and hyper-parameter settings for various vision tasks, this survey presents multiple lateral comparisons over different datasets and restrictions. More importantly, we summarize a series of promising components designed for each task, including: (a) \textit{shallow local convolution with hierarchical structure for backbone}; (b) \textit{spatial prior acceleration with sparse attention for neck detector}; and (c) \textit{general-purpose mask prediction scheme for segmentation}.
  
  \item[(3)] \textit{In-depth Analysis.} We further provide well-thought insights from the following aspects: (a) How visual Transformers bridge the traditional sequential tasks to the visual ones (Why does Transformer work effectively in CV); (b) the correspondence between the visual Transformers and other neural networks; (c) the double edges of the visual Transformers; and (d) the correlation of the learnable embeddings (i.e., class token, object query, mask embedding) adopted in different tasks and data stream types. Finally, we outline some future research directions. For example, the encoder-decoder Transformer backbone can unify multiple visual tasks and data stream types through query embeddings.
\end{itemize}

\begin{figure}[!htbp]
    \centering
    \includegraphics[width=2in]{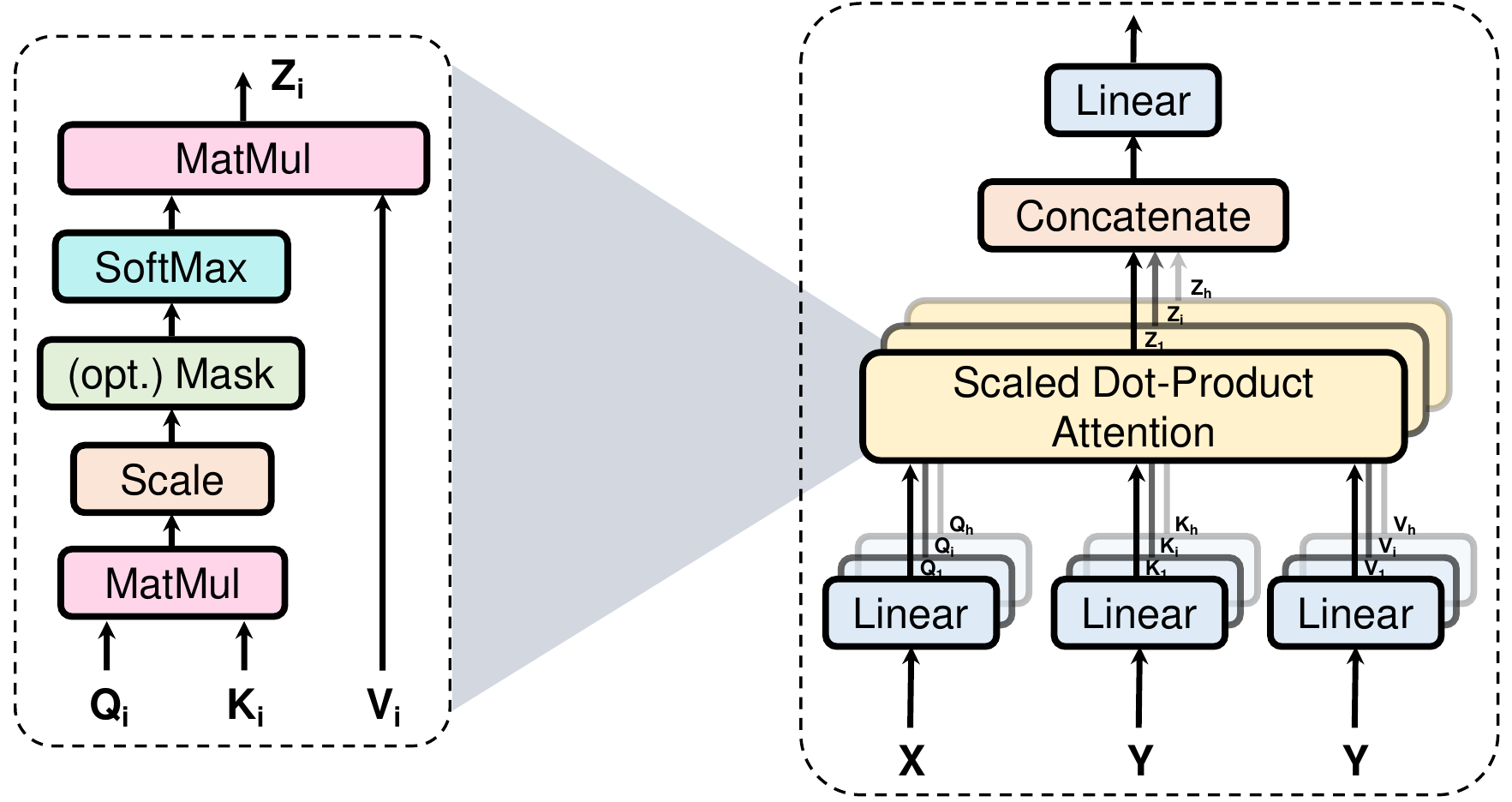}
    \caption{The structure of the attention layer. Left: Scaled Dot-Product Attention. Right: Multi-Head Attention Mechanism. }
    \vspace{-0.2cm}
    \label{fig:02}
\end{figure}

The rest of this paper is organized as follows. An overview of the architectures and the critical components for the vanilla sequential Transformers are introduced in Sec.~\ref{sec:02}. A comprehensive taxonomy for the Transformer backbones is summarized in Sec.~\ref{sec:03} with a brief discussion of their applications for image classification. We then review contemporary Transformer detectors, including Transformer necks and backbones in Sec.~\ref{sec:04}. Sec.~\ref{sec:05} clarifies the mainstream and its variants for the visual Transformers in the segmentation field according to their embedding forms (i.e., patch embedding and query embedding). Sec.~\ref{sec:03}-\ref{sec:05} also briefly analyzes a specific aspect of their corresponding fields with performance evaluation. In addition to 2D visual recognition, Sec.~\ref{sec:add01} briefly introduces the recently-developed 3D visual recognition from the perspective of point clouds. Sec.~\ref{sec:add02} further overviews the fusion approaches within the visual Transformers for multiple data stream types (e.g., multi-view, multi-modality, visual-linguistic pre-training, and visual grounding). Finally, Sec.~\ref{sec:06} provides three aspects for further discussion and points out some promising research directions for future investment. 

\section{Original Transformers}\label{sec:02}
The original Transformer~\cite{2017Attention} is first applied to the task for sequence-to-sequence auto-regression. Compared with previous sequence transduction models~\cite{sutskever2014sequence,greff2016lstm}, such original Transformer inherits the encoder-decoder structure but discards the recurrence and convolutions entirely by using \textit{multi-head attention mechanisms} and \textit{point-wise feed-forward networks}. In the following sub-sections, we will provide an architectural overview of the original Transformers and describe their four key components.

\definecolor{myorange}{RGB}{235,150,20}
\definecolor{myblue}{RGB}{0,159,230}
\definecolor{myred}{RGB}{198,50,10}
\definecolor{mypurple}{RGB}{151,50,228}
\definecolor{mypink}{RGB}{212,92,129}

\NewEnviron{dash_rec_pink}{
\par
\begin{tikzpicture}
\node[rectangle,minimum width=0.4\textwidth] (m) {\begin{minipage}{0.965\textwidth}\BODY\end{minipage}};
\draw[dashed,mypink,line width=0.7pt, rounded corners] (m.south west) rectangle (m.north east);
\end{tikzpicture}
}

\NewEnviron{dash_rec_purple}{
\par
\begin{tikzpicture}
\node[rectangle,minimum width=0.4\textwidth] (m) {\begin{minipage}{0.965\textwidth}\BODY\end{minipage}};
\draw[dashed,mypurple,line width=0.7pt, rounded corners] (m.south west) rectangle (m.north east);
\end{tikzpicture}
}

\NewEnviron{dash_rec_red}{
\par
\begin{tikzpicture}
\node[rectangle,minimum width=0.4\textwidth] (m) {\begin{minipage}{0.965\textwidth}\BODY\end{minipage}};
\draw[dashed,myorange,line width=0.7pt, rounded corners] (m.south west) rectangle (m.north east);
\end{tikzpicture}
}

\NewEnviron{dash_rec_blu}{
\par
\begin{tikzpicture}
\node[rectangle,minimum width=0.4\textwidth] (m) {\begin{minipage}{0.965\textwidth}\BODY\end{minipage}};
\draw[dashed,myblue,line width=0.7pt, rounded corners] (m.south west) rectangle (m.north east);
\end{tikzpicture}
}

\NewEnviron{dash_rec_gre}{
\par
\begin{tikzpicture}
\node[rectangle,minimum width=0.4\textwidth] (m) {\begin{minipage}{0.965\textwidth}\BODY\end{minipage}};
\draw[dashed,myred,line width=0.7pt, rounded corners] (m.south west) rectangle (m.north east);
\end{tikzpicture}
}

\NewEnviron{dash_rec}{
\par
\begin{tikzpicture}
\node[rectangle,minimum width=0.4\textwidth] (m) {\begin{minipage}{0.45\textwidth}\BODY\end{minipage}};
\draw[dashed,black,line width=0.7pt, rounded corners] (m.south west) rectangle (m.north east);
\end{tikzpicture}
}

\forestset{
myforest/.style={
delay={where content={}{shape=coordinate, for siblings={edge={myorange, line width=0pt}}}{}},
for tree={
    grow=east,
    reversed=true,
    child anchor=west,
    base=left,
    font=\tiny,
    rectangle,
    draw={hiddendraw, line width=0.7pt},
    rounded corners,align=left,
    minimum width=2.5em,
    edge={black, line width=0.55pt},
    l+=1.9mm,
    s sep=4.3pt,
    inner xsep=2pt,
    inner ysep=1.15pt,},
}
}

\begin{figure}[htbp]
\centering
\vspace{-0.2cm}
\textbf{Visual Transformers}
\vspace{0.1cm}
{\begin{dash_rec}
\vspace{0.15cm}
\quad \footnotesize{\textbf{Classification}} \small
\vspace{0.1cm}
\begin{dash_rec_red}
\begin{forest}
forked edges,
myforest,
for tree={draw={myorange, line width=0.7pt},},
where level=1{text width=6.7em,font=\tiny}{},
    [,phantom
        [Original Visual Transformer
            [SA-Net~\cite{ramachandran2019stand}{,}
             FAN~\cite{cordonnier2020on}{,}
             ViT~\cite{dosovitskiy2021an}{.} 
            ,text width=15.3em]
        ]
        [Transformer Enhanced CNN
            [VTs~\cite{wu2020visual}{,}
             BoTNet~\cite{srinivas2021bottleneck}{.} 
            ,text width=15.3em]
        ]
        [CNN Enhanced Transformer
            [Soft Inductive Bias{: }
             DeiT~\cite{touvron2021training}{,}
             ConViT~\cite{d2021convit}{.} \\
             Straightforward{: }
             CeiT~\cite{yuan2021incorporating}{,} 
             LocalViT~\cite{li2021localvit}{,} 
             CPVT~\cite{chu2021conditional}{,}
             ResT~\cite{zhang2021rest}{.} \\
             Combination{: }
             Early Conv.~\cite{xiao2021early}{,}
             CoAtNet~\cite{dai2021coatnet}{.} 
             ,text width=15.3em]
        ]
        [Transformer with Local Attn.
            [Local Only{: }
             HaloNet~\cite{vaswani2021scaling}{,}
             Swin~\cite{liu2021swin}{,} 
             VOLO~\cite{yuan2021volo}{.} \\
             Local-Global{: }
             TNT~\cite{han2021transformer}{,}
             Twins~\cite{chu2021twins}{,}
             ViL~\cite{zhang2021multi}{,} 
             Focal~\cite{yang2021focal}{.} 
             ,text width=15.3em]
        ]
        [$\,$ $\,$ Hierarchical Transformer
            [T2T~\cite{yuan2021tokens}{,}
             PVT~\cite{wang2021pyramid}{,}
             PiT~\cite{heo2021rethinking}{,}
             PVT v2~\cite{wang2021pvtv2}{,}
             CvT~\cite{wu2021cvt}{.} 
            ,text width=15.3em]
        ]
        [ $\quad$ $\,$ $\,$Deep$\quad$ Transformer
            [Structure Improvement{: }
             CaiT~\cite{touvron2021going}{,}
             DeepViT~\cite{zhou2021deepvit}{,}
             Refiner~\cite{zhou2021refiner}{.}  \\
             Loss Regulation{: }
             Diverse Patch~\cite{gong2021diverse}{.} 
             ,text width=15.3em]
        ]
        [Self-Supervised Transformer
            [Generative{: }
             iGPT~\cite{chen2020generative}{,}
             MST~\cite{li2021mst}{,}
             BEIT~\cite{bao2021beit}{,}
             MAE~\cite{he2021mae}{.}  \\
             Discriminative{: }
             MoCo v3~\cite{chen2021empirical}{,}
             DINO~\cite{caron2021emerging}{,}
             MoBY~\cite{xie2021self}{.} 
             ,text width=15.3em]
        ]
    ]
 \end{forest}
\end{dash_rec_red}
\vspace{0.1cm}
\quad \footnotesize{\textbf{Detection}} \small
\vspace{0.08cm}
\begin{dash_rec_blu}
\begin{forest}
forked edges,
myforest,
for tree={draw={myblue, line width=0.7pt},},
where level=1{text width=6.7em,font=\tiny,}{},
where level=2{text width=4.8em,font=\tiny,}{},
    [, phantom
        [$\quad$ $\,$Transformer $\quad$ Neck
            [Original Transformer
                [DETR~\cite{carion2020end}{,}
                 Pix2seq~\cite{chen2021pix2seq}{.} 
                ,text width=8.9em]
            ]
            [Sparse Attention
                [Deformable DETR~\cite{zhu2021deformable}{,}
                 ACT~\cite{zheng2020end}{,}\\
                 PnP-DETR~\cite{wang2021pnp}{,}
                 Sparse-DETR~\cite{roh2021sparse}{.} 
                 ,text width=8.9em]
            ]
            [Spatial Prior
                [One-Stage{: } 
                 SMCA~\cite{gao2021fast}{,}\\
                 \hspace{0.85cm} Conditional DETR~\cite{meng2021conditional}{,}\\
                 \hspace{0.85cm} Anchor DETR~\cite{wang2021anchor}{,}\\
                 \hspace{0.85cm} DAB-DETR~\cite{liu2022dab}{,} \\
                 \hspace{0.85cm} SAP-DETR~\cite{liu2022sap}{.} \\
                 Two-Stage{: }
                 \hspace{-0.02cm}Deformable DETR~\cite{zhu2021deformable}{,}\\
                 \hspace{0.85cm} Efficient DETR~\cite{yao2021efficient}{,}\\
                 \hspace{0.85cm} Dynamic DETR~\cite{dai2021dynamic}{.} 
                 ,text width=8.9em]
            ]
            [Structural Redesign
                [TSP~\cite{sun2021rethinking}{,}
                 YOLOS~\cite{fang2021you}{.} 
                 ,text width=8.9em]
            ]
            [Pre-trained Model
                    [UP-DETR~\cite{dai2021up}{,}
                     FP-DETR~\cite{wang2021fp}{.} 
                    ,text width=8.9em]
            ]
            [Matching Optimiz.
                    [DN-DETR~\cite{li2022dn}{,}
                     DINO~\cite{zhang2022dino}{.} 
                    ,text width=8.9em]
            ]
        ]
        [$\,$ $\,$ Transformer $\,$ Backbone
            [General{: }
             Focal~\cite{yang2021focal}{,} 
             PVT~\cite{wang2021pyramid}{,}
             ViL~\cite{zhang2021multi}{,}
             Swin~\cite{liu2021swin}{.} \\
             Specialized{: }
             FPT~\cite{zhang2020feature}{,} 
             HRFormer~\cite{yuan2021hrformer}{,}
             HRViT~\cite{gu2021hrvit}{.} 
            ,name=TB_for_DP,text width=14em]
        ]
    ]
\end{forest}
\end{dash_rec_blu}
\vspace{0.1cm}
\quad \footnotesize{\textbf{Segmentation}} \small
\vspace{0.08cm}
\begin{dash_rec_gre}
\begin{forest}
forked edges,
myforest,
for tree={draw={myred, line width=0.7pt},},
where level=1{text width=6.7em,font=\tiny,}{},
where level=2{text width=4.8em,font=\tiny,}{},
    [, phantom
        [\quad Patch-Based Transformer
            [SETR~\cite{zheng2021rethinking}{,}
             TransUNet~\cite{chen2021transunet}{,}
             SegFormer~\cite{xie2021segformer}{.} 
            ,text width=14em]
        ]
        [\quad Query-Based Transformer
            [Object Query
                [Serial{: }
                 Panoptic DETR~\cite{carion2020end}{.} \\
                 Paralleled{: }
                 Cell-DETR~\cite{prangemeier2020attention}{,}\\
                 \hspace{0.85cm}VisTR~\cite{wang2021end}{.} \\
                 Cascaded{: }
                 QueryInst~\cite{fang2021queryinst}{.} 
                 ,text width=7.6em]
            ]
            [Mask Embedding
                [Box-auxiliary{: }
                 ISTR~\cite{hu2021istr}{,}\\
                 \hspace{1.136cm}SOLQ~\cite{dong2021solq}{.}  \\
                 Box-Free{: }
                 Max-DeepLab~\cite{wang2021max}{,}\\
                 \hspace{0.83cm}Segmenter~\cite{strudel2021segmenter}{,}\\
                 \hspace{0.83cm}Maskformer~\cite{cheng2021per}{.} 
                 ,text width=7.6em]
            ]
        ]
    ]
\end{forest}
\end{dash_rec_gre}
\vspace{0.1cm}
\quad \footnotesize{\textbf{3D Visual Recognition}} \small
\vspace{0.08cm}
\begin{dash_rec_purple}
\begin{forest}
forked edges,
myforest,
for tree={draw={mypurple, line width=0.7pt},},
where level=1{text width=6.7em,font=\tiny,}{},
where level=2{text width=4.8em,font=\tiny,}{},
    [, phantom
        [Representation Learning
            [Basic: Point Transformer~\cite{zhao2021point}{,}
             PCT~\cite{guo2021pct}{,} \\
             \hspace{0.50cm}3DCTN~\cite{lu20223dctn}{,}
             Fast Point Transformer~\cite{park2021fast}{.} \\
             Fine-Grained: Pointformer~\cite{pan20213d}{,}
             SST~\cite{fan2021embracing}{,}\\
             \hspace{1.02cm}VoTr~\cite{mao2021Voxeltf}{,}
             VoxSeT~\cite{he2022Voxelst}{.} \\
             Self-Supervised: Point-BERT~\cite{yu2021point}{,}
             Point-MAE~\cite{pang2022masked}{,} \\
             \hspace{1.23cm}MaskPoint~\cite{Liu2022MaskedDF}{.} 
             ,text width=14em]
        ]
        [Cognition Mapping
            [Point-Based: 3DETR~\cite{misra2021end}{,}
             Group-Free~\cite{liu2021group}{,}
             CT3D~\cite{sheng2021improving}{.} \\
             Camera-Based: MonoDTR~\cite{huang2022mono}{,}
             MonoDETR~\cite{zhang2022mono}{,}\\
             \hspace{1.14cm}DETR3D~\cite{wang2022detr3d}{,}
             TransFusion~\cite{bai2022transfusion}{.} 
            ,text width=14em]
        ]
        [Specific Processing
            [PoinTr~\cite{yu2021pointr}{,}
             SnowflakeNet~\cite{xiang2021snow}{,}
             PointRecon~\cite{choe2021deep}{.} 
            ,text width=14em]
        ]
    ]
\end{forest}
\end{dash_rec_purple}
\vspace{0.1cm}
\quad \footnotesize{\textbf{Multi-Sensory Data Stream}} \small
\vspace{0.08cm}
\begin{dash_rec_pink}
\begin{forest}
forked edges,
myforest,
for tree={draw={mypink, line width=0.7pt},},
where level=1{text width=6.7em,font=\tiny,}{},
where level=2{text width=4.8em,font=\tiny,}{},
    [, phantom
        [Homologous Stream
            [Interactive Fusion: MVT~\cite{chen2021mvt}{,}
             MVDeTr~\cite{hou2021multiview}{,} \\
             \hspace{1.40cm}TransFuser~\cite{prakash2021multi}{,}
             COTR~\cite{jiang2021cotr}\\
             \hspace{1.40cm}Wang et al.~\cite{wang2021multi}{,}
             FUTR3D~\cite{chen2022futr3d}\\
             \hspace{1.40cm}TransformerFusion~\cite{bozic2021transformerfusion}{,}
             mmFormer~\cite{zhang2022mmformer}\\
             Transfer Fusion: Tulder er al.~\cite{tulder2021multi}{,}
             Long et al.~\cite{long2021multi}{,}\\
             \hspace{1.25cm}DRT~\cite{song2021deep}{.} 
             ,text width=14em]
        ]
        [Heterologous Stream
            [Vis.-Lin. Pre-train.: VideoBETR~\cite{sun2019videobert}{,}
             ViLBERT~\cite{lu2019vilbert}{,}\\
             \hspace{1.45cm}LXMERT ~\cite{tan2019lxmert}
             VisualBERT~\cite{li2019visualbert}{,}\\
             \hspace{1.45cm}VL-BERT~\cite{su2019vl}{,}
             UNITER~\cite{chen2020uniter}{,}\\
             \hspace{1.45cm}Oscar~\cite{li2020oscar}{,}
             Unified~\cite{zhou2020unified}{,}\\
             \hspace{1.45cm}ViLT~\cite{kim2021vilt}{,}
             VinVl~\cite{zhang2021vinvl}{,}\\
             \hspace{1.45cm}CLIP~\cite{radford2021learning}{,}
             DALL-E~\cite{ramesh2021zero}{,}\\
             \hspace{1.45cm}ALIGN~\cite{jia2021scaling}{,}
             UniT~\cite{hu2021unit}{,}\\
             \hspace{1.45cm}SimVLM~\cite{wang2021simvlm}{,}
             Data2Vec~\cite{baevski2022data2vec}{.} \\
             Visual Grounding: MDETR~\cite{kamath2021mdetr}{,}
             TransVG~\cite{deng2021transvg}{,}\\
             \hspace{1.40cm}VGTR~\cite{du2021visual}{,}
             Referring Transformer~\cite{li2021referring}{,}\\
             \hspace{1.40cm}Pseudo-Q~\cite{jiang2022pseudo}{,}
             LanguageRefer~\cite{roh2022languagerefer}{,}\\
             \hspace{1.40cm}TransRefer3D~\cite{he2021transrefer3d}{,}
             MVT(2022)~\cite{huang2022multi}{,}\\
              \hspace{1.40cm}TubeDETR~\cite{yang2022tubedetr}{.} 
            ,text width=14em]
        ]
    ]
\end{forest}
\end{dash_rec_pink}
\end{dash_rec}}
\caption{Taxonomy of Visual Transformers}
\label{fig:03}
\end{figure}
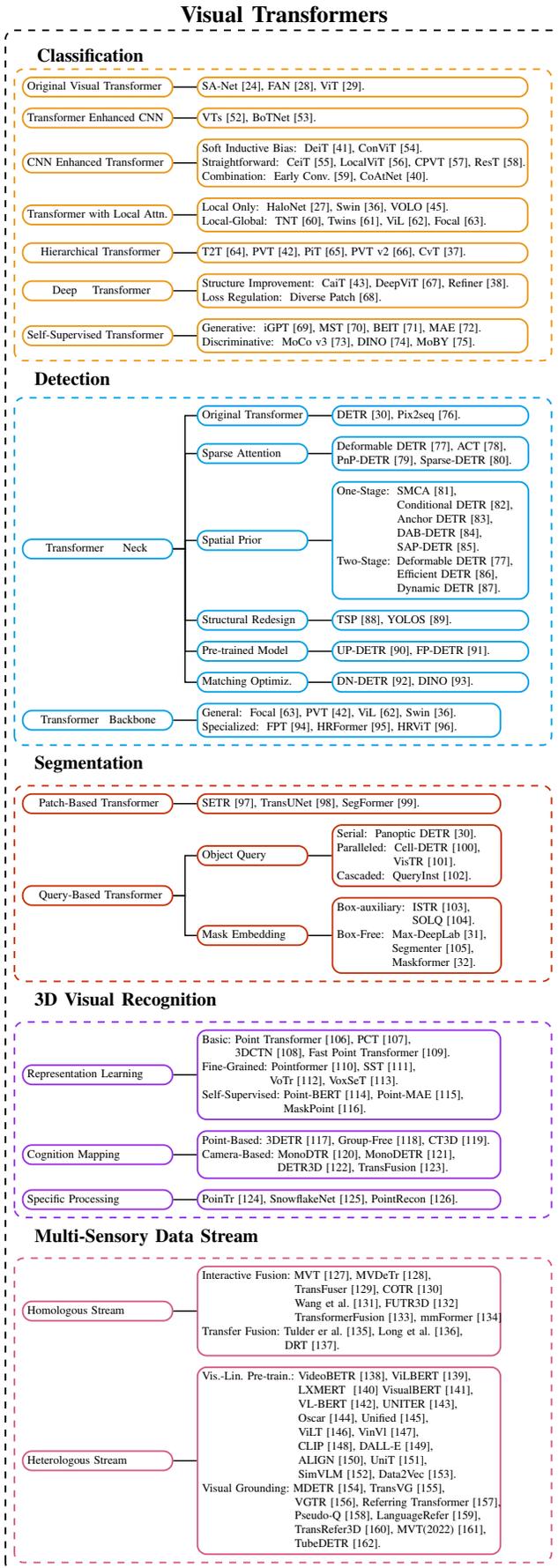


\begin{figure}[!htbp]
    \centering
    \includegraphics[width=2.3in]{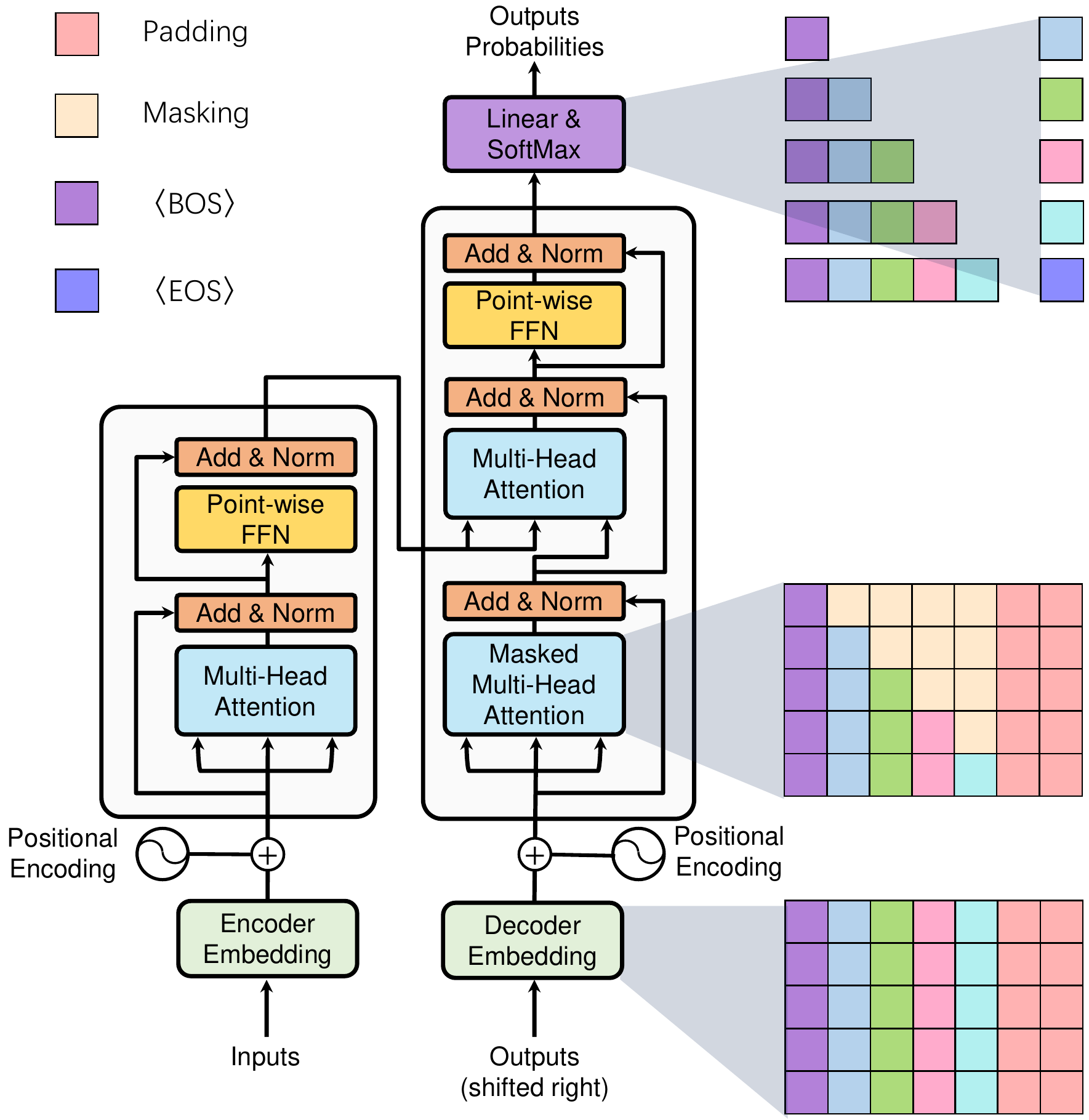}
    \caption{The overall architecture of Transformer~\cite{2017Attention}.The 2D lattice represents the each states of queries during training (best viewed in color). }
    \label{fig:04}
\end{figure}

\subsection{(Multi-Head) Attention Mechanism}\label{sec:021}
The mechanism with one single head attention can be grouped into two parts: 1) {\em A transformation layer} maps the input sequences $X\in\mathbb{R}^{n_x \times d_x}, Y\in \mathbb{R}^{n_y \times d_y}$ into three different vectors (query $Q$, key $K$, and value $V$), where $n$ and $d$ are the length and the dimension of the inputs, respectively. 2) {\em An attention layer}, as shown in Fig.~\ref{fig:02}, explicitly aggregates the query with the corresponding key, assigns them to the value and updates the output vector.

The formula for the transformation layer is defined as 
\begin{equation}\label{eq:13}
    \begin{array}{l}
    Q = XW^Q, \quad  K = YW^K, \quad  V = YW^V,
    \end{array}
\end{equation}
where $W^Q \in \mathbb{R}^{d_x \times d^k}$, $W^K \in \mathbb{R}^{d_y \times d^k}$, and $W^V \in \mathbb{R}^{d_y \times d^v}$ are linear matrices. $d^k$ and $d^v$ are the dimension of the query-key pair and the value which are projected from $Y$ and $X$, respectively. Such two sequence inputs are referred as the cross-attention mechanism. It can also be regarded as a self-attention when $Y=X$. In form, self-attention is applied to both Transformer encoder and decoder, while the cross-attention severs as a junction within the decoder. 

Then, the scale-dot attention mechanism is formulated as
\begin{equation}\label{eq:12}
    \begin{array}{l}
    \text{Attention}(Q, K, V) = \text{Softmax}\left(\dfrac{QK^T}{\sqrt{d_k}}\right)V,
    \end{array}
\end{equation}

\noindent where the attention weights are generated by a dot-product operation between $Q$ and the $K$, a scaling factor $\sqrt{d_k}$ and a softmax operation are supplied to translate the attention weights into a normalized distribution. The resulting weights are assigned to the corresponding value elements, thereby yielding the final output vector.

Due to the restricted feature subspace, the modeling capability of the single-head attention block is quite coarse. To tackle this issue, as shown in Fig.~\ref{fig:02}, a Multi-Head Self-Attention mechanism (MHSA) is proposed to linearly project the input into multiple feature sub-spaces and process them by using several independent attention heads (layers) parallelly. The resulting vectors are concatenated and mapped to the final outputs. The process of MHSA can be formulated as
\begin{equation}\label{eq:14}
    \begin{array}{l}
    Q_i = XW^{Q_i},\; K_i=XW^{K_i},\; V_i=XW^{V_i},\vspace{1ex}\\
    Z_i=\text{Attention}(Q_i,K_i,V_i),\  i=1 \dots h, \vspace{1ex}\\
    \text{MultiHead}(Q,K,V)=\text{Concat}(Z_1,Z_2,...,Z_h)W^O, 
    \end{array}
\end{equation}
where $h$ is the head number, $W^O\in\mathbb{R}^{hd_v\times d_{model}}$ denotes the output projected matrix, $Z_i$ denotes the output vector of each head, $W^{Q_i}\in\mathbb{R}^{d_{model}\times d_k}$, $W^{K_i}\in\mathbb{R}^{d_{model}\times d_k}$, and $W^{V_i}\in\mathbb{R}^{d_{model}\times d_v}$ are three different groups of matrices. Multi-head attention separates the inputs into $h$ independent attention heads with $d_{model}/h$-dimensional vectors, and integrates each head features dependently. Without extra costs, multi-head attention enriches the diversity of the feature subspaces.

\subsection{Position-wise Feed-Forward Networks}\label{sec:024}
The output of MHSA is then fed into two successive feed-forward networks (FFN) with a ReLU activation as
\begin{equation}\label{eq:15}
    \begin{array}{l}
        \text{FFN}(x)=\text{RELU}(W_1x+b_1)W_2+b_2.
    \end{array}
\end{equation}
This position-wise feed-forward layer can be viewed as a point-wise convolution, which treats each position equally but uses different parameters between each layer. 

\subsection{Positional Encoding}\label{sec:025}
Since the Transformer/Attention operates on the input embedding simultaneously and identically, the order of the sequence is neglected. To make use of the sequential information, a common solution is to append an extra positional vector to the inputs, hence term the ``positional encoding''. There are many choices for positional encoding. For example, a typical choice is cosine functions with different frequencies as
\begin{equation}\label{eq:16}
    \begin{array}{l}
        PE_{(pos,i)}=
        \begin{cases}
            \text{sin}(pos \cdot \omega_k)& \text{if} \quad i=2k\\
            \text{cos}(pos \cdot \omega_k)& \text{if} \quad i=2k+1,\\
        \end{cases}
    \vspace{1ex}\\
     \quad \omega_k = \dfrac{1}{10000^{2k/d}}, \quad k=1,\cdots,d/2,
    \end{array}
\end{equation}
where $pos$ and $d$ are the position and the length of the vector, respectively, and $i$ is the index of each element within vector.

\subsection{Transformer Model}\label{sec:026}
Fig.~\ref{fig:04} shows the overall Transformer models with the encoder-decoder architecture. Specifically, it consists of $N$ successive encoder blocks, each of which is composed of two sub-layers. 1) An MHSA layer aggregates the relationship within the encoder embeddings. 2) A position-wise FFN layer extracts feature representations. For the decoder, it also involves $N$ consecutive blocks that follow a stack of the encoders. Compared with the encoder, each decoder block appends to a multi-head cross-attention layer to aggregate both decoder embeddings and encoder outputs, where $Y$ corresponds to the former, and $X$ is the latter as shown in Eq.~(\ref{eq:13}). Moreover, all of the sub-layers in both encoder and decoder employ a residual connection~\cite{he2016deep} and a Layer Normalization~\cite{ba2016layer} to enhance the scalability of the Transformer. In order to record the sequential information, each input embedding is attached with a positional encoding at the beginning of the encoder stack and the decoder stack. Finally, a softmax operation are used for predicting the next word.

\begin{figure}[!htbp]
    \centering
    \includegraphics[width=3.1in]{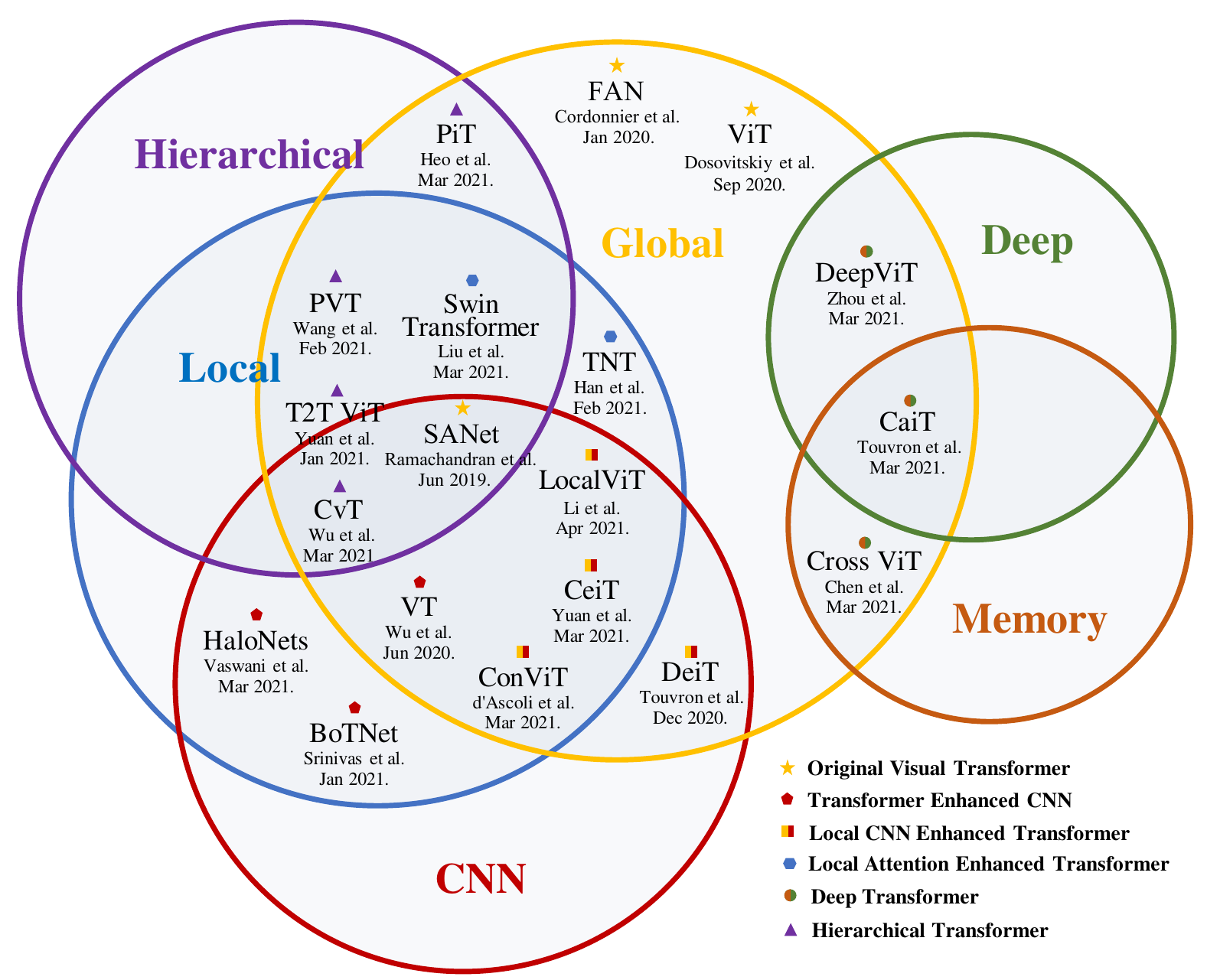}
    \caption{Taxonomy of Visual Transformer Backbone 
    (best viewed in color).}
    \label{fig:05}
\end{figure}

In an auto-regressive language model, the Transformer is originated from the machine translation tasks. Given a sequence of words, the Transformer vectorizes the input sequence into the word embeddings, adds the positional encodings, and feeds the resulting sequence of the vectors into an encoder. During training, as illustrated in Fig.~\ref{fig:04}, Vaswani et al. design a masking operation according to the rule for the auto-regressive task~\cite{2017Attention}, where the current position only depends on the outputs of the previous positions. Based on this masking, the Transformer decoder is able to process the sequence of the input labels parallelly. During the inference time, the sequence of the previously-predicted words is processed by the same operation to predict the next word.

\section{Transformer for Classification}\label{sec:03}
Following the prominent developments of the Transformers in NLP~\cite{devlin2018bert,radford2018improving,radford2019language,brown2020language}, recent works attempt to introduce visual Transformers for image classification. This section comprehensively reviews over 40 visual Transformers and groups them into six categories, as shown in Fig.~\ref{fig:05}. We start with introducing the Fully-Attentional Network~\cite{ramachandran2019stand,cordonnier2020on} and the Vision Transformer (ViT)~\cite{dosovitskiy2021an}, such \textit{Original Visual Transformer} first demonstrates its efficacy on multiple classification benchmarks. Then we discuss \textit{Transformer Enhanced CNN} methods that utilize Transformer to enhance the representation learning in CNNs. Due to the negligence of local information in the original ViT, the \textit{CNN Enhanced Transformer} employs an appropriate convolutional inductive bias to augment the visual Transformer, while the~\textit{Local Attention Enhanced Transformer} redesigns patch partition and attention blocks to improve their locality. Following the hierarchical and deep structures in CNNs~\cite{brahma2015deep}, the \textit{Hierarchical Transformer} replaces the fixed-resolution columnar structure with a pyramid stem, while the \textit{Deep Transformer} prevents the attention map from over-smooth and increases its diversity in the deep layer. Moreover, we also review the existing visual Transformers with \textit{Self-Supervised Learning}. Finally, we make a brief discussion based on intuitive comparisons for further investigation. More visual Transformers' milestones are introduced in App.~\ref{sup:development}.

\subsection{Original Visual Transformer}\label{sec:031}
Inspired by the tremendous achievements of the Transformers in the NLP field~\cite{devlin2018bert,radford2018improving,radford2019language,brown2020language}, the previous technology trends for the vision tasks~\cite{wang2018non,hu2018squeeze,chen20182,huang2019ccnet,cao2019gcnet} incorporate the attention mechanisms with the convolution models to augment the models' receptive field and global dependency. 

Beyond such hybrid models, Ramachandran et al. contemplate whether the attention can completely replace the convolution, and then present a Stand-Alone self-attention network (SANet)~\cite{ramachandran2019stand}, which has achieved superior performance on the vision tasks as compared with the original baseline. Given a ResNet~\cite{he2016deep} architecture, the authors straightforwardly replace the spatial convolution layer ($3\times3$ kernel) in each bottleneck block with a locally spatial self-attention layer and keep other structures the same as the original setting in ResNet. Moreover, lots of ablations have shown that the positional encodings and convolutional stem can further improve the network efficacy. 

Following~\cite{ramachandran2019stand}, Cordonnier et al. pioneer \textit{a prototype design} (called Fully-Attentional Network in their original paper)~\cite{cordonnier2020on}, including a fully vanilla Transformer and a quadratic positional encoding. The authors also theoretically prove that a convolutional layer can be approximated by a single MHSA layer with relative positional encoding and sufficient heads. With the ablations on CIFAR-10~\cite{krizhevsky2009learning}, they further verify that such a prototype design does learn to attend a grid-like pattern around each query pixel, as their theoretical conclusion.

\begin{figure}[!htbp]
    \centering
    \includegraphics[width=3.2in]{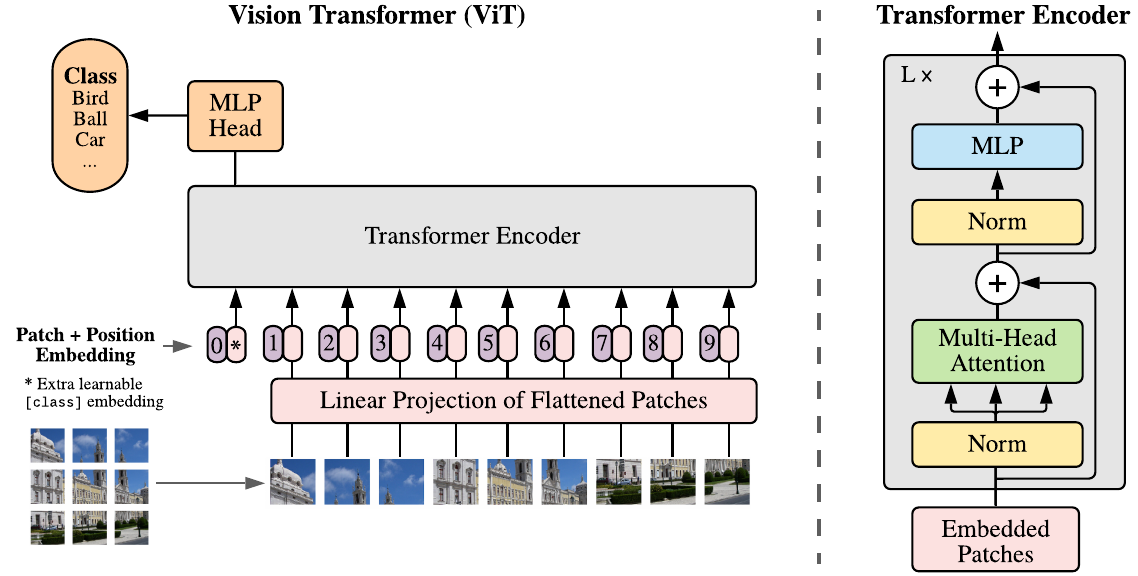}
    \caption{Illustration of ViT. The flatten image patches with an additional class token are fed into the vanilla Transformer encoder after positional encoding. Only the class token can be predicted for classification. (from~\cite{dosovitskiy2021an}.)}
    \label{fig:06}
\end{figure}

Different from~\cite{cordonnier2020on} that only focuses on lite scale model, the Vision Transformer (ViT)~\cite{dosovitskiy2021an} further explores the effectiveness of the vanilla Transformer with large-scale pre-trained learning, and such a pioneer work impacts the community significantly. Because the vanilla Transformers only accept the  sequential inputs, the input image in ViT is firstly split into a series of non-overlapped patches and they are then projected into patch embeddings. Then a 1D learnable positional encoding is added on the patch embeddings to retain the spatial information, and the joint embeddings are then fed into the encoder, as shown in Fig.~\ref{fig:06}. Similar to BERT~\cite{devlin2018bert}, a learned $[class]$ token is attached with the patch embeddings to aggregate the global representation and it serves as the final output for classification. Moreover, a 2D interpolation complements the pre-trained positional encoding to maintain the consistent order of the patches when the feeding images are in  arbitrary resolution. By pre-training with a large-scale private dataset (JFT-300M~\cite{JFT300M}), ViT has achieved similar or even superior results on multiple image recognition benchmarks (ImageNet~\cite{deng2009imagenet} and CIFAR-100~\cite{krizhevsky2009learning}) as compared with the most prevailing CNNs methods. However, its generalization capability tends to be eroded with limited training data.

\subsection{Transformer-enhanced CNNs}\label{sec:032}
As described in Section~\ref{sec:02}, the Transformer has two keys: {\em MHSA} and {\em FFN}. There exists an approximation between the convolutional layer and the  MHSA~\cite{cordonnier2020on}, and Dong et al. suggest that the Transformer can further mitigate the strong bias of MHSA with the help of skip connections and FFN~\cite{dong2021attention}. Recently, some methods attempt to integrate the Transformer into CNNs to enhance representation learning. VTs~\cite{wu2020visual} decouples semantic concepts for the input image into different channels and relates them densely through the encoder block, namely VT-block. Such VT-block substitutes the last convolution stage to enhance the CNN model's ability on semantic modelling. Unlike previous approaches that directly replace convolution with attention structure, Vaswani et al. propose a conceptual redefinition that the successive bottleneck blocks with MHSA can be formulated as the Bottleneck Transformer (BoTNet)~\cite{srinivas2021bottleneck} blocks. The relative position encoding~\cite{shaw2018self} is adopted to further mimic the original Transformer. Based on ResNet~\cite{he2016deep}, BoTNet outperforms the most CNN models with similar parameter settings on the ImageNet benchmark and further demonstrates the efficacy of hybrid models.

\subsection{CNN-enhanced Transformer}\label{sec:033}
Inductive bias is defined as a set of assumptions on data distribution and solution space, whose manifestations within convolution are the locality and the translation invariance~\cite{battaglia2018relational}. As the covariance within local neighborhoods is large and tends to be gradually stationary across an image, CNNs can process an image effectively with the help of the biases. Nevertheless, strong biases also limit the upper bound of CNNs when sufficient data are available. Recent efforts attempt to leverage an appropriate CNN bias to enhance Transformer.

Touvron et al. propose a Data-efficient image Transformer (DeiT)~\cite{touvron2021training} to moderate the ViT's dependence on large datasets. In addition to the existing strategies for data augmentation and regularization, a teacher-student distillation strategy is applied for auxiliary representation learning. The student model is the ViT, where a distilled token is attached to the patch embeddings and it is supervised by the pseudo labels from the teacher model. Extensive experiments have demonstrated that CNN is a better teacher model than the Transformer. Surprisingly, the distilled student Transformer even outperforms its teacher CNN model. These observations are explained in~\cite{abnar2020transferring}: the teacher CNN transfers its inductive bias in a soft way to the student Transformer through knowledge distillation. Based on ViT's architecture, DeiT-B attains the top-1 accuracy of 85.2\% without external data. ConViT~\cite{d2021convit} appends a parallel convolution branch with vanilla Transformer to impose inductive biases softly. The main idea of the convolution branch is a learnable embedding that is first initialized to approximate the locality as similar as the convolution and then explicitly gives each attention head freedom to escape the locality by adjusting a learned gating parameter. CeiT~\cite{yuan2021incorporating} and LocalViT~\cite{li2021localvit} extract the locality by directly adding a depth-wise convolution in FFN. As point-wise convolution is equal to position-wise FFN, they extend FFN to an inverted residual block~\cite{sandler2018mobilenetv2} to build a depth-wise convolutional framework. Based on the assumption of positional encoding~\cite{zhang2021rest} and the observation in~\cite{islam2020much}, ResT~\cite{zhang2021rest} and CPVT~\cite{chu2021conditional} try to adapt the inherent positional information of the convolution to the arbitrary size of inputs instead of interpolating the positional encoding. Including CvT~\cite{wu2021cvt}, these methods replace the linear patch projection and positional encoding with the convolution stacks. Both methods benefit from such convolutional position embedding, especially for small model. 

Besides the ``internal'' fusion, many approaches focus on an ``apparent'' combination according to different visual Transformer's structures. For standard columnar structure, Xiao et al. substitute the original patchify stem (single non-overlapped large kernel) with several stacked stride-2 $3 \times 3$ kernels~\cite{xiao2021early}. Such a Convolutional Stem significantly improves ViT by 1-2\% on accuracy for ImageNet-1k and facilitates its stability and generalization for the downstream tasks. For hierarchical structures, Dai et al.~\cite{dai2021coatnet} investigate an optimal combination of hybrid models to benefit the performance trade-off. By comparing a series of hybrid models, they propose a Convolution and Attention Network (CoAtNet) to leverage the strength of both CNNs and Transformer. The authors observe that depth-wise convolution can be naturally integrated into the attention block, and stacking convolution vertically in the shallow layer is more effective than the original hierarchical methods. It has achieved the SoTA performance across multiple datasets.

\subsection{Local Attention Enhanced Transformer}\label{sec:034}
The coarse patchify process in ViT~\cite{dosovitskiy2021an} neglects the local image information. In addition to adding CNNs, various local attention mechanisms are proposed to dynamically attend the neighbour elements and augment the local extraction ability. 

\begin{figure}[!htbp]
    \centering
    \setlength{\abovecaptionskip}{-0.1cm}
    \includegraphics[width=3.3in]{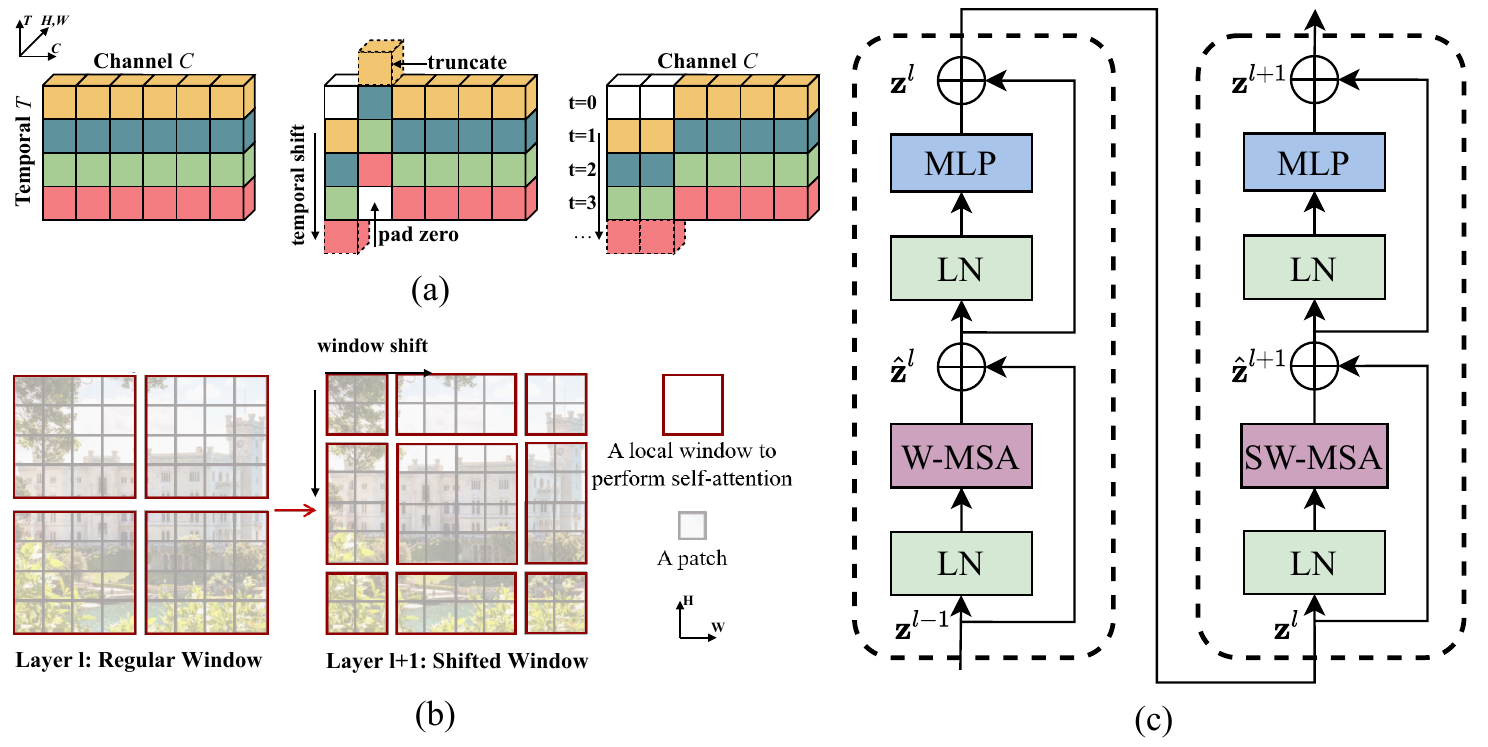}
    \caption{An overview of Swin Transformer and TSM. (a) TSM with bi-direction and uni-direction operation. (b) The shifted window method. (c) Two successive Transformer blocks of Swin Transformer. The regular and shifted window correspond to W-MSA and SW-MSA, respectively. (from~\cite{lin2019tsm,liu2021swin}).}
    \label{fig:07}
\end{figure}

One of the representative methods is the Shifted windows (Swin) Transformer~\cite{liu2021swin}. Similar to TSM~\cite{lin2019tsm} (Fig.~\ref{fig:07}(a)), Swin utilizes a shifted window along the spatial dimension to model the global and boundary features. In detail, two successive window-wise attention layers can facilitate the cross-window interactions (Fig.~\ref{fig:07}(b)-(c)), similar to the receptive field expansion in CNNs. Such operation also reduces the computational complexity from $O(2n^2C)$ to $O(4M^2nC)$ in one attention layer, where $n$ and $M$ denote the patch length and the window size, respectively. Swin Transformer achieves 84.2\% accuracy on ImageNet and the latest SoTA on multiple dense prediction benchmarks (see Sec.~\ref{sec:042}).

Inspired by~\cite{pang2017convolution}, Han et al. leverage a Transformer-iN-Transformer (TNT)~\cite{han2021transformer} model to aggregate both patch- and pixel-level representations. Each layer of TNT consists of two successive blocks, an inner block models the pixel-wise interaction within each patch, and an outer block extracts the global information. Twins~\cite{chu2021twins} employs a spatially separable self-attention mechanism, similar to depth-wise convolution~\cite{sandler2018mobilenetv2} or window-wise TNT~\cite{han2021transformer}, to model the local-global representation. Another separate form is ViL~\cite{zhang2021multi}, which replaces the single class token with a series of local embeddings (termed as global memory). These local embeddings only perform an inner attention and an interaction with their corresponding 2D spatial neighbors. VOLO~\cite{yuan2021volo} proposes an outlook attention, which is similar to a patch-wise dynamic convolution, to focus on the finer-level features, including three operations: unfold, linear-wights attention, and refold. Based on~\cite{jiang2021token}, it achieves SoTA results on ImageNet without external data.

\subsection{Hierarchical Transformer}\label{sec:035}
Due to the columnar structure of ViT~\cite{dosovitskiy2021an} with a fixed resolution across the entire Transformer layers, it loses the fine-grained features and suffers from heavy computational costs. Followed by the hierarchical models, Tokens-to-Token ViT (T2T-ViT)~\cite{yuan2021tokens} first introduces a paradigm of hierarchical Transformer and employs an overlapping unfold operation for down-sampling. However, such operation brings heavy memory and computation costs. Therefore, Pyramid Vision Transformer (PVT)~\cite{wang2021pyramid} leverages a non-overlapping patch partition to reduce feature size. Furthermore a spatial-reduction attention (SRA) layer is applied in PVT to further reduce the computational cost by learning low-resolution key-value pairs. Empirically, PVT adapts the Transformer to the dense prediction tasks on many benchmarks which demand large inputs and fine-grained features with computational efficiency. Moreover, both PiT~\cite{heo2021rethinking} and CvT~\cite{wu2021cvt} utilize pooling and convolution to perform token downsampling, respectively. In detail, CvT~\cite{wu2021cvt} improves the SRA of PVT~\cite{wang2021pyramid} by replacing the linear layer with a convolutional projection. Based on the convolutional bias, CvT~\cite{wu2021cvt} can adapt to arbitrary size inputs without positional encodings.

\subsection{Deep Transformer}\label{sec:036}
Empirically, increasing model's depth always strengthens its learning capacity~\cite{he2016deep}. Recent works apply a deep structure to Transformer and massive experiments are conducted to investigate its scalability by analyzing cross-patch~\cite{gong2021diverse} and cross-layer~\cite{zhou2021deepvit,zhou2021refiner} similarities, and the contribution of residual blocks~\cite{touvron2021going}. In the deep Transformer, the features from the deeper layers tend to be less representative (attention collapse~\cite{zhou2021deepvit}), and the patches are mapped into the indistinguishable latent representations (patch over-smoothing~\cite{gong2021diverse}). To address such limitations mentioned above, current methods present the corresponding solutions from two aspects.

From the aspect of model's structure, Touvron et al. present efficient Class-attention in image Transformers (CaiT~\cite{touvron2021going}), including two stages: 1) Multiple self-attention stages without class token. In each layer, a learned diagonal matrix initialized by small values is exploited to update the channel weights dynamically, thereby offering a certain degree of freedom for channel adjustment. 2) Last few class-attention stages with frozen patch embeddings. A later class token is inserted to model global representations, similar to DETR~\cite{carion2020end} with an encoder-decoder structure. This explicit separation is based on the assumption that the class token is invalid for the gradient of patch embeddings in the forward pass. With distillation training strategy~\cite{touvron2021training}, CaiT achieves a new SoTA on imagenet-1k (86.5\% top-1 accuracy) without external data. Although deep Transformer suffers from attention collapse and over-smoothing problems,it still largely preserves the diversity of the attention map between different heads. Based on this observation, Zhou et al. propose Deep Vision Transformer (DeepViT)~\cite{zhou2021deepvit} that aggregates different head attention maps and re-generates a new one by using a linear layer to increase cross-layer feature diversity. Furthermore, Refiner~\cite{zhou2021refiner} applies a linear layer to expand the dimension of the attention maps (indirectly increasing the head number) for diversity promotion. Then, a Distributed Local Attention (DLA) is employed to achieve better modeling of both the local features and the global ones, which is implemented by a head-wise convolution effecting on the attention map.

From the aspect of training strategy, Gong et al. present three Patch Diversity losses for deep Transformer that can significantly encourage patches' diversity and offset over-smoothing problem~\cite{gong2021diverse}. Similar to~\cite{gao2019representation}, a patch-wise cosine loss minimizes pairwise cosine similarity among patches. A patch-wise contrastive loss regularizes the deeper patches by their corresponding one in the early layer. Inspired by Cutmix~\cite{yun2019cutmix}, a patch-wise mixing loss mixes two different images and forces each patch to only attend to the patches from the same image and ignore unrelated ones. Compared with LV-ViT~\cite{jiang2021token}, their similar loss function is based on a distinctive motivation. The former focuses on the patch diversity, while the latter focuses on data augmentation about token labeling.

\subsection{Transformers with Self-Supervised Learning}\label{sec:037}
Following the grateful success of self-supervised in the NLP field~\cite{devlin2018bert}, recent works also attempt to design various self-supervised learning schemes for the visual Transformers in both generative and discriminative ways. 

For the generative models, Chen et al. propose an image Generative Pre-training Transformer (iGPT)~\cite{chen2020generative} for self-supervised visual learning. Different from the patch embedding of ViT~\cite{dosovitskiy2021an}, iGPT directly resizes and flattens the image to a lower resolution sequences. The resized sequences are then input into a GPT-2~\cite{brown2020language} for auto-regressive pixel prediction. iGPT demonstrates the effectiveness of the Transformer in the visual tasks without any help from image-specific knowledge, but its considerable computation cost is hard to be accepted (roughly 2500 V100-days for pre-training). Instead of generating such raw pixels directly, Bao et al. propose a BERT-style~\cite{devlin2018bert} visual Transformer (BEiT)~\cite{bao2021beit} by reconstructing the masked image in the latent space. Similar to the dictionary in BERT~\cite{devlin2018bert}, dVAE~\cite{ramesh2021zero} vectorizes the image into discrete visual tokens. The resulting visual token serves as pseudo label to pre-train ViT for masked patch construction. 

For the discriminative models, Chen et al.~\cite{chen2021empirical} go back to basics and investigate the effects of several fundamental components for stabilized self-supervised ViT training. They observe that the unstable training process mildly affects the eventual performance, and extend MoCo series to MoCo v3, containing a series of training strategies such as freezing projection layer. Following DeiT~\cite{touvron2021training}, Caron et al. further extend the teacher-student recipe to self-supervised learning and propose DINO (2021)~\cite{caron2021emerging}. The core concepts of DINO can be summarized into three points. A momentum encoder inherited SwAV~\cite{caron2020unsupervised}, serves as a teacher model that outputs the centered pseudo labels over a batch. An online encoder without the prediction head serves as a student model to fit the teacher's output. A standard cross-entropy loss connects self-training with knowledge distillation. More interestingly, self-supervised ViT can learn flourishing features for segmentation, which are normally unattainable by the supervised models.

\begin{figure*}[!htbp]
    \includegraphics[width=6.45in]{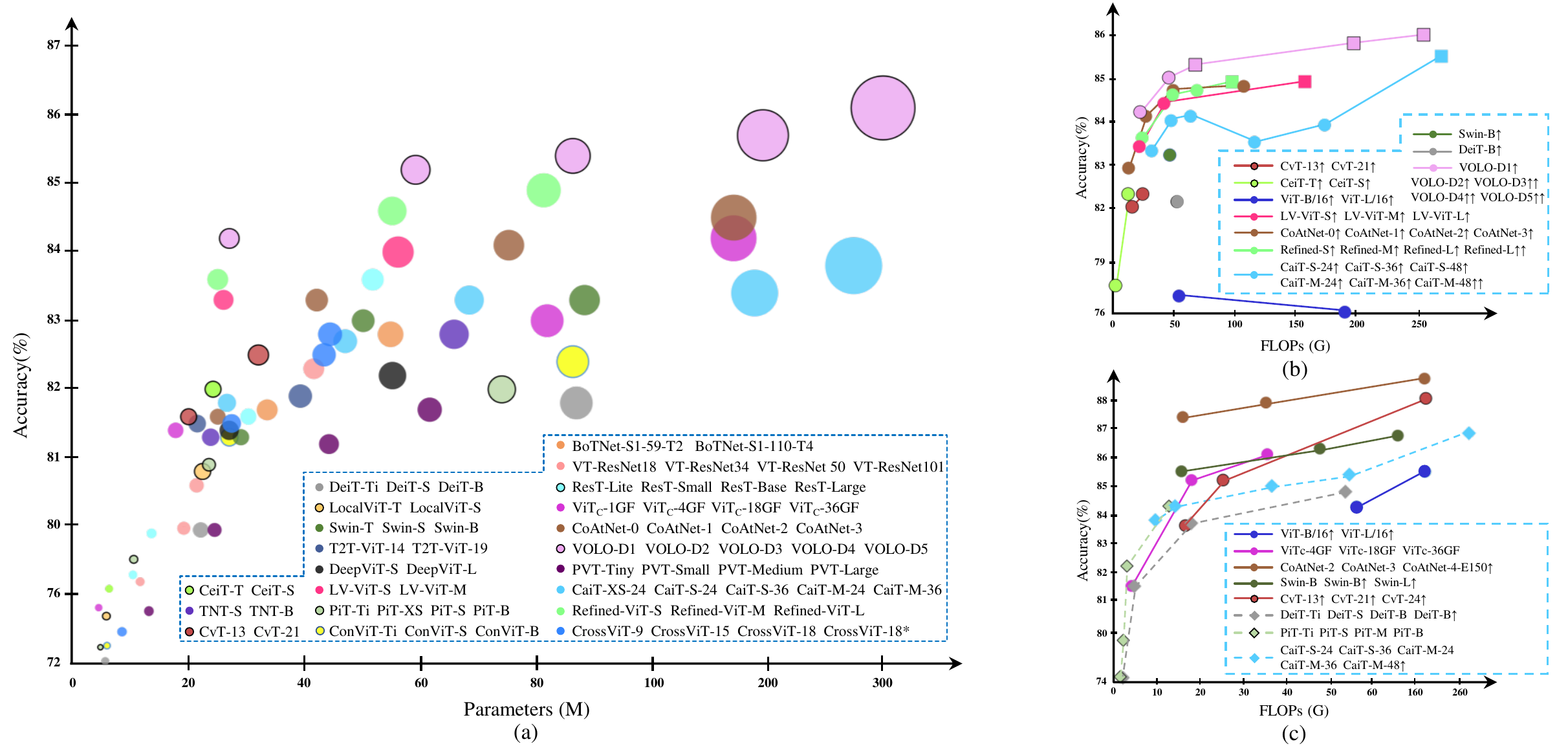}
    \setlength{\abovecaptionskip}{-0.15cm}
    \setlength{\belowcaptionskip}{-0.4cm}
    \centering
    \caption{Comparisons of recent visual Transformers on ImageNet-1k benchmark, including ViT~\cite{dosovitskiy2021an}, DeiT~\cite{touvron2021training}, BoTNet~\cite{srinivas2021bottleneck}, VTs~\cite{wu2020visual}, ConViT~\cite{d2021convit}, CeiT~\cite{yuan2021incorporating}, LocalViT~\cite{li2021localvit}, TNT~\cite{han2021transformer}, Swin~\cite{liu2021swin}, PiT~\cite{heo2021rethinking}, T2T-ViT~\cite{yuan2021tokens}, PVT~\cite{wang2021pyramid}, CvT~\cite{wu2021cvt}, DeepViT~\cite{zhou2021deepvit}, CaiT~\cite{touvron2021going}, Cross ViT~\cite{chen2021crossvit} (best viewed in color). (a) The bubble plot of the mentioned models with $224^2$ resolution input, the size of cycle denotes GFLOPs. (b) Comparison of visual Transformers with high-resolution inputs, the square indicates $448^2$ input resolution. (c) The accuracy plot of some pre-trained models on ImageNet-21k.}
    \label{fig:08}
\end{figure*}

\begin{table*}[htbp]\footnotesize
  \centering \small
  \caption{ Top-1 accuracy comparison of visual Transformers on ImageNet-1k. ``1k Only'' denotes training on ImageNet-1K only; ``21k Pre.'' denotes pre-training on ImageNet-21K and fine-tuning on ImageNet-1K; ``Distill.'' denotes applying distillation training scheme of DeiT~\cite{touvron2021training}; The color of ``legend'' corresponding to each model also denotes same model in Fig.~\ref{fig:08}.}
  \setlength \tabcolsep{1.5pt}
  \begin{spacing}{0.63}
  \scalebox{0.85}{%
  \begin{tabular}{lccccc}
    \toprule[0.15em]
    \multirow{2}[2]{*}[0.7ex]{\textbf{Method}} & \multicolumn{1}{c}{\multirow{2}[2]{*}[0.1ex]{ \makecell[c]{\textbf{\#Params.}\\ \textbf{(M)}}}} & \multicolumn{1}{c}{\multirow{2}[2]{*}[0.1ex]{\makecell[c]{\textbf{FLOPs}\\\textbf{(G)}}}} & \multicolumn{2}{c}{\textbf{ImageNet-1k Top-1 Acc.}} & \multicolumn{1}{c}{\multirow{2}[2]{*}[0.7ex]{\makecell{\textbf{Legend}}}}\\
    \cmidrule{4-5}    
    \multicolumn{1}{c}{} & & & \multicolumn{1}{c}{1K} & \multicolumn{1}{c}{21K/Distill.($\dagger$/$\Upsilon$)} & \\ 
    \toprule
    ViT-B/16↑$_{384}$~\cite{dosovitskiy2021an} & 86.8    & 49.4   & 77.9  & 83.97$\dagger$ & \multirow{2}[0]{*}{\includegraphics[height=0.2cm]{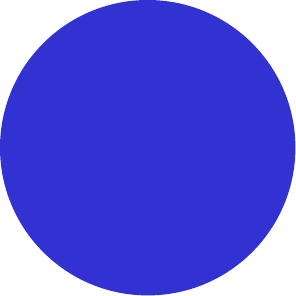}} \\
    ViT-L/16↑$_{384}$~\cite{dosovitskiy2021an} & 304.7   & 174.8  & 76.5  & 85.15$\dagger$ &  \\
    \specialrule{0.5pt}{0.7pt}{1.5pt}
    VT-Rest18~\cite{wu2020visual} & 11.7  & 1.57  & 76.8  & -     & \multirow{4}[0]{*}{\includegraphics[height=0.2cm]{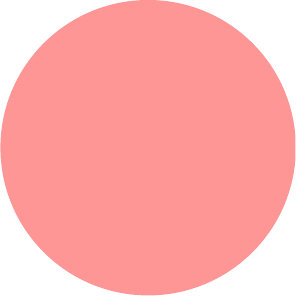}} \\
    VT-Rest34~\cite{wu2020visual} & 19.2  & 3.24  & 79.9  & -     &  \\
    VT-Rest50~\cite{wu2020visual} & 21.4  & 3.41  & 80.6  & -     &  \\
    VT-Rest101~\cite{wu2020visual} & 41.5  & 7.13  & 82.3  & -     &  \\
    \specialrule{0.5pt}{0.7pt}{1.5pt}
    BoTNet-T2~\cite{srinivas2021bottleneck} & 33.5  & 7.3   & 81.7  & -     & \multirow{3}[0]{*}{\includegraphics[height=0.2cm]{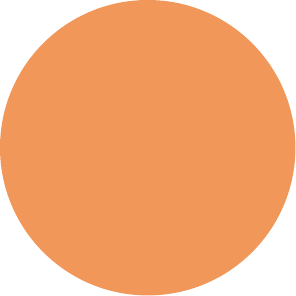}} \\
    BoTNet-T4~\cite{srinivas2021bottleneck} & 54.7  & 10.9  & 82.8  & -     &  \\
    BoTNet-T5↑$_{256}$~\cite{srinivas2021bottleneck} & 75.1  & 19.3  & 83.5  & -     &  \\
    \specialrule{0.5pt}{0.7pt}{1.5pt}
    DeiT-Ti~\cite{touvron2021training} & 5.7   & 1.1   & 72.2  & 74.5$\Upsilon$  & \multirow{4}[0]{*}{\includegraphics[height=0.2cm]{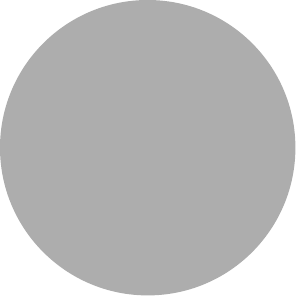}} \\
    DeiT-S~\cite{touvron2021training} & 22.1  & 4.3   & 79.8  & 81.2$\Upsilon$   &  \\
    DeiT-B~\cite{touvron2021training} & 86.6  & 16.9 & 81.8  & 83.4$\Upsilon$   &  \\
    DeiT-B↑$_{384}$~\cite{touvron2021training} & 86.8  & 49.4  & 83.1  & 84.5$\Upsilon$   &  \\
    \specialrule{0.5pt}{0.7pt}{1.5pt}
    ConViT-Ti~\cite{d2021convit} & 6     & 1     & 73.1  & -     & \multirow{3}[0]{*}{\includegraphics[height=0.2cm]{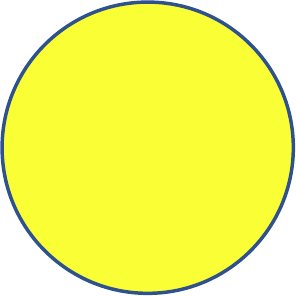}} \\
    ConViT-S~\cite{d2021convit} & 27    & 5.4   & 81.3  & -     &  \\
    ConViT-B~\cite{d2021convit} & 86    & 17    & 82.4  & -     &  \\
    \specialrule{0.5pt}{0.7pt}{1.5pt}
    LocalViT-T~\cite{li2021localvit} & 5.9   & 1.3   & 74.8  & -     & \multirow{2}[0]{*}{\includegraphics[height=0.2cm]{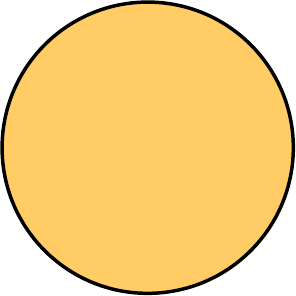}} \\
    LocalViT-S~\cite{li2021localvit} & 22.4  & 4.6   & 80.8  & -     &  \\
    \specialrule{0.5pt}{0.7pt}{1.5pt}
    CeiT-T~\cite{yuan2021incorporating} & 6.4   & 1.2   & 76.4  & -     & \multirow{4}[0]{*}{\includegraphics[height=0.2cm]{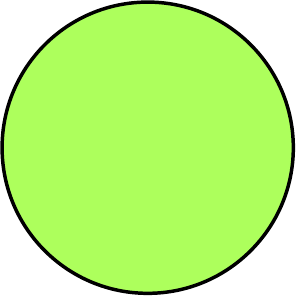}} \\
    CeiT-S~\cite{yuan2021incorporating} & 24.2  & 4.5   & 82.0    & -     &  \\
    CeiT-T↑$_{384}$~\cite{yuan2021incorporating} & 6.4   & 3.6   & 78.8  & -     &  \\
    CeiT-S↑$_{384}$~\cite{yuan2021incorporating} & 24.2  & 12.9  & 83.3  & -     &  \\
    \specialrule{0.5pt}{0.7pt}{1.5pt}
    ResT-Small~\cite{zhang2021rest} & 13.7  & 1.9   & 79.6  & -     & \multirow{3}[0]{*}{\includegraphics[height=0.2cm]{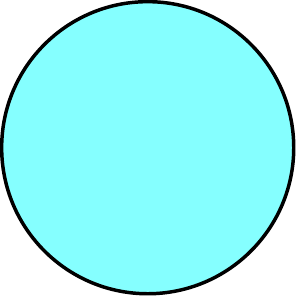}} \\
    ResT-Base~\cite{zhang2021rest} & 30.3  & 4.3   & 81.6  & -     &  \\
    ResT-Large~\cite{zhang2021rest} & 51.6  & 7.9   & 83.6  & -     &  \\
    \specialrule{0.5pt}{0.7pt}{1.5pt}
    ViT$_C$-1GF~\cite{zhang2021multi} & 4.6   & 1.1   & 75.3  & -     & \multirow{4}[0]{*}{\includegraphics[height=0.2cm]{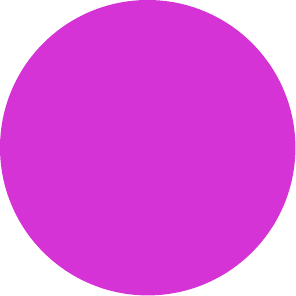}} \\
    ViT$_C$-4GF~\cite{zhang2021multi} & 17.8  & 4.0     & 81.4  & 81.2$\dagger$  &  \\
    ViT$_C$-18GF~\cite{zhang2021multi} & 81.6  & 17.7  & 83.0    & 84.9$\dagger$  &  \\
    ViT$_C$-36GF~\cite{zhang2021multi} & 167.8 & 35    & 84.2  & 85.8$\dagger$  &  \\
    \specialrule{0.5pt}{0.7pt}{1.5pt}
    CoAtNet-0~\cite{dai2021coatnet} & 25    & 4.2   & 81.6  & -     & \multirow{5}[0]{*}{\includegraphics[height=0.2cm]{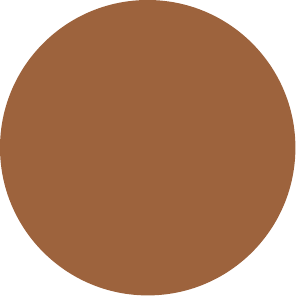}} \\
    CoAtNet-1~\cite{dai2021coatnet} & 42    & 8.4   & 83.3  & -     &  \\
    CoAtNet-2~\cite{dai2021coatnet} & 75    & 15.7  & 84.1  & 87.1$\dagger$  &  \\
    CoAtNet-3~\cite{dai2021coatnet} & 168   & 34.7  & 84.5  & 87.6$\dagger$  &  \\
    CoAtNet-4↑$_{384}$~\cite{dai2021coatnet} & 275   & 189.5 & -     & 88.4$\dagger$  &  \\
    \specialrule{0.5pt}{0.7pt}{1.5pt}
    TNT-S~\cite{han2021transformer} & 23.8  & 5.2   & 81.3  & -     & \multirow{4}[0]{*}{\includegraphics[height=0.2cm]{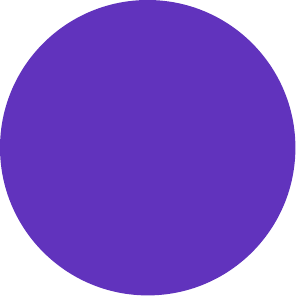}} \\
    TNT-B~\cite{han2021transformer} & 65.6  & 14.1  & 82.8  & -     &  \\
    TNT-S↑$_{384}$~\cite{han2021transformer} & 23.8  & -     & 83.1  & -     &  \\
    TNT-B↑$_{384}$~\cite{han2021transformer} & 65.6  & -     & 83.9  & -     &  \\
    \specialrule{0.5pt}{0.7pt}{1.5pt}
    Swin-T~\cite{liu2021swin} & 29 & 4.5 & 81.3 & -  & \multirow{4}[0]{*}{\includegraphics[height=0.2cm]{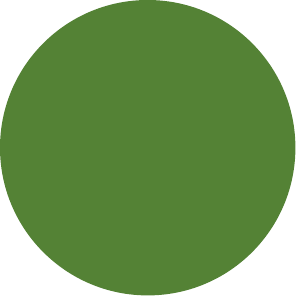}}\\
    Swin-S~\cite{liu2021swin} & 50 & 8.7 & 83.0 & - & \\
    Swin-B~\cite{liu2021swin} & 88 & 15.4 & 83.3 & 85.2$\dagger$ & \\
    Swin-L↑$_{384}$~\cite{liu2021swin} & 197 & 104 & - & 86.4$\dagger$ & \\
    \specialrule{0.5pt}{0.7pt}{1.5pt}
    LV-ViT-S~\cite{jiang2021token} & 26    & 6.6   & 83.3  & -     & \multirow{5}[0]{*}{\includegraphics[height=0.2cm]{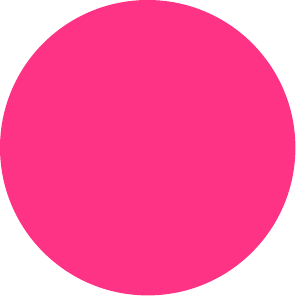}} \\
    LV-ViT-M~\cite{jiang2021token} & 56    & 16.0    & 84.0    & -     &  \\
    LV-ViT-L↑$_{288}$~\cite{jiang2021token} & 150   & 59.0    & 85.3  & -     &  \\
    LV-ViT-M↑$_{384}$~\cite{jiang2021token} & 56    & 42.2  & 85.4  & -     &  \\
    LV-ViT-L↑$_{448}$~\cite{jiang2021token} & 150   & 157.2 & 85.9  & -     &  \\
    \bottomrule[0.15em]
    \end{tabular}}
    \scalebox{0.85}{%
    \begin{tabular}{lccccc}
    \toprule[0.15em]
    \multirow{2}[2]{*}[0.7ex]{\textbf{Method}} & \multicolumn{1}{c}{\multirow{2}[2]{*}[0.1ex]{ \makecell[c]{\textbf{\#Params.}\\ \textbf{(M)}}}} & \multicolumn{1}{c}{\multirow{2}[2]{*}[0.1ex]{\makecell[c]{\textbf{FLOPs}\\\textbf{(G)}}}} & \multicolumn{2}{c}{\textbf{ImageNet-1k Top-1 Acc.}} & \multicolumn{1}{c}{\multirow{2}[2]{*}[0.7ex]{\makecell{\textbf{Legend}}}}\\
    \cmidrule{4-5}    
    \multicolumn{1}{c}{} & & & \multicolumn{1}{c}{1K} & \multicolumn{1}{c}{21K/Distill.($\Upsilon$) } & \\ 
    \toprule
    VOLO-D1~\cite{yuan2021volo} & 27    & 6.8   & 84.2  & -     & \multirow{8}[0]{*}{\includegraphics[height=0.2cm]{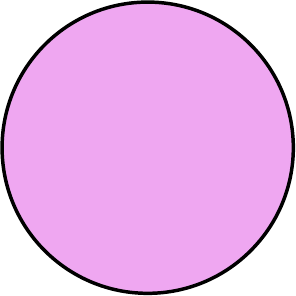}} \\
    VOLO-D2~\cite{yuan2021volo} & 59    & 14.1  & 85.2  & -     &  \\
    VOLO-D3~\cite{yuan2021volo} & 86    & 20.6  & 85.4  & -     &  \\
    VOLO-D4~\cite{yuan2021volo} & 193   & 43.8  & 85.7  & -     &  \\
    VOLO-D5~\cite{yuan2021volo} & 296   & 69.0    & 86.1  & -     &  \\
    VOLO-D3↑$_{448}$~\cite{yuan2021volo} & 86    & 67.9  & 86.3  & -     &  \\
    VOLO-D4↑$_{448}$~\cite{yuan2021volo} & 193   & 197   & 86.8  & -     &  \\
    VOLO-D5↑$_{448}$~\cite{yuan2021volo} & 296   & 304   & 87.0    & -     &  \\
    \specialrule{0.5pt}{0.985pt}{1.785pt}
    T2T-ViT-14~\cite{yuan2021tokens} & 21.5  & 5.2   & 81.5  & -     & \multirow{2}[0]{*}{\includegraphics[height=0.2cm]{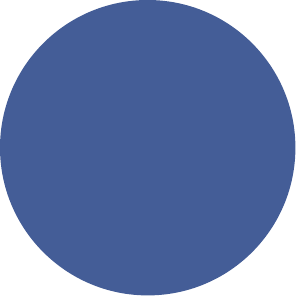}} \\
    T2T-ViT-19~\cite{yuan2021tokens} & 39.2  & 8.9   & 81.9  & -     &  \\
    \specialrule{0.5pt}{0.985pt}{1.785pt}
    PVT-Ti~\cite{wang2021pyramid} & 13.2  & 1.9   & 75.1  & -     & \multirow{6}[0]{*}{\includegraphics[height=0.2cm]{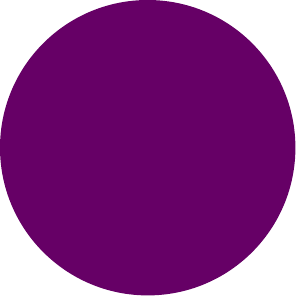}} \\
    PVT-S~\cite{wang2021pyramid} & 24.5  & 3.8   & 79.8  & -     &  \\
    PVT-M~\cite{wang2021pyramid} & 44.1  & 6.7   & 81.2  & -     &  \\
    PVT-L~\cite{wang2021pyramid} & 61.4  & 9.8   & 81.7  & -     &  \\
    \specialrule{0pt}{0.985pt}{0pt}
    \hdashline
    \specialrule{0pt}{0pt}{1.785pt}
    PVTv2-B2~\cite{wang2021pvtv2} & 25.4  & 4.0  & 82.0  & -     &  \\
    PVTv2-B4~\cite{wang2021pvtv2} & 62.6  & 10.1  & 83.6  & -     &  \\
    \specialrule{0.5pt}{0.985pt}{1.785pt}
    PiT-Ti~\cite{heo2021rethinking} & 4.9   & 0.7   & 73.0    & 74.6$\Upsilon$   & \multirow{4}[0]{*}{\includegraphics[height=0.2cm]{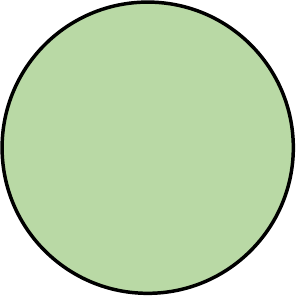}} \\
    PiT-XS~\cite{heo2021rethinking} & 10.6  & 1.4   & 78.1  & 79.1$\Upsilon$   &  \\
    PiT-S~\cite{heo2021rethinking} & 23.5  & 2.9   & 80.9  & 81.9$\Upsilon$   &  \\
    PiT-B~\cite{heo2021rethinking} & 73.8  & 12.5  & 82.0    & 84.0$\Upsilon$     &  \\
    \specialrule{0.5pt}{0.985pt}{1.785pt}
    CvT-13~\cite{wu2021cvt} & 20    & 4.5   & 81.6  & -     & \multirow{5}[0]{*}{\includegraphics[height=0.2cm]{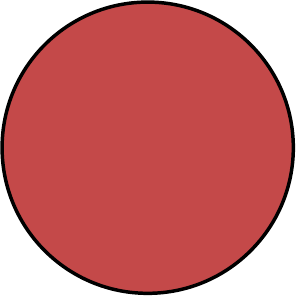}} \\
    CvT-21~\cite{wu2021cvt} & 32    & 7.1   & 82.5  & -     &  \\
    CvT-13↑$_{384}$~\cite{wu2021cvt} & 20    & 16.3  & 83.0    & 83.3$\dagger$  &  \\
    CvT-21↑$_{384}$~\cite{wu2021cvt} & 32    & 24.9  & 83.3  & 84.9$\dagger$  &  \\
    CvT-W24↑$_{384}$~\cite{wu2021cvt} & 277   & 193.2 & -     & 87.7$\dagger$  &  \\
    \specialrule{0.5pt}{0.985pt}{1.785pt}
    DeepViT-S~\cite{zhou2021deepvit} & 27    & 6.2   & 82.3  & -     & \multirow{2}[0]{*}{\includegraphics[height=0.2cm]{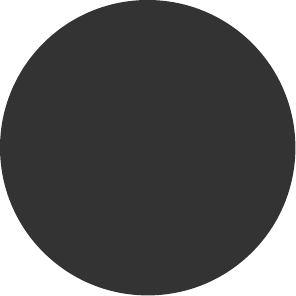}} \\
    DeepViT-L~\cite{zhou2021deepvit} & 55    & 12.5  & 83.1  & -     &  \\
    \specialrule{0.5pt}{0.985pt}{1.785pt}
    CaiT-XS-24~\cite{touvron2021going} & 26.6  & 5.4   & 81.8  & 82.0$\Upsilon$     & \multirow{5}[0]{*}{\includegraphics[height=0.2cm]{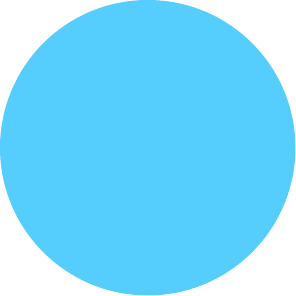}} \\
    CaiT-S-24~\cite{touvron2021going} & 46.9  & 9.4   & 82.7  & 83.5$\Upsilon$   &  \\
    CaiT-S-36~\cite{touvron2021going} & 68.2  & 13.9  & 83.3  & 84.0$\Upsilon$     &  \\
    CaiT-M-24~\cite{touvron2021going} & 185.9 & 36.0    & 83.4  & 84.7$\Upsilon$   &  \\
    CaiT-M-36~\cite{touvron2021going} & 270.9 & 53.7  & 83.8  & 85.1$\Upsilon$   &  \\
    \specialrule{0.5pt}{0.985pt}{1.785pt}
    DiversePatch-S12~\cite{gong2021diverse} & 22    & -     & 81.2  & -     &  \\
    DiversePatch-S24~\cite{gong2021diverse} & 44    & -     & 82.2  & -     &  \\
    DiversePatch-B12~\cite{gong2021diverse} & 86    & -     & 82.9  & -     &  \\
    DiversePatch-B24~\cite{gong2021diverse} & 172   & -     & 83.3  & -     &  \\
    DiversePath-B12↑$_{384}$~\cite{gong2021diverse} & 86    & -     & 84.2  & -     &  \\
    \specialrule{0.5pt}{0.985pt}{1.785pt}
    Refined-ViT-S~\cite{zhou2021refiner} & 25    & 7.2   & 83.6  & -     & \multirow{5}[0]{*}{\includegraphics[height=0.2cm]{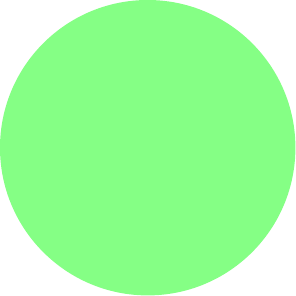}} \\
    Refined-ViT-M~\cite{zhou2021refiner} & 55    & 13.5  & 84.6  & -     &  \\
    Refined-ViT-L~\cite{zhou2021refiner} & 81    & 19.1  & 84.9  &       &  \\
    Refined-ViT-M↑$_{384}$~\cite{zhou2021refiner} & 55    & 49.2  & 85.6  & -     &  \\
    Refined-ViT-L↑$_{384}$~\cite{zhou2021refiner} & 81    & 69.1  & 85.7  & -     &  \\
    \specialrule{0.5pt}{0.985pt}{1.785pt}
    CrossViT-9~\cite{chen2021crossvit} & 8.6   & 1.8   & 73.9  & -     & \multirow{5}[0]{*}{\includegraphics[height=0.2cm]{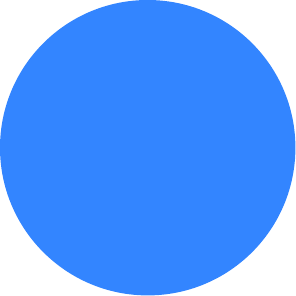}} \\
    CrossViT-15~\cite{chen2021crossvit} & 27.4  & 5.8   & 81.5  & -     &  \\
    CrossViT-18~\cite{chen2021crossvit} & 43.3  & 9.0     & 82.5  & -     &  \\
    CrossViT-15*↑$_{384}$~\cite{chen2021crossvit} & 28.5  & 21.4  & 83.5  & -     &  \\
    CrossViT-18*↑$_{384}$~\cite{chen2021crossvit} & 44.6  & 32.4  & 83.9  & -     &  \\
    \bottomrule[0.15em]
    \end{tabular}}
    \end{spacing}
  \label{tab:01}
\end{table*}

\subsection{Discussion}\label{sec:038}
\subsubsection{Algorithm Evaluation and Comparative Analysis}\label{sec:0381}
In our taxonomy, all the existing supervised models are grouped into six categories. Tab.~\ref{tab:01} summarizes the performances of these existing visual Transformers on  ImageNet-1k benchmarks. To evaluate them objectively and intuitively, we use the following three figures to illustrate their performances on ImageNet-1k under different configurations. Fig.~\ref{fig:08}(a) summarizes the accuracy of each model under $224^2$ inputs size. Fig.~\ref{fig:08}(b) takes the FLOPs as the horizontal axis, which focuses on their performances under higher-resolution. Fig.~\ref{fig:08}(c) focuses on the pre-trained models with external datasets. From these comparison results, we briefly summarize several performance improvements on efficiency and scalability as follows.

\begin{itemize}
  \item   Compared with the most structure-improved methods, the basic training strategies like DeiT~\cite{touvron2021training} and LV-ViT~\cite{jiang2021token}, are more universal for various models, tasks, and inputs.
  
  \item The locality is indispensable for the Transformer, which is reflected by the dominant of VOLO~\cite{yuan2021volo} and Swin~\cite{liu2021swin} on classification and dense prediction tasks, respectively.
  
  \item The convolutional patchify stem (ViT$_c$~\cite{xiao2021early}) and early convolutional stage (CoAtNet~\cite{dai2021coatnet}) can significantly boost the accuracy of the Transformers, especially for large models. We speculate the reason is because these designs introduce a more stringent high-level features than the non-overlapping patch projection in ViT~\cite{dosovitskiy2021an}.
  
  \item The deep Transformer, such as Refined-ViT~\cite{zhou2021refiner} and CaiT~\cite{touvron2021going}, has great potential. As the model size grows quadratically with the channel dimension, the trade-off in deep Transformer is considered for further investigation.
  
  \item CeiT~\cite{yuan2021incorporating} and CvT~\cite{wu2021cvt} show significant advantages in training a small or medium model (0$\--$40M), which suggests that such kinds of hybrid attention blocks for lightweight models are worth further exploring.
\end{itemize}

\subsubsection{Brief Discussion on Alternatives}\label{sec:0383}
During the development of the visual Transformers, the most common question is whether the visual Transformers can replace the traditional convolution completely. By reviewing the history of the performance improvements in the last year, there is no sign of relative inferiority here. The visual Transformers have returned from a pure structure to a hybrid form, and the global information has gradually returned to a mixed stage with the locality bias. Although the visual Transformers can be equivalent to CNN or even has a better modeling capability, such a simple and effective convolution operation is enough to process the locality and the semantic features in the shallow layer. In the future, the spirit of combining both of them shall drive more breakthroughs for image classification. 

\section{Transformer for Detection}\label{sec:04}
In this section, we review visual Transformers for object detection, which can be grouped into two folds: \textit{Transformer as the neck} and \textit{Transformer as the backbone}. For the neck detectors, we mainly focus on a new representation specified to the Transformer structure, called object query, that a set of learned parameters aggregate instance features from input images. The recent variants try to solve an optimal fusion paradigm in terms of either convergence acceleration or performance improvement. Besides these neck design, a proportion of backbone detectors also take specific strategies into consideration. Finally, we evaluate them and analyze some potential methods for these detectors.

\subsection{Transformer Neck}\label{sec:041}
We first review DETR~\cite{carion2020end} and Pix2seq~\cite{chen2021pix2seq}, the \textit{original Transformer detectors} that reformulate two different paradigms for object detection. Subsequently, we mainly focus on the DETR-based variants, improving such Transformer detector in accuracy and convergence from five aspects: \textit{sparse attention}, \textit{spatial prior}, \textit{structural redesign}, \textit{assignment optimization}, and \textit{pre-training model}.

\subsubsection{The Original Detectors}\label{sec:0411}
DEtection with TRansformer (DETR)~\cite{carion2020end} is the first end-to-end Transformer detector that eliminates hand-designed representations~\cite{ren2017faster,redmon2016you,lin2017focal,zhao2019object} and non-maximum suppression (NMS) post-processing, which redefines the object detection as a set prediction problem. In detail, a small set of learnable positional encodings, called object queries, that are parallelly fed into the Transformer decoder to extract the instance information from the image features. Then, these object queries are independently predicted to be a detection result. Instead of the vanilla k-class classification, a special class, no object label ($\varnothing$) is added for k+1 class classification. During the training process, a bipartite matching strategy is applied between the predicted objects and the ground-truth to identify one-to-one label assignment, 
\begin{figure}[htbp]
    \centering
    \includegraphics[width=3.5in]{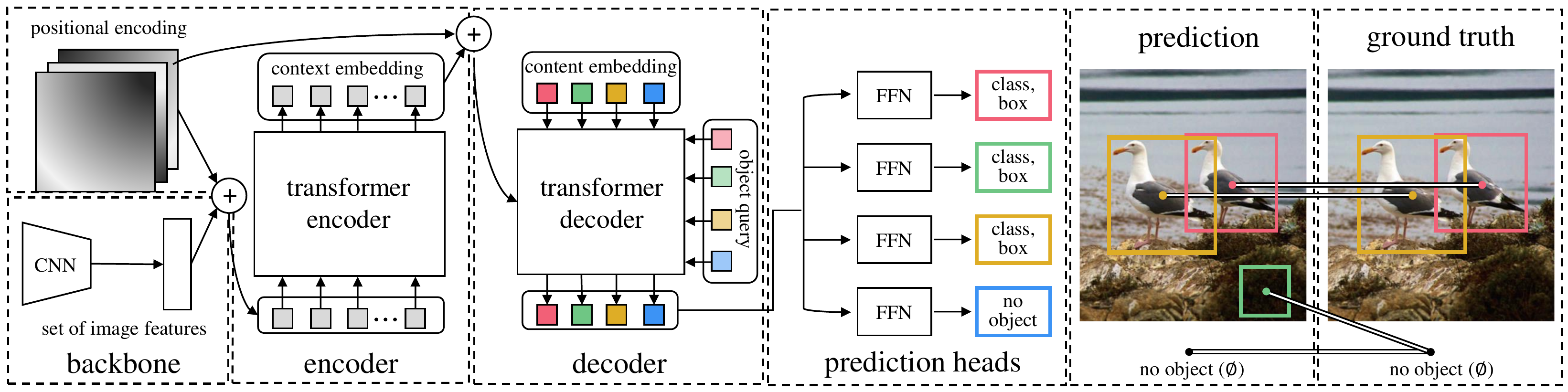} 
    \caption{An overview of DETR. (Modified from~\cite{carion2020end}.)}
    \label{fig:09}
\end{figure}
hence removing the redundant predictions at the inference time without NMS. In back propagation, a Hungarian loss includes a log-likelihood loss for all classification results and a box loss for all the matched pairs. More details about the Hungarian matching strategy are available in the App.~\ref{sup:detr}.

Overall, DETR provides a new paradigm for end-to-end object detection. The object query gradually learns an instance representation during the interaction with image features. The bipartite matching allows a direct set prediction and easily joints to the one-to-one label assignment, hence eliminating traditional post-processing. DETR achieves competitive performance on the COCO benchmark but suffers from slow convergence as well as poor performance on small objects.

Another pioneered work is Pix2seq~\cite{chen2021pix2seq}, treating generic object detection as a language modeling task. Given an image input, a vanilla sequential Transformer is executed to extract features and generate a series of object descriptions (i.e. class labels and bounding boxes) auto-regressively. Such a simplified but more elaborate image caption method is derived under the assumption that if a model learns about both location and label of an object, it can be taught to produce a description with specified sequence~\cite{chen2021pix2seq}. Compared with DETR, Pix2seq attains a better result on small objects. How to combine both kinds of concepts is worthy of further consideration.

\subsubsection{Transformer with Sparse Attention}\label{sec:0412}
In DETR, the dense interaction across both object queries and image features costs unbearable resources and slows down the convergence of DETR. Therefore, the most recent efforts aim to design a data-dependent sparse attention to address these issues.

Following~\cite{dai2017deformable}, Zhu et al. develop Deformable DETR to ameliorate both training convergence and detection performance significantly via multi-scale deformable attention~\cite{zhu2021deformable}. Compared with original DETR, the deformable attention module only samples a small set of key (reference) points for full features aggregation. Such sparse attention can be easily stretched to multi-scale feature fusion without the help of FPN~\cite{lin2017feature}. Moreover, an iterative bounding box refinement and a two-stage prediction strategy (Sec.~\ref{sec:0413}) are developed to further enhance the detection accuracy. Empirically, Deformable DETR achieves a higher accuracy (especially for small objects) with $10\times$ less training epochs and reduces the computing complexity to $O(2N_qC^2+\text{min}(HWC^2,N_qKC^2))$ with $1.6\times$ faster inference speed. Please see the App.~\ref{sup:deformable-detr} for more details of deformable attention mechanism.

By visualizing the attention map of DETR~\cite{carion2020end}, Zheng et al. observe that the semantically similar and spatially close elements always have a similar attention map in the encoder~\cite{zheng2020end}. Then they present an Adaptive Clustering Transformer (ACT), leveraging a multi-round sensitivity hashing to dynamically cluster the queries into different prototypes. The attention map of the prototype is then broadcast to their corresponding queries. Unlike the redesign of on sparse attention, Wang et al. introduce a Poll and Pool (PnP) sampling model~\cite{wang2021pnp} to extract the fine foreground features and condense the contextual background features into a smaller one. Such fine-coarse tokens are then fed into DETR to generate the detection results. Instead of the input sparsification, Sparse DETR~\cite{roh2021sparse} applies a hysteretic scoring network (corresponding to the Poll operation in~\cite{wang2021pnp}) to update the expected tokens selectively within the transformer encoder, where the top-k selected tokens are supervised by pseudo labels from the binarized decoder cross-attention map with BCE loss. 

\subsubsection{Transformer with Spatial Prior}\label{sec:0413}
Unlike anchor or other representations directly generated by content and geometry features~\cite{ren2017faster,tian2019fcos}, object queries implicitly model the spatial information with random initialization, which is weakly related with the bounding box. The mainstream for spatial prior applications are the one-stage detector with empirical spatial information and the two-stage detector with geometric coordinates initialization or Region-of-Interest (RoI) features.

In one-stage methods, Gao et al. suggest Spatially Modulated Cross-Attention (SMCA)~\cite{gao2021fast} to estimate the object queries' spatial prior explicitly. Specifically, a Gaussian-like weight map generated by object queries is multiplied with the corresponding cross-attention map to augment the RoI for convergence acceleration. Furthermore, both intra-scale and multi-scale self-attention layers are utilized in the Transformer encoder for multi-scale feature aggregation, and the scale-selection weights generated from object queries are applied for scale-query adaptation. Meng et al.~\cite{meng2021conditional} extract the spatial attention map from the cross-attention formulation and observe that the extreme region of such attention map has larger deviations at the early training. Consequently, they propose Conditional DETR where a new spatial prior mapped from reference points is adopted in the cross-attention layer, thereby attending to extreme regions of each object explicitly. The reference point is predicted by the object query or serves as a learned parameter replacing the object query. Following~\cite{meng2021conditional}, Anchor DETR~\cite{wang2021anchor} suggests to explicitly learn the 2D anchor points ([$cx,cy$]) and different anchor patterns instead of the high-dimensional spatial embedding. Similar to~\cite{ren2017faster}, the pattern embeddings are assigned to the meshed anchor points so that they can detect different scale objects anywhere. DAB-DETR~\cite{liu2022dab} then extends the 2D concept to a 4D anchor box ([$cx,cy,w,h$]) to explicitly provide proposal bounding box information during the cross-attention. With the auxiliary decoder loss of the coordinates offset~\cite{zhu2021deformable}, such a 4D box can be dynamically refined layer-by-layer in the decoder. However, the same reference boxes/points may severely deteriorate queries' saliency and confuses detector due to the indiscriminative spatial prior. By assigning query-specific reference points to object queries, SAP-DETR~\cite{liu2022sap} only predicts the distance from each side of the bounding box to these points. Such a query-specific prior discrepancies queries' saliency and pave the way for fast model convergency.

\begin{table}[htbp]
  \centering
  \caption{Comparison between Transformer necks and representative CNNs with ResNet-50 backbone on COCO 2017 val set. }
  \setlength \tabcolsep{1.55pt}
    \begin{spacing}{0.65}
    \resizebox{\linewidth}{!}{%
    \begin{tabular}{lccccccc}
    \toprule
    \multicolumn{1}{l}{\textbf{Method}}  & \textbf{Epochs} & \textbf{\makecell[c]{FLOPs \\ (G)}} & \textbf{\makecell[c]{\#Para.  \\ (M)}} & \textbf{FPS} & \textbf{MS} & \textbf{AP}/$\textbf{AP}_{50}$/$\textbf{AP}_{75}$ & $\textbf{Ap}_S$/$\textbf{Ap}_M$/$\textbf{Ap}_{L}$ \\
    \specialrule{0.9pt}{1.6pt}{3.6pt}
    \multicolumn{8}{l}{\textit{CNN Backbone with Other Representations}} \\
    \specialrule{0.5pt}{0.8pt}{1.7pt}
    FCOS~\cite{tian2019fcos,sun2021rethinking} & 36    & 177   &  - & 17    &    $\checkmark$   & 41.0 /59.8/44.1  & 26.2/44.6/52.2 \\
    Faster R-CNN~\cite{ren2017faster}        & 37    & 180   & 42    & 26    &   $\checkmark$    & 40.2/61.0/43.8  & 24.2/43.5/52.0 \\
    Faster R-CNN+~\cite{ren2017faster}         & 109   & 180   & 42    & 26    &   $\checkmark$    & 42.0 /62.1/45.5  & 26.6/45.4/53.4 \\
    Mask R-CNN~\cite{he2017mask}        & 36    & 260   & 44  & -     & $\checkmark$    & 41.0/61.7/44.9  &  \  - \   / \  - \   / \  - \  \\
    Cas. Mask R-CNN~\cite{cai2018cascade}          & 36    & 739   & 82    & 18    & $\checkmark$    & 46.3/64.3/50.5  & \  - \   / \  - \   / \  - \ \\
    \specialrule{0.5pt}{0.8pt}{1.7pt}
    \multicolumn{8}{l}{\textit{Transformer Model as Neck}} \\
    \specialrule{0.5pt}{0.8pt}{1.7pt}
    DETR-R50~\cite{carion2020end}   & 500   & 86    & 41    & 28    &   \ding{55}   & 42.0/62.4/44.2  & 20.5/45.8/61.1 \\
    DETR-DC5~\cite{carion2020end}   & 500   & 187   & 41    & 12    &   \ding{55}    & 43.3/63.1/45.9  & 22.5/47.3/61.1 \\
    \specialrule{0.5pt}{0.8pt}{1.7pt}
    Pix2seq~\cite{chen2021pix2seq}     &  300     &    -   & 37    &   -    &   \ding{55}   & 43.0/61.0/45.6  & 25.1/46.9/59.4 \\
    Pix2seq-DC5~\cite{chen2021pix2seq}      &  300     &   -    & 38    &   -    &   \ding{55}    & 43.2/61.0/46.1  & 26.6/47.0/58.6 \\
    \specialrule{0.5pt}{0.8pt}{1.7pt}
    Defor. DETR~\cite{zhu2021deformable}  & 50    & 78    & 34    & 23    &  \ding{55}     & 39.7/60.1/42.4  & 21.2/44.3/56.0 \\
    Defor. DETR-DC5~\cite{zhu2021deformable} & 50    & 128   & 34    & 22    &   \ding{55}    & 41.5/61.8/44.9  & 24.1/45.3/56.0 \\
    Defor. DETR-Iter~\cite{zhu2021deformable}  & 50    & 173   & 40    & 19    & $\checkmark$ & 43.8/62.6/47.7  & 26.4/47.1/58.0 \\
    Defor. DETR-Two~\cite{zhu2021deformable}  & 50    & 173   & 40    & 19    &   $\checkmark$    & 46.2/65.2/50.0    & 28.8/49.2/61.7 \\
    \specialrule{0.5pt}{0.8pt}{1.7pt}
    ACT-DC5 (L=16)~\cite{zheng2020end}  & MTKD~\cite{zheng2020end}      & 156   & -     & 14    &   \ding{55}  & 40.6/ \,\  - \,\ / \,\  - \,\      & 18.5/44.3/59.7 \\
    ACT-DC5 (L=32)~\cite{zheng2020end}  & MTKD~\cite{zheng2020end}      & 169   & -     & 16    &   \ding{55}    & 43.1/ \,\  - \,\ / \,\  - \,\     & 22.2/47.1/61.4 \\
    \specialrule{0.5pt}{0.8pt}{1.7pt}
    PnP-DETR-0.33~\cite{wang2021pnp}     & 500   & 77  & - &   -    &    \ding{55}   & 41.1/61.5/43.7  & 20.8/44.6/60.0 \\
    PnP-DETR-0.5~\cite{wang2021pnp}    & 500   & 79  & - &   -    &    \ding{55}   & 41.8/62.1/44.4  & 21.2/45.3/60.8 \\
    PnP-DETR-DC5-0.5~\cite{wang2021pnp}     & 500   & 136 & - &   -    &   \ding{55}    & 43.1/63.4/45.3  & 22.7/46.5/61.1 \\
    \specialrule{0.5pt}{0.8pt}{1.7pt}
    Sparse-DETR-0.1~\cite{roh2021sparse}     &    50   & 105   & 41    & 25  &   $\checkmark$    & 45.3/65.8/49.3  & 28.4/48.3/60.1 \\
    Sparse-DETR-0.5~\cite{roh2021sparse}     &   50    & 136   & 41    & 21  &   $\checkmark$    & 46.3/66.0/50.1  & 29.0/49.5/60.8 \\
    \specialrule{0.5pt}{0.8pt}{1.7pt}
    SMCA~\cite{gao2021fast}  & 50     & 86   & 40    & 22    &   \ding{55}   & 41.0/ \,\  - \,\ / \,\  - \,\     & 21.9/44.3/59.1 \\
    SMCA+~\cite{gao2021fast} & 108   & 86   & 40    & 22    &   \ding{55}    & 42.7/ \,\  - \,\ / \,\  - \,\      & 22.8/46.1/60.0 \\
    SMCA~\cite{gao2021fast}    & 50    & 152   & 40    & 10    & $\checkmark$ & 43.7/63.6/47.2  & 24.2/47.0 /60.4 \\
    SMCA+~\cite{gao2021fast}   & 108   & 152   & 40    & 10    &   $\checkmark$    & 45.6/65.5/49.1  & 25.9/49.3/62.6 \\
    \specialrule{0.5pt}{0.8pt}{1.7pt}
    Condit. DETR~\cite{meng2021conditional}  & 108   & 90    & 44    & 17     &     \ding{55}  & 43.0/64.0/45.7  & 22.7/46.7/61.5 \\
    Condit. DETR-DC5~\cite{meng2021conditional}  & 108   & 195   & 44    & 11     &   \ding{55}   & 45.1/65.4/48.5  & 25.3/49.0/62.2 \\
    \specialrule{0.5pt}{0.8pt}{1.7pt}
    Anchor-DETR~\cite{wang2021anchor}     & 50    & 85 & 39    &   20   &   \ding{55}    & 42.1/63.1/44.9  & 22.3/46.2/60.0 \\
    Anchor-DETR-DC5~\cite{wang2021anchor}      & 50    & 151 & 39    & 14   &  \ding{55}     & 44.2/64.7/47.5  & 24.7/48.2/60.6 \\
    \specialrule{0.5pt}{0.8pt}{1.7pt}
    DAB-DETR~\cite{liu2022dab}  & 50    & 90   & 44    & 17    &   \ding{55}    & 42.2/63.1/44.7  & 21.5/45.7/60.3 \\
    DAB-DETR-DC5~\cite{liu2022dab}    & 50    & 194   & 44    &   11   &   \ding{55}    & 44.5/65.1/47.7    & 25.3/48.2/62.3 \\
    \specialrule{0.5pt}{0.8pt}{1.7pt}
    SAP-DETR~\cite{liu2022sap} & 50    & 92   & 47    & 16    &   \ding{55}    & 43.1/63.8/45.4  & 22.9/47.1/62.1 \\
    SAP-DETR-DC5~\cite{liu2022sap}    & 50    & 197   & 47    &   9    &   \ding{55}    & 46.0/65.5/48.9    & 26.4/50.2/62.6 \\
    \specialrule{0.5pt}{0.8pt}{1.7pt}
    Efficient DETR~\cite{yao2021efficient}  & 36    & 159   & 32    & -     & $\checkmark$ & 44.2/62.2/48.0    & 28.4/47.5/56.6 \\
    Efficient DETR*~\cite{yao2021efficient}  & 36    & 210   & 35    & -     &   $\checkmark$    & 45.1/63.1/49.1  & 28.3/48.4/59.0 \\
    \specialrule{0.5pt}{0.8pt}{1.7pt}
    Dynamic DETR~\cite{dai2021dynamic} & 50    &    -   &  58   &   -    &   $\checkmark$    & 47.2/65.9/51.1  & 28.6/49.3/59.1 \\
    \specialrule{0.5pt}{0.8pt}{1.7pt}
    TSP-FCOS~\cite{sun2021rethinking}  & 36    & 189   & 52  & 15    & $\checkmark$ & 43.1/62.3/47.0   & 26.6/46.8/55.9 \\
    TSP-RCNN~\cite{sun2021rethinking} & 36    & 188   & 64  & 11    &   $\checkmark$    & 43.8/63.3/48.3  & 28.6/46.9/55.7 \\
    TSP-RCNN+~\cite{sun2021rethinking} & 96    & 188   & 64  & 11    &    $\checkmark$   & 45.0/64.5/49.6  & 29.7/47.7/58.0 \\
    \specialrule{0.5pt}{0.8pt}{1.7pt}
    YOLOS-S(800$\times$)~\cite{fang2021you}  & 150   & 194 & 31  & 6 &   \ding{55}   & 36.1/56.4/37.1  & 15.3/38.5/56.1 \\
    YOLOS-S(784$\times$)~\cite{fang2021you}    & 150   & 172  & 28  & 6 & $\checkmark$  & 37.6/57.6/39.2  & 15.9/40.2/57.3 \\
    YOLOS-B~\cite{fang2021you} & 150   & 538   & 127   &   3   & \ding{55}     & 42.0/62.2/44.5  & 19.5/45.3/62.1 \\
    \specialrule{0.5pt}{0.8pt}{1.7pt}
    UP-DETR~\cite{dai2021up}  & 150   & 86    & 41    & 28    &   \ding{55}    & 40.5/60.8/42.6  & 19.0/44.4/60.0 \\
    UP-DETR+~\cite{dai2021up}  & 300   & 86    & 41    & 28    &   \ding{55}    & 42.8/63.0/45.3  & 20.8/47.1/61.7 \\
    \specialrule{0.5pt}{0.8pt}{1.7pt}
    FP-DETR-Base~\cite{wang2021fp}  &   50    &   -    & 36    &    -   &   \ding{55}   & 43.3/63.9/47.7  & 27.5/46.1/57.0 \\
    \specialrule{0.5pt}{0.8pt}{1.7pt}
    DN-DETR~\cite{li2022dn}  & 50    & 94    & 44    &   17   &  \ding{55}    & 44.1/64.4/46.7  & 22.9/48.0/63.4 \\
    DN-DETR-DC5~\cite{li2022dn} & 50    & 202   & 44    &  8    &  \ding{55}     & 46.3/66.4/49.7  & 26.7/50.0/64.3 \\
    DN-Defor.-DETR~\cite{li2022dn} & 50    & 196   & 48   &  23     &  $\checkmark$     & 46.3/66.4/49.7  & 26.7/50.0/64.3 \\
    \specialrule{0.5pt}{0.8pt}{1.7pt}
    DINO-4scale~\cite{zhang2022dino}  & 36    &  279     &   47    &   24    &   $\checkmark$    & 50.5/68.3/55.1  & 32.7/53.9/64.9 \\
    DINO-5scale~\cite{zhang2022dino}   & 36    &  860     &   47    &   10   &   $\checkmark$    & 51.0/69.0/55.6  & 34.1/53.6/65.6 \\
    \specialrule{0.9pt}{1.6pt}{3.6pt}
    \multicolumn{8}{l}{\small{\makecell[l]{``MS'' denotes to multi-scale features.\\
    Both GFLOPs and Params are measured by Detectron2. FPS is measured on a single A100 GPU.}}}\\
    \bottomrule
    \end{tabular}}
  \end{spacing}
  \label{tab:02}
\end{table}%


In two-stage methods, Zhe et al. empower the Top-K region proposals from the outputs of the encoder to initialize the decoder embedding instead of the learned content query~\cite{zhu2021deformable}. Efficient DETR~\cite{yao2021efficient} also adopts a similar initialization operation for dense proposals and refines them in the decoder to get sparse prediction by using a shared detection head with the dense parts. More interestingly, it is observed that small stacking decoder layers bring slight performance improvement, but even more stacks yield even worse results. Dynamic DETR~\cite{dai2021dynamic} regards the object prediction in a coarse-to-fine process. Different from the previous RoI-based initialization detectors, a cross-attention between the pool region features and the object embeddings is used to perform decoder embedding updates. The RoI features associated with the box embedding are then refined accordingly in the next layer.

\subsubsection{Transformer with Redesigned Structure}\label{sec:0414}
Besides the modifications focusing on the cross-attention, some works redesign an encoder-only structure to avoid the problem of the decoder directly. TSP~\cite{sun2021rethinking} inherits the idea of set prediction~\cite{carion2020end} and dismisses the decoder and the object query to accelerate convergence. Such encoder-only DETR reuses previous representations~\cite{ren2017faster, tian2019fcos}, and generates a set of fixed-size Features of Interests (FoI)~\cite{tian2019fcos} or proposals~\cite{ren2017faster} that are subsequently fed into the Transformer encoder. In addition, a matching distillation is applied to resolve the instability of the bipartite matching, especially in the early training stage. Fang et al.~\cite{fang2021you} combine the encoder-decoder neck of DETR and the encoder-only backbone of ViT into an encoder-only detector and develop YOLOS, a pure sequence-to-sequence Transformer to unify the classification and detection tasks. It inherits ViT's structure and replaces the single class token with fixed size learned detection tokens. These object tokens are first pre-trained on the transfer ability for the classification tasks and then fine-tuned on the detection benchmark.

\subsubsection{Transformer with Bipartite Matched Optimization} 
In DETR~\cite{carion2020end}, the bipartite matching strategy forces the prediction results to fulfil one-to-one label assignment during the training scheme. Such a training strategy simplifies detection pipeline and directly builds up an end-to-end system without the help of NMS. To deeply understand the efficacy of the end-to-end detector, Sun et al. devote to exploring a theoretical view of one-to-one prediction~\cite{sun2021makes}. Based on multiple ablation and theoretical analyses, they conclude that the classification cost for one-to-one matching strategy serves as the key component for significantly avoiding duplicate predictions. Even so, DETR is suffering from multiple problems caused by bipartite matching. Li et al.~\cite{li2022dn} exploit a denoising DETR (DN-DETR) to mitigate the instability of bipartite matching. Concretely, a series of objects with slight perturbation is supposed  to reconstruct their actual coordinates and classes. The main ingredients of the denoising (or reconstruction) part are an attention mask that prevents information leakage between the matching and noised parts, and a specified label embedding to indicate the perturbation. Recently, Zhang et al.~\cite{zhang2022dino} present an improved denoising training model called DINO (2022) by incorporating a contrastive loss for the perturbation groups. Based on DN-DETR~\cite{li2022dn}, DINO attaches a ``no object'' class for the negative example if the distance is far enough from the perturbation, which avoids redundant prediction due to the confusion of multiple reference points near an object. As a result, DINO attains the current SoTA on the COCO dataset.

\subsubsection{Transformer Detector with Pre-Training} 
Inspired by the pre-trained linguistic Transformer~\cite{devlin2018bert,radford2019language}, Dai et al. devise an Unsupervised Pre-training DETR (UP-DETR)~\cite{dai2021up} to assist the convergence for supervised training. The objective of pre-training is to localize the random cropped patches from a given image. Specifically, each patch is assigned to a set of queries and predicted independently via the attention mask. An auxiliary reconstruction loss forces the detector to preserve the feature discrimination so as to avoid over-bias towards the localization in pre-training. FP-DETR~\cite{wang2021fp} devotes to narrowing the gap between upstream and downstream tasks. During the pre-training, a fully encoder-only DETR like YOLOS~\cite{fang2021you} views query positional embeddings as a visual prompt to enhance target area attention and object discrimination. A task adapter implemented by self-attention is used to enhance object interaction during fine-tuning. 

\subsection{Transformer Backbone}\label{sec:042}
We have reviewed numerous Transformer-based backbones for image classification~\cite{dosovitskiy2021an,touvron2021training} in Sec.~\ref{sec:03}. These backbones can be easily incorporated into various frameworks (e.g., Mask R-CNN~\cite{he2017mask}, RetinaNet~\cite{lin2017focal}, DETR~\cite{carion2020end}, etc.) to perform dense prediction tasks. For example, the hierarchical structure like PVT~\cite{wang2021pyramid,wang2021pvtv2}, constructs the visual Transformer as a high-to-low resolution process to learn multi-scale features. The locally enhanced structure constructs the backbone as a local-to-global combination, which can efficiently extract both short-range and long-range visual dependencies and avoid quadratic computational overhead, such as Swin-Transformer~\cite{liu2021swin}, ViL~\cite{zhang2021multi}, and Focal Transformer~\cite{yang2021focal}. App.~\ref{sup:comparison} includes more detailed comparisons of these models for the dense prediction tasks. In addition to the generic Transformer backbone, the Feature Pyramid Transformer (FPT)~\cite{zhang2020feature} combines the characteristics across both the spaces and the scales, by using self-attention, top-down cross-attention, and bottom-up cross channel attention. Following~\cite{sun2019deep}, HRFormer~\cite{yuan2021hrformer} introduces the advantages of multi-resolution to the Transformer along with non-overlapping local self-attention. HRViT~\cite{gu2021hrvit} redesigns a heterogeneous branch and a cross-shaped attention block to further optimize the trade-off between efficiency and accuracy.

\subsection{Discussion}\label{sec:043}
We summarize five folds of the Transformer neck detectors in Tab.~\ref{tab:02}, and more details of Transformer backbone for dense prediction tasks are referred to in Tab.~\ref{tab:s1}. The majority of Transformer neck promotions concentrate on the following five aspects: 1) The sparse attention model and the scoring network are proposed to address the problem of redundant feature interaction. These methods can significantly alleviate computational costs and accelerate model convergence. 2) The explicit spatial prior, which is decomposed into the selected feature initialization and the positional information extracted by learned parameters, would enable the detector to predict the results precisely. 3) Multi-scale features and layer-by-layer updating are extended in the Transformer decoder for small object refinement. 4) The improved bipartite matching strategy is beneficial to avoid redundant prediction as well as perform end-to-end object detection. 5) The encoder-only structure reduces the overall Transformer stack layers but increases the FLOPs excessively, while the encoder-decoder structure is a good trade-off between FLOPs and Parameters, but the deeper decoder layers may cause the problems of long training time and over-smooth.

Moreover, there are many Transformer backbones for improving classification performance, but few works are developed for the dense prediction tasks. In the future, we anticipate that the Transformer backbone would cooperate with the deep high-resolution network to solve dense prediction tasks.

\section{Transformer for Segmentation}\label{sec:05}
\textit{Patch-Based} and \textit{Query-Based Transformer} are the two major ways for Segmentation. The latter can be further grouped into \textit{Object Query} and \textit{Mask Embedding} methods.

\begin{figure}[!htbp]
    \centering
    \includegraphics[width=3.2in]{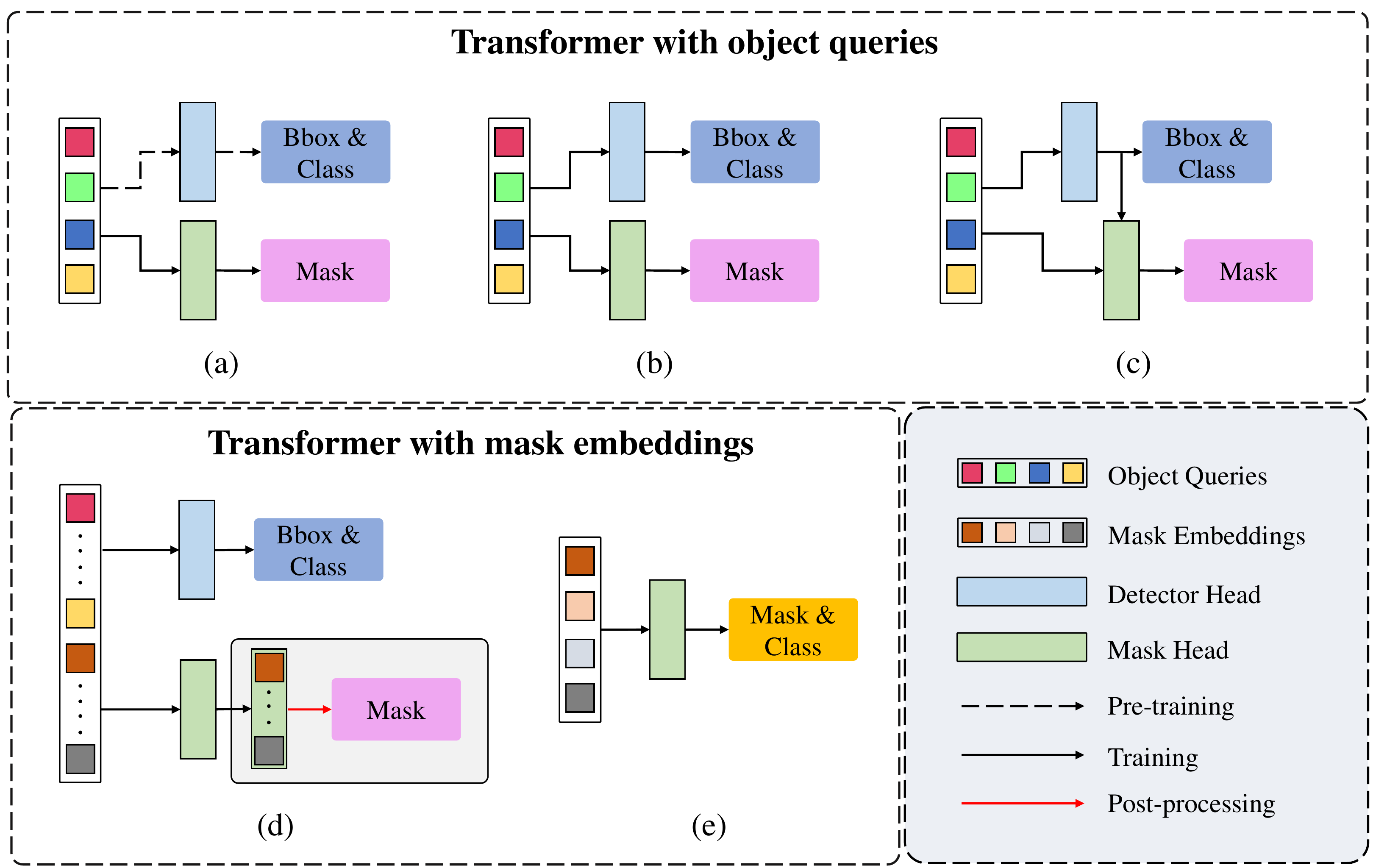} 
    \caption{Query-based frameworks for segmentation tasks. (a) Transfer learning for fine-tuning mask head.  (b) Multi-task learning for two independent task. (c) Cascade learning for fine-grained mask generation on the coarse region prediction. (d) The query embeddings are independently supervised by mask embeddings and boxes. (e) The box-free model directly predicts masks without box branch and views segmentation task as a mask prediction problem.}
    \label{fig:10}
\end{figure}

\subsection{Patch-Based Transformer}\label{sec:051}
Because of the receptive field expansion strategy~\cite{chen2017deeplab}, CNNs require multiple decoder stacks to map the high-level features into the original spatial resolution. Instead, patch-based Transformer can easily incorporate with a simple decoder for segmented mask prediction because of its global modelling capability and resolution invariance. Zheng et al. extend ViT~\cite{dosovitskiy2021an} for semantic segmentation tasks, and present SEgmentation TRansformer (SETR)~\cite{zheng2021rethinking} by employing three fashions of the decoder to perform per-pixel classification: naive up-sampling, progressive up-sampling, and multi-level feature aggregation (MLA). SETR demonstrates the feasibility of the visual Transformer for the segmentation tasks, but it also brings unacceptably extra GPU costs. TransUNet~\cite{chen2021transunet} is the first for medical image segmentation. Formally, it can be viewed as either a variant of SETR with MLA decoder~\cite{zheng2021rethinking}, or a hybrid model of U-Net~\cite{ronneberger2015u} and Transformer. Thanks to the strong global modeling capability of Transformer encoder, Segformer~\cite{xie2021segformer} designs a lightweight decoder with only four MLP layers. Segformer shows superior performance as well as stronger robustness than CNNs when tested with multiple corrupted types of images.

\subsection{Query-Based Transformer}\label{sec:052}
Query embeddings are a set of scratch semantic/instance representations gradually learning from the image inputs. Unlike patch embeddings, queries can more ``fairly'' integrate the information from features and naturally join with the set prediction loss~\cite{carion2020end} for post-processing elimination. Existed query-based models can be grouped into two categories. One is driven by both the tasks of detection and segmentation, simultaneously (called \textit{object queries}). The other is only supervised by the segmentation task (called \textit{mask embeddings}).

\subsubsection{Object Queries}\label{sec:0521}
There are three training manners for object query based methods (Fig.~\ref{fig:10}(a)-(c)). With the success of DETR~\cite{carion2020end} for the object detection tasks, the authors extend it to panoptic segmentation (hence termed Panoptic DETR~\cite{carion2020end}) by training a mask head based on the pre-trained object queries (Fig.~\ref{fig:10}(a)). In detail, a cross-attention block between the object queries and the encoded features is applied to generate an attention map for each object. 
\begin{figure}[htbp]
    \centering
    \includegraphics[width=3.5in]{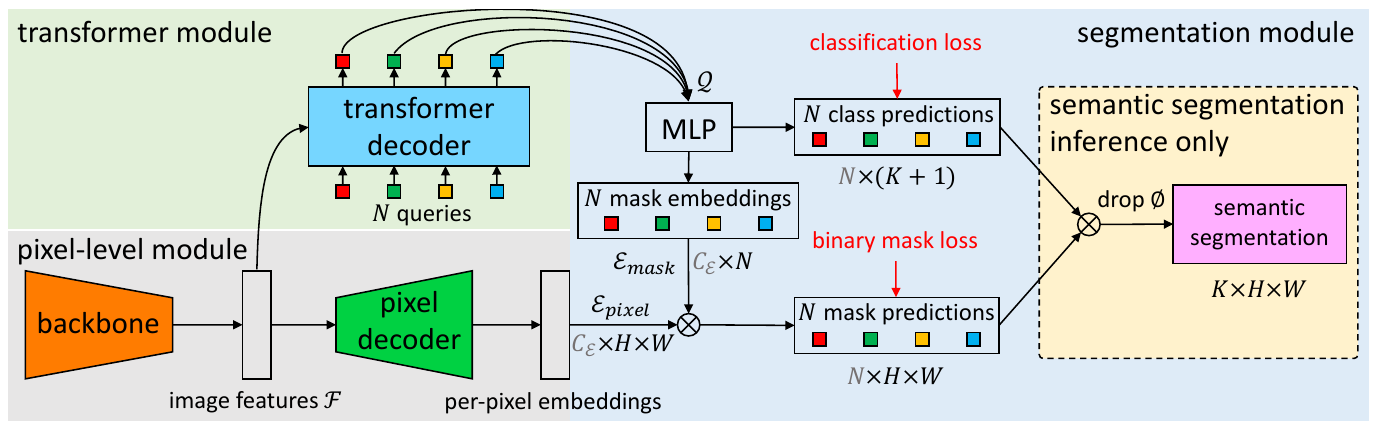}
    \caption{Illustration of Maskformer. (from~\cite{cheng2021per}.)}
    \label{fig:11}
\end{figure}
After an up-sampling FPN-style CNN, a spatial argmax operation fuses the resulting binary masks to a non-overlapping prediction. Instead of using a multi-stage serial training process, Cell-DETR and VisTR develop a parallel model for end-to-end instance segmentation (Fig.~\ref{fig:10}(b)). Based on DETR~\cite{carion2020end}, Cell-DETR leverages a cross-attention block to extract instance-wise features from the box branch and fuses the previous backbone features to augment the CNN decoder for accurate instance mask segmentation of biological cells. Another extension is VisTR~\cite{wang2021end} that directly formulates the video instance segmentation (VIS) task as parallel sequence prediction. Apart from the similar structure as Cell-DETR~\cite{prangemeier2020attention}, the key of VisTR is a bipartite matching loss at the instance sequence level to maintain the order of outputs, so as to adapt DETR~\cite{carion2020end} to VIS for direct one-to-one predictions. Unlike prior works that treat detection and mask generation branches separately, QueryInst~\cite{fang2021queryinst} builds a hybrid cascaded network (Fig.~\ref{fig:10}(c)), where the previous box outputs together with the shared queries serve as the inputs of the mask head for accurate mask segmentation. Notably, QueryInst leverages the shared queries to keep the instance correspondences across multi-stage, so that mitigating the problem of inconsistent objects in previous non-query based methods~\cite{cai2018cascade,sun2021sparse}. QueryInst obtains the latest SoTA results on the COCO datasets.

\subsubsection{Mask Embeddings} \label{sec:0522}
The other framework makes efforts to use queries to predict mask directly, and we refer to this learned mask-based query as mask embeddings. Unlike the object queries, mask embeddings are only supervised by the segmentation tasks. As shown in Fig.~\ref{fig:10}(d), two disjoint sets of queries are employed parallelly for different tasks, and the box learning is viewed as an auxiliary loss for further enhancement. For semantic and box-free instance segmentation, a series of query-based Transformers predict the mask directly without the help of the box branch (Fig.~\ref{fig:10}(e)).

From the auxiliary training perspective, the core is how to enable 1D sequence outputs to be supervised by 2D mask labels directly. To this end, ISTR~\cite{hu2021istr} empowers a mask pre-coding method to encode the ground-truth spatial mask into low-dimensional mask embedding for instance segmentation. Similarly, Dong et al. propose a more straightforward pipeline SOLQ~\cite{dong2021solq} and explore three reversible compression encoding methods for mask embeddings. In detail, a set of unified queries is applied to perform multiple representation learning parallelly: classification, box regression, and mask encoding. Based on the original DETR~\cite{chen2017deeplab}, SOLQ adds a mask branch to produce mask embedding loss. Both ISTR and SOLQ obtain comparable results and outperform previous methods even with approximation-based suboptimal embeddings. However, there exists a huge gap between $AP^{box}$ and $AP^{seg}$ (Tab.~\ref{tab:03}).

From the box-free perspective, Wang et al. pioneer a new paradigm Max-DeepLab~\cite{wang2021max} that directly predicts panoptic masks from the query without the help of the box branch. Specifically, it forces the query to predict the corresponding mask via a PQ-style bipartite matching loss and a dual-path Transformer structure. Given a set of mask embeddings and an image input, Max-DeepLab processes them separately in both Transformer and CNN path, and then generates a binary mask and a class for each query, respectively. Max-DeepLab achieves new SoTA with 51.3\% PQ on COCO test-dev set, but leads to heavy computational costs due to its dual-path high-resolution processing. Segmenter~\cite{strudel2021segmenter} views the semantic segmentation task as a sequence-to-sequence problem. In detail, a set of mask embeddings that represent different semantic classes are fed into the Transformer encoder together with image patches, and then a set of labeled masks are predicted for each patch via an argmax operation.

Unlike the conventional semantic segmentation methods that predict mask at the pixel level, Cheng et al. reformulate the semantic segmentation task as a mask prediction problem and enable this output format to the query-based Transformer, which is called Maskformer~\cite{cheng2021per}. Different from Max-DeepLab~\cite{wang2021max}, Maskformer leverages a simple Transformer decoder without redundant connection as well as a sigmoid activation for overlapping binary masks selection. It not only outperforms the current per-pixel classification SoTA on large-class semantic segmentation datasets, but also generalizes the panoptic segmentation task with a new SoTA result (Tab.~\ref{tab:03}).

\subsection{Discussion}\label{sec:053}
We summarize the aforementioned Transformers according to three different tasks. Table~\ref{tab:03}(a) focuses on ADE20K (170 classes). It can be shown that when trained on the datasets with large numbers of classes, the segmentation performance of visual Transformers is improved significantly. Table~\ref{tab:03}(b) focuses on COCO test dataset for instance segmentation. Clearly, the visual Transformers with mask embeddings surpass most prevailing models for both segmentation and detection tasks. However, there is a huge performance gap between $AP^{box}$ and $AP^{seg}$. With the cascaded framework, QueryInst~\cite{fang2021queryinst} attains the SoTA among various Transformer models. It is worthy of further study for combining the visual Transformers with the hybrid task cascade structures. Table~\ref{tab:03}(c) focuses on panoptic segmentation. Max-DeepLab~\cite{wang2021max} is general to solve both foreground and background in the panoptic segmentation task via a mask prediction format, while Maskformer~\cite{cheng2021per} successfully employs this format for semantic segmentation and unifies both semantic and instance segmentation tasks into a single model. Based on their performances in the panoptic segmentation field, we can conclude that the visual Transformers could unify multiple segmentation tasks into one box-free framework with mask prediction.

\begin{table}[htbp]
  \centering
  \caption{Comparison between CNN-based and Transformer-based model on ADE20K and COCO for different segmentation tasks. ``+MS'' denotes the performance trained with multi-scale inputs.}
  \setlength \tabcolsep{4pt}
  \begin{spacing}{0.58}
  \resizebox{\linewidth}{!}{%
    \begin{tabular}{lccccccc}
    \specialrule{0.5pt}{0.5pt}{0.3pt}
    \specialrule{0.5pt}{0.3pt}{1.7pt}
    \multicolumn{8}{l}{\textit{(a) ADE20K Val. Set for Semantic Segmentation}}  \\
    \specialrule{0.5pt}{0.7pt}{0.3pt}
    \specialrule{0.5pt}{0.3pt}{1.2pt}
    
    \textbf{Method} & \multicolumn{1}{c}{\textbf{Backbone}}   & \textbf{\makecell[c]{image \\ size}} & \textbf{\makecell[c]{\#Params. \\ (M)}} & \textbf{\makecell[c]{FLOPs \\ (G)}} & \textbf{FPS} & \textbf{mIoU}  & \textbf{+MS} \\
    \toprule
    \multirow{6}[0]{*}{UperNet\makecell[c]{ \cite{cheng2021per} \\ \cite{xiao2018unified} \\ \cite{chen2021searching} }} & R-50~\cite{he2016deep}   & $\text{512}$   & 67    & 238     & 23.4& 42.1  & 42.8 \\
          & R-101~\cite{he2016deep}    & $\text{512}$   & 86    & 257  & 20.3  & 43.8  & 44.9 \\ 
          & Swin-T~\cite{liu2021swin}   & $\text{512}$   & 60    & 236 & 18.5  & 44.5  & 46.1 \\ 
          & Swin-S~\cite{liu2021swin}   & $\text{512}$   & 81    & 259 & 15.2  & 47.6  & 49.3 \\ 
          & Swin-B$^\dagger$~\cite{liu2021swin}   & $\text{640}$  & 121   & 471 & 8.7  & 50.0    & 51.6 \\ 
          & Swin-L$^\dagger$~\cite{liu2021swin}   & $\text{640}$   & 234   & 647 & 6.2  & 52.0    & 53.5 \\ 
    \specialrule{0.5pt}{0.8pt}{1.7pt}
    \multirow{3}[0]{*}{Segformer~\cite{xie2021segformer}} & MiT-B3   & $\text{512}$   & 47  & 79 & -  & 49.4  & 50.0 \\
          & MiT-B4   & $\text{512}$   & 64  & 96 & 15.4  & 50.3  & 51.1 \\
          & MiT-B5   & $\text{640}$  & 85  & 183 & 9.8  & 51.0    & 51.8 \\
    \specialrule{0.5pt}{0.8pt}{1.7pt}
    \multirow{3}[0]{*}{Segmenter~\cite{strudel2021segmenter}} & ViT-S/16$^\dagger$~\cite{dosovitskiy2021an}   & $\text{512}$   & 27    &   -    & 34.8 & 45.3  & 46.9\\
          & ViT-B/16$^\dagger$~\cite{dosovitskiy2021an}   & $\text{512}$   & 106   &   -    & 24.1 & 48.5  & 50.0 \\
          & ViT-L/16$^\dagger$~\cite{dosovitskiy2021an}   & $\text{640}$    & 334   &   -    & - & 51.8  & 53.6\\
    \specialrule{0.5pt}{0.8pt}{1.7pt}
    \multirow{7}[0]{*}{MaskFormer~\cite{cheng2021per}} & R-50~\cite{he2016deep}  & $\text{512}$   & 41    & 53    & 24.5 & 44.5  & 46.7 \\
          & R-101~\cite{he2016deep}   & $\text{512}$   & 60    & 73    & 19.5 & 45.5  & 47.2\\
          & Swin-T~\cite{liu2021swin}  & $\text{512}$   & 42    & 55    & 22.1 & 46.7  & 48.8\\
          & Swin-S~\cite{liu2021swin}  & $\text{512}$   & 63    & 79    & 19.6 & 49.8  & 51.0 \\
          & Swin-B$^\dagger$~\cite{liu2021swin}  & $\text{640}$    & 102   & 195   & 12.6 & 52.7  & 53.9\\
          & Swin-L$^\dagger$~\cite{liu2021swin}  & $\text{640}$    & 212   & 375   & 7.9 & 54.1  & 55.6\\

    \specialrule{0.5pt}{0.5pt}{0.3pt}
    \specialrule{0.5pt}{0.3pt}{1.7pt}
    \multicolumn{8}{l}{\textit{(b): COCO Test-Dev for Instance Segmentation}} \\
    \specialrule{0.5pt}{0.7pt}{0.3pt}
    \specialrule{0.5pt}{0.3pt}{1.2pt}
    
    \textbf{Method} & \textbf{Backbone} & \textbf{Epochs} & $\textbf{AP}^{box}$/$\textbf{AP}^{seg}$   & $\textbf{AP}_S^{seg}$ & $\textbf{Ap}_M^{seg}$ & $\textbf{Ap}_L^{seg}$    & \textbf{FPS} \\
    \toprule
    \multirow{2}[0]{*}{\makecell[c]{Mask R-CNN}~\cite{he2017mask}} & \multicolumn{1}{c}{R-50-FPN~\cite{he2016deep}} & 36    & 41.3/37.5  & 21.1  & 39.6  & 48.3    & 15.3 \\
          & \multicolumn{1}{c}{R-101-FPN~\cite{he2016deep}} & 36    & 43.1/38.8  & 21.8  & 41.4  & 50.5    & 11.8 \\
    \specialrule{0.5pt}{0.8pt}{1.7pt}
    \multirow{2}[0]{*}{\makecell[c]{Blend Mask}~\cite{chen2020blendmask}} & \multicolumn{1}{c}{R-50-FPN~\cite{he2016deep}} & 36    & 43.0/37.8  & 18.8  & 40.9  & 53.6    & 15.0 \\
          & \multicolumn{1}{c}{R-101-FPN} & 36    & 44.7/39.6  & 22.4  & 42.2  & 51.4   & 11.5 \\
    \specialrule{0.5pt}{0.8pt}{1.7pt}
    \multirow{2}[0]{*}{\makecell[c]{SOLO v2}~\cite{wang2020solov2}} & \multicolumn{1}{c}{R-50-FPN~\cite{he2016deep}} & 36    & 40.7/38.2  & 16.0    & 41.2  & 55.4    & 10.5 \\
          & \multicolumn{1}{c}{R-101-FPN~\cite{he2016deep}} & 36    & 42.6/39.7  & 17.3  & 42.9  & 57.4    & 9.0 \\
    \specialrule{0.5pt}{0.8pt}{1.7pt}
    \multirow{2}[0]{*}{ISTR~\cite{hu2021istr}} & \multicolumn{1}{c}{R-50-FPN~\cite{he2016deep}} & 36    & 46.8/38.6  & 22.1  & 40.4  & 50.6    & 13.8 \\
          & \multicolumn{1}{c}{R-101-FPN~\cite{he2016deep}} & 36    & 48.1/39.9  & 22.8  & 41.9  & 52.3   & 11.0 \\
    \specialrule{0.5pt}{0.8pt}{1.7pt}
    \multirow{3}[0]{*}{SOLQ~\cite{dong2021solq}} & \multicolumn{1}{c}{R-50~\cite{he2016deep}} & 50    & 47.8/39.7  & 21.5  & 42.5  & 53.1  & - \\
          & \multicolumn{1}{c}{R-101~\cite{he2016deep}} & 50    & 48.7/40.9  & 22.5  & 43.8  & 54.6    & - \\
          & \multicolumn{1}{c}{Swin-L$^\dagger$~\cite{liu2021swin}} & 50   & 55.4/45.9  & 27.8  & 49.3  & 60.5    & - \\
    \specialrule{0.5pt}{0.8pt}{1.7pt}
    \multirow{4}[1]{*}{\makecell[c]{QueryInst}~\cite{fang2021queryinst}} & \makecell[c]{R-50-FPN~\cite{he2016deep}} & 36     & 44.8/40.1  & 23.3  & 42.1  & 52.0     & 10.5 \\
          & R-50-FPN~\cite{he2016deep} & 36    & 45.6/40.6  & 23.4  & 42.5  & 52.8    & 7.0 \\
          & R-101-FPN~\cite{he2016deep} & 36   & 47.0/41.7  & 24.2  & 43.9  & 53.9      & 6.1 \\
          & Swin-L$^\dagger$~\cite{liu2021swin} & 50      & 56.1/49.1  & 31.5  & 51.8  & 63.2    & 3.3 \\

    \specialrule{0.5pt}{0.5pt}{0.3pt}
    \specialrule{0.5pt}{0.3pt}{1.7pt}
    \multicolumn{8}{l}{\textit{(c): COCO Panopticon Minival. for Panoptic Segmentation}} \\
    \specialrule{0.5pt}{0.7pt}{0.3pt}
    \specialrule{0.5pt}{0.3pt}{1.2pt}
    
    \textbf{Method} & \textbf{Backbone} & \textbf{Epochs} & \textbf{\makecell[c]{\#Params. \\ (M)}} & \textbf{\makecell[c]{FLOPs \\ (G)}} & \textbf{PQ} & $\textbf{PQ}^{Th}$ & $\textbf{PQ}^{St}$  \\
    \toprule
    \multirow{2}[0]{*}{DETR~\cite{carion2020end}} & R-50~\cite{he2016deep} & \multirow{2}[0]{*}{500+25}   & 43  & 137 & 43.4  & 48.2  & 36.3 \\
          & R-101~\cite{he2016deep} &      & 62  & 157 & 45.1  & 50.5  & 37.0     \\
    \specialrule{0.5pt}{0.8pt}{1.7pt}
    \multirow{2}[0]{*}{MaxDeepLab~\cite{wang2021max}} & Max-S & \multirow{2}[0]{*}{54}  & 62  & 162 & 48.4  & 53.0    & 41.5   \\
          & Max-L &       & 451   & 1846 & 57.0    & 42.2 & 51.1    \\
    \specialrule{0.5pt}{0.8pt}{1.7pt}
    \multirow{6}[0]{*}{MaskFormer~\cite{wang2021not}} & R-50~\cite{he2016deep} & \multirow{6}[0]{*}{300} & 45    & 181 & 46.5  & 51.0    & 39.8   \\
          & R-101~\cite{he2016deep} &       & 64    & 248 & 47.6  & 52.5  & 40.3   \\
          & Swin-T~\cite{liu2021swin} &       & 42    & 179 & 47.7  & 51.7  & 41.7   \\
          & Swin-S~\cite{liu2021swin} &       & 63    & 259 & 49.7  & 54.4  & 42.6   \\
          & Swin-B~\cite{liu2021swin} &       & 102   & 411 & 51.1  & 56.3  & 43.2   \\
          & Swin-L$^\dagger$~\cite{liu2021swin} &       & 212   & 792 & 52.7  & 58.5  & 44.0     \\
         
    \specialrule{0.5pt}{0.5pt}{0.3pt}
    \specialrule{0.5pt}{0.3pt}{1.2pt}
    \multicolumn{8}{l}{\small{$\dagger$ denotes the model pre-trained on ImageNet-21k}}  \\
    \specialrule{0.5pt}{0.5pt}{0.3pt}
    \specialrule{0.5pt}{0.3pt}{1.2pt}
    \end{tabular}}
  \end{spacing}
  \label{tab:03}
\end{table}

\section{Transformer for 3D Visual Recognition}\label{sec:add01}
With the rapid development of 3D acquisition technology, stereo/monocular images and LiDAR (Light Detection And Ranging) point clouds become the popular sensory data for 3D recognition. Discriminated from the RGB(D) data, point cloud representation pays more attention to distance, geometry, and shape information. Notably, such a geometric feature is significantly suitable for Transformer on account of its characteristic on sparseness, disorder, and irregularity. Following the success of 2D visual Transformers, substantial approaches are developed for 3D visual analysis. This section exhibits a compact review for 3D visual Transformers following \textit{Representation learning}, \textit{Cognition mapping}, and \textit{Specific processing}.

\subsection{Representation Learning}\label{sec:add011}
Compared with conventional hand-designed networks, visual Transformer is more appropriate for learning semantic representations from point clouds, in which such irregular and permutation invariant nature can be transformed into a series of parallel embeddings with positional information. In view of this, Point Transformer~\cite{zhao2021point} and PCT~\cite{guo2021pct} firstly demonstrate the efficacy of the visual Transformer for 3D representation learning. The former merges a hierarchical Transformer~\cite{zhao2021point} with the down-sampling strategy~\cite{qi2017pointnet++} and extends their previous vector attention block~\cite{zhao2020exploring} to 3D point clouds. The latter first aggregates neighbour points and then processes such neighbour embeddings on a global off-set Transformer where a knowledge transfer from Graph Convolution Network (GCN) is applied for noise mitigation. Notably, the positional encoding, a significant operation of the visual Transformer, is diminished in both the approaches because of points' inherent coordinate information. PCT directly processes on the coordinates without positional encodings, while Point Transformer adds a learnable relative positional encoding for further enhancement. Lu et al. leverage a local-global aggregation module 3DCTN~\cite{lu20223dctn} to achieve local enhancement and cost-efficiency. Given the multi-stride down-sampling groups, an explicit graph convolution with max-pooling operation are used to aggregate the local information within each group. The resulting group embeddings are concatenated and fed into the improved transformer~\cite{zhao2021point,guo2021pct} for global aggregation. Park et al. present Fast Point Transformer~\cite{park2021fast} to optimize the model efficiency by using voxel-hashing neighbor search, voxel-bridged relative positional encoding, and cosine similarity based local attention. 

For dense prediction, Pan et al. propose a customized point-based backbone Pointformer~\cite{pan20213d} for attending the local and global interactions separately within each layer. Different from previous local-global forms, a coordinate refinement operation after the local attention is adopted to update the centroid point instead of the surface one. And a local-global cross attention model fuses the high-resolution features, followed by global attention. Fan et al. return to a Single-stride Sparse Transformer (SST)~\cite{fan2021embracing} rather than the down-sampling operation to address the problem for small scale detection. Similar to Swin~\cite{liu2021swin}, a shifted group in continuous Transformer block is adopted to attend to each group of tokens separately, which further mitigates the computation problem. In voxel-based methods, Voxel Transformer (VoTr)~\cite{mao2021Voxeltf} separately operate on the empty and non-empty voxel positions effectively via local attention and dilated attention blocks. VoxSeT~\cite{he2022Voxelst} further decomposes the  self-attention layer into two cross-attention layers, and a group of latent codes link them to preserve global features in a hidden space. 

Following the mentioned methods in Sec.~\ref{sec:037}, a series of self-supervised Transformers are also extended to 3D spaces~\cite{yu2021point,pang2022masked,Liu2022MaskedDF}. Specifically, Point-BERT~\cite{yu2021point} and Point-MAE~\cite{pang2022masked} directly transfer the previous works~\cite{bao2021beit,he2021mae} to point clouds, while MaskPoint~\cite{Liu2022MaskedDF} changes the generative training scheme by using a contrastive decoder as similar as DINO (2022)~\cite{zhang2022dino} for binary noise/part classification. Based on large experiments, we can conclude that such generative/contrastive self-training methods empower visual Transformers to be valid in either images or points.

\subsection{Cognition Mapping}\label{sec:add012}
Given rich representation features, how to directly map the instance/semantic cognition to the target outputs also arouse considerable interests. Different from 2D images, the objects in 3D scenes are independent and can be intuitively represented by a series of discrete surface points. To bridge the gap, some existed methods transfer domain knowledge into 2D prevailing models. Following~\cite{carion2020end}, 3DETR~\cite{misra2021end} extends an end-to-end module for 3D object detection via farthest point sampling and Fourier positional embeddings for object queries initialization. Group-Free 3D DETR~\cite{liu2021group} applies a more specified and stronger structure than~\cite{misra2021end}. In detail, it directly selects a set of candidate sample points from the extracted point clouds as the object queries and updates them in the decoder layer-by-layer iteratively. Moreover, the $K$-closed inside points are assigned positive and supervised by a binary objectiveness loss in both sampler and decoder heads. Sheng et al. proposes a typical two-stage method that leverages a Channel-wise Transformer 3D Detector (CT3D)~\cite{sheng2021improving} to simultaneously aggregate proposal-aware embedding and channel-wise context information for the point features within each proposal.

For monocular sensors, both MonoDTR~\cite{huang2022mono} and MonoDETR~\cite{zhang2022mono} utilize an auxiliary depth supervision to estimate pseudo Depth Positional Encodings (DPEs) during the training process. In MonoDETR~\cite{huang2022mono}, DPEs are first attached with the image features for Transformer encoder and then serve as the inputs of the DETR-like~\cite{carion2020end} decoder to initialize the object queries. In MonoDETR~\cite{zhang2022mono}, both visual features and DPEs are first extracted by two different encoders parallelly and then interact with object queries via two successive cross-attention layers. Based on foreground depth supervision and narrow categorisation interval, MonoDETR obtains the SoTA result on the KITTI benchmark. DETR3D~\cite{wang2022detr3d} introduces a multi-camera 3D object detection paradigm where both 2D images and 3D positions are associated by the camera transformation matrices and a set of 3D object queries. TransFusion~\cite{bai2022transfusion} further takes the advantages of both LiDAR points and RGB images by interacting with object queries through two Transformer decoder layers successively. More multi-sensory data fusion are introduced in Sec.~\ref{sec:add021}. 

\subsection{Specific Processing}\label{sec:add013}
Limited by sensor resolution and view angle, point clouds are afflicted with incompletion, noise, and sparsity problems in real-world scenes. To this end, PoinTr~\cite{yu2021pointr} represents the original point cloud as a set of local point proxies and leverages a geometry-aware encoder-decoder Transformer to migrate the centre point proxies towards incomplete points direction. SnowflakeNet~\cite{xiang2021snow} formulates the process of completing point clouds as a snowflake-like growth, which progressively generates child points from their parent points implemented by a point-wise splitting deconvolution strategy. A skip-Transformer for adjacent layers further refines the spatial-context features between parents and children to enhance their connection regions. Choe et al. unify various generation tasks (e.g. denosing, completing and super-resolution) into a Point cloud Reconstruction problem, hence termed PointRecon~\cite{choe2021deep}. Based on voxel hashing, it covers the absolute-scale local geometry and utilizes a PointTransformer-like~\cite{zhao2021point} structure to aggregate each voxel (the query) with its neighbours (the value-key pair) for fine-grained conversion from the discrete voxel to a group of point sets. Moreover, an amplified positional encoding is adapted to the voxel local attention scheme, implemented by using a negative exponential function with L1-loss as weights for vanilla positional encodings. Notably, compared with masked generative self-training, the completion task directly generates a set of complete points without the explicit spatial prior of incomplete points.

\section{Transformer for Multi-Sensory Data Stream}\label{sec:add02}
In the real world, multiple sensors are always used complementarily rather than a single one. To this end, recent works start to explore different fusing methods to cooperate multi-sensory data stream effectively. Compared with the typical CNNs, Transformer is naturally appropriate for multi-stream data fusion because of its nonspecific embedding and dynamically interactive attention mechanism. This section details these methods according to their data stream sources: \textit{Homologous Stream} and \textit{Heterologous Stream}.

\subsection{Homologous Stream}\label{sec:add021}
Homologous stream is a set of multi-sensory data with similar inherent characteristics, such as multi-view, multi-dimension, and multi-modality visual stream data. They can be categorized into two groups: \textit{Interactive Fusion} and \textit{Transfer Fusion}, according to their fusion mechanism.

\subsubsection{Interactive Fusion}
The classical fusion pattern of CNN adopts a channel concatenation operation. However, the same positions from different modalities might be anisotropic, which is unsuitable for the translation-invariant bias of CNN. Instead, the spatial concatenation operation of Transformer enables different modalities to interact beyond the local restriction. 

For the local interaction, MVT~\cite{chen2021mvt} spatially concatenates the patch embeddings from different views and strengthens their interaction via a modal-agnostic Transformer encoder. Considering the redundant features from different modalities, MVDeTr~\cite{hou2021multiview} projects each view of features onto the ground plane and extends the multi-scale deformable attention~\cite{zhu2021deformable} to a multi-view design. TransFuser~\cite{prakash2021multi}, COTR~\cite{jiang2021cotr}, 
and mmFormer~\cite{zhang2022mmformer} deploy a hybrid model. TransFuser models image and LiDAR inputs separately by using two different convolution backbones and links the intermediate feature maps via a Transformer encoder together with a residual connection. COTR shares the CNN backbone for each of view images and inputs the resulted features into a Transformer encoder block with a spatially expanded mesh-grid positional encoding. mmFormer exploits a modality-specific Transformer encoder for each sequence of MRI image and a modality-correlated Transformer encoder for multi-modal modeling.

For the global interaction, Wang et al.~\cite{wang2021multi} leverage a shared backbone to extract the features for different views. Instead of pixel/patch wise concatenation in COTR~\cite{jiang2021cotr}, the extracted view-wise global features are spatially concatenated to perform view fusion within a Transformer. Considering the angular and position discrepancy across different camera views, TransformerFusion~\cite{bozic2021transformerfusion} first converts each view feature into an embedding vector with the intrinsics and extrinsics of their camera views. These embeddings are then fed into a global Transformer whose attention weights are used for a frame selection so as to compute efficiently. To unify the multi-sensory data in 3D detection, FUTR3D~\cite{chen2022futr3d} projects the object queries in DETR-like decoder into a set of 3D reference points. These points together with 
their related features are subsequently sampled from different modalities and spatially concatenated to update the object queries.

\subsubsection{Transfer Fusion}
Unlike the interactive fusion implemented by the Transformer encoder with self-attention, the other fusing form is more like a transfer learning from the source data to the target one via a cross-attention mechanism. For instance, Tulder et al.~\cite{tulder2021multi} insert two cooperative cross-attention Transformers into the intermediate backbone features for bridging the unregistered multi-view medical images. Instead of the pixel-wise attention form, a token-pixel cross-attention is further developed to alleviate arduous computation. Long et al.~\cite{long2021multi} propose a epipolar spatio-temporal Transformer for multi-view image depth estimation. Given a single video containing a series of static multi-view frames, the neighbour frames are first concatenated and the epipolar is then warped into the centre camera space. The resulted frame volume finally serves as the source data to perform fusion with the centre frame through a cross-attention block. With the spatially-aligned data streams, DRT~\cite{song2021deep} first explicitly models the relation map between different data streams by using a convolution layer. The resulting maps are subsequently fed into a dual-path cross-attention to build both local and global relationships parallelly, thereby it can collect more regional information for glaucoma diagnosis. 

\subsection{Heterologous Stream}\label{sec:add022}
Visual Transformers also perform excellently on heterologous data fusion, especially in visual-linguistic representation learning. Although different tasks may adopt different training schemes, such as supervised/self-supervised learning or compact/large-scale datasets, we here categorize them into two representative groups only according to their cognitive forms: 1) \textit{Visual-Linguistic Pre-Training} including Vision-Language Pre-training (VLP) and Contrastive Language-Image Pre-training (CLIP), 2) and \textit{Visual Grounding} such as Phrase Grounding (PG), Referring Expression Comprehension (REC). More comparisons please see Tab.~\ref{tab:s2}.

\subsubsection{Visual-Linguistic Pre-Training} \label{sec:add0221}
Due to limited annotated data, early VLP methods commonly rely on off-the-shelf object detector~\cite{anderson2018bottom} and text encoder~\cite{devlin2018bert} to extract data-specific features for joint distribution learning. Given an image-text pair, an object detector pre-trained on Visual Genome (VG)~\cite{krishna2017vg} first extracts a set of object-centric RoI features from the image. The RoI features serving as visual tokens are then merged with text embeddings for pre-defined tasks pre-training. Basically, these methods are grouped into dual-stream and single-stream fusion.

The dual-stream methods, including ViLBERT~\cite{lu2019vilbert} and LXMERT~\cite{tan2019lxmert}, apply a vision-language cross-attention layer between two data-specific frameworks for multi-modal transferring fusion. Concretely, ViLBERT~\cite{lu2019vilbert} is pre-trained through Masked Language Modeling (MLM), Masked Region Classification (MRC), and Image Text Alignment (ITA) on Conceptual Captions (CC)~\cite{sharma2018cc3m} with 3M image-text pairs. LXMERT~\cite{tan2019lxmert} extends the pre-training datasets to a large-scale combination and further indicates that the pre-trained task-specific (BERT~\cite{devlin2018bert}) weights initialization is harmful to the pre-training of multi-sensory data fusion. 

VideoBERT~\cite{sun2019videobert} is the first single-stream VLP method, which clusters latent space features of each video frame as visual tokens and organizes their corresponding text embeddings by using a captioning API. Subsequently, these features are together fed into a cross-modality self-attention layer for joint representation learning. Following~\cite{sun2019videobert}, VisualBERT~\cite{li2019visualbert} extends such a single-stream framework for various image-text tasks and adds a segment embedding to distinguish between textual and visual tokens. VL-BERT~\cite{su2019vl} suggests that unmatched image-caption pairs over the ITA pre-training may decrease the accuracy of downstream tasks. And the authors further introduce both text-only corpus and unfrozen detector strategies for pre-training enhancement. Instead, such a ``harmful'' pre-training strategy is refuted by UNITER~\cite{chen2020uniter}, and the authors deploy an optimal transport loss to explicitly build Word-Region Alignment (WRA) at the instance level. To the same end, Oscar~\cite{li2020oscar} uses shared linguistic semantic embeddings of a salient object class (called tag) as an anchor point to link both region and its paired words. Zhou et al. propose Unified VLP~\cite{zhou2020unified} to handle both generation and understanding tasks via a shared Transformer encoder-decoder with two customized attention masks. Without extra auxiliary training, Unified VLP only adopts MLM during pre-training and attains superior results on Visual Question Answering (VQA)~\cite{antol2015vqa} and Visual Captioning (VC)~\cite{vinyals2015show} tasks.

However, these methods rely heavily on the visual extractor or predefined visual vocabulary, leading to a bottleneck of the VLP expressive upper bound. To address this issue, VinVL~\cite{zhang2021vinvl} develops an improved object detector for VLP pre-training on multiple large-scale dataset combination. Instead of the object-centric RoI features, ViLT~\cite{kim2021vilt} initializes the interaction Transformer weights from a pre-trained ViT, and adopts whole word masking and image augmentation strategy for VLP pre-training. UniT~\cite{hu2021unit} follows the architecture of DETR and applies a wide range of task for unified Transformer pre-training via different task-specific output heads simultaneously. SimVLM~\cite{wang2021simvlm} adopts ~\cite{dai2021coatnet} to obtain image features and designs a Prefix Language Modeling as pre-training objective to generalize zero-shot image captioning.

Besides the conventional pre-training scheme with multi-task supervision, another recent line has been developed for contrastive learning. The most representative work is CLIP~\cite{radford2021learning}. Based on the 400M Internet image-text pairs datasets, both image and text encoder are jointly trained by a contrastive loss for ITA. Different from previous methods, CLIP enables the pre-trained model with a linear classifier to zero-shot transfer to the most visual downstream tasks efficiently by embedding the whole semantics of the objective dataset’s classes.  Based on extensive experiments on over 30 existing CV tasks (e.g., classification and action recognition), CLIP attains superior results to classical supervised methods, demonstrating that such task-agnostic pre-training is also generalized well in the CV field. ALIGN~\cite{jia2021scaling} further expands a noisy dataset of over one billion image alt-text pairs rather than the elaborate filtering or post-processing steps in CLIP~\cite{radford2021learning}.
\begin{figure}[htbp]
    \centering
    \includegraphics[width=3.5in]{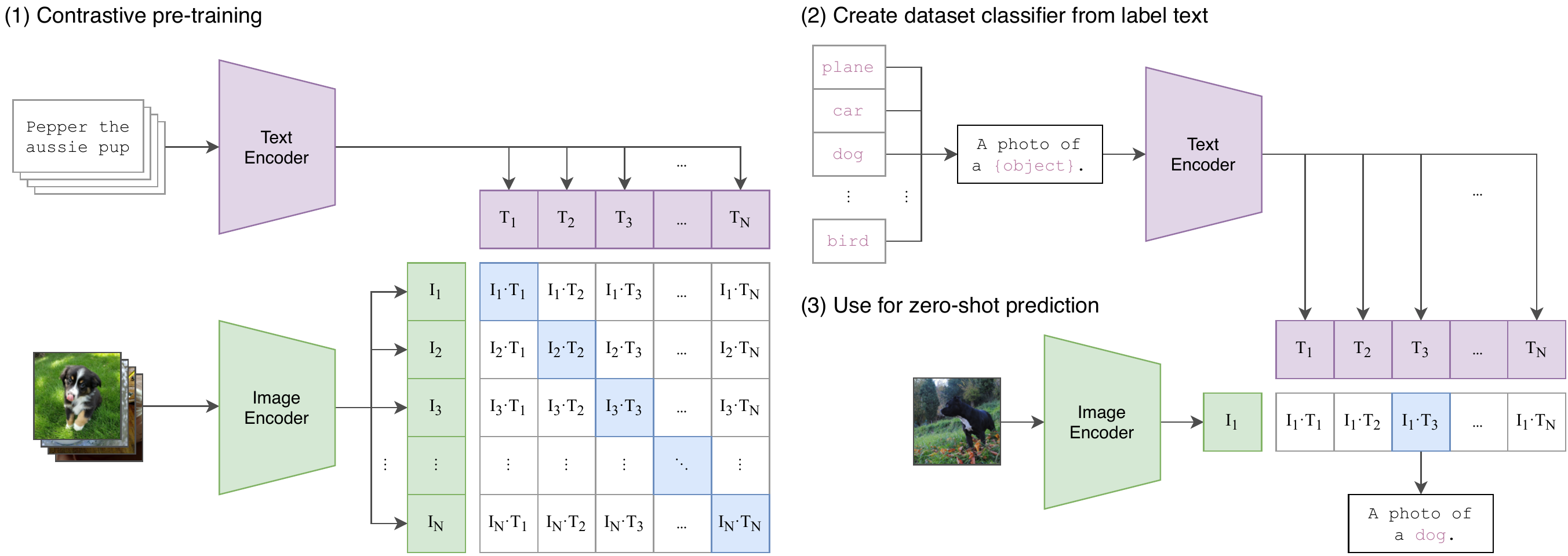}
    \caption{The overview of CILP (from~\cite{radford2021learning}).}
    \label{fig:12}
\end{figure}
Combining masked modeling and contrastive learning pre-training strategy, Data2Vec~\cite{baevski2022data2vec} proposes a self-distilled network treating the masked features as a type of data augmentation, whose structure is analogous to DINO (2021)~\cite{caron2021emerging}. By testing on different sensory benchmarks (voice, image, and language), it achieves competitive or better results compared with the existing self-supervised methods.

\subsubsection{Visual Grounding} Compared with VLP, visual grounding has more concrete target signal supervision whose objective is to locate the target objects according to their corresponding descriptions. In the image space, Modulated DETR (MDETR)~\cite{kamath2021mdetr} extends its previous work~\cite{carion2020end} to phrase grounding pre-training that locates and assigns the bounding box to each instance phrase in one description. Based on the proposed combined dataset from many existing ones, MDETR is first pre-trained on the 1.3M aligned text-image pairs for PG and then fine-tuned on other downstream tasks. During pre-training, the image-text pair features are separately processed by two specific extractors, and fed into a DETR-like Transformer for salient object localization. Besides the box loss, two auxiliary losses are adopted to enforce network to model an alignment between image feature and their corresponding phrase tokens. With the large-scale image-text pairs pre-training, MDETR can be easily generalized in few-shot learning, even on long-tail data. Different from MDETR~\cite{kamath2021mdetr} adding two auxiliary losses for box-phrase alignments, Referring Transformer~\cite{li2021referring} directly initializes object queries with phrase-specific embeddings for PG, which explicitly reserves an one-to-one phrase assignment for final bounding box prediction. VGTR~\cite{du2021visual} reformulates the REC as a task for single salient object localization from the language features. In detail, a text-guided attention mechanism encapsulates both self-attention block and text-image cross-attention one to update the image features simultaneously. The resulted image features, which serve as the key-value pairs, interact with language queries when regressing bounding box coordinates in the decoder. Following ViT~\cite{dosovitskiy2021an}, TransVG~\cite{deng2021transvg} keeps the class token to aggregate the image and language features simultaneously for the mentioned object localization in REC. Pseudo-Q~\cite{jiang2022pseudo} focuses on REC for the unsupervised learning, where a pseudo-query generation module based on a pre-trained detector and a series of attributes\&relationship generation algorithm is applied to generate a set of pseudo phrase descriptions, and a query prompt is introduced to match feature proposals and phrase queries for REC adaptation. 

In the 3D spaces, LanguageRefer~\cite{roh2022languagerefer} redefines the multi-stream data reasoning as a language modeling problem, whose core idea is to omit point cloud features and infuse the predicted class embeddings together with a caption into a language model to get a binary prediction for object selection. Following the conventional two-stream methods, TransRefer3D~\cite{he2021transrefer3d} further enhances the relationship of the object features by using a cross-attention between asymmetric object relation maps and linguistic features. Considering the specific view for varied descriptions, Huang et al. present a Multi-View Transformer (MVT 2022)~\cite{huang2022multi} for 3D visual grounding. Given a shared point cloud feature for each object, MVT first appends the converted bounding box coordinates to the shared objects in order to get specific view features. These multi-view features are then fed into a stack of the Transformer decoders for text data fusion. Finally, the multi-view features are merged by an order-independent aggregation function and converted to the grounding score. MVT achieves the SoTA performance on Nr3D and Sr3D datasets~\cite{achlioptas2020referit3d}. In the video space, a specific 3D data (with temporal dimension), Yang et al. propose TubeDETR~\cite{yang2022tubedetr} to address the problem of Spatio-Temporal Video Grounding (STVG). Concretely, a slow-fast encoder sparsely samples the frames and performs cross-modal self-attention between the sampled frames and the text features in the slow branch, and aggregates the updated sample features into the full-frame features from fast branch via a broadcast operation. A learnable query attached with different time encodings, called time-specific queries in the decoder is then predicted as either a time-aligned bounding box or ``no object''. It attains SoTA results on STVG leaderboards.

\section{Discussion and Conclusion}\label{sec:06}
This section briefly conducts a summary of the performance improvements provided in Sec.~\ref{sec:061}, some critical issues discussed in Sec.~\ref{sec:062}, future research directions suggested in Sec.~\ref{sec:063}, and final conclusion given in Sec.~\ref{sec:064}. 

\subsection{Summary of Recent Improvements}\label{sec:061}
We briefly summarize the major performance improvements for three fundamental CV tasks as follows.

(1) For classification, a deep hierarchical Transformer backbone is valid for decreasing the computational complexity~\cite{wang2021pyramid} and avoiding the feature over-smooth~\cite{zhou2021deepvit,touvron2021going,gong2021diverse,zhou2021refiner} in the deep layer. Meanwhile, the early-stage convolution~\cite{dai2021coatnet} is enough to capture the low-level features, which can significantly enhance the robustness and reduce the computational complexity in the shallow layer. Moreover, both the convolutional projection~\cite{yuan2021incorporating,li2021localvit} and the local attention mechanism~\cite{liu2021swin,yuan2021volo} can improve the locality of the visual Transformers. The former~\cite{chu2021conditional,zhang2021rest} may also be a new approach to replace the positional encoding.
  
(2) For detection, the Transformer necks benefit from the encoder-decoder structure with less computation than the encoder-only Transformer detector~\cite{fang2021you}. Thus, the decoder is necessary but it requires more spatial prior~\cite{zhu2021deformable,gao2021fast,meng2021conditional, wang2021anchor, liu2022dab, yao2021efficient, dai2021dynamic} owing to its slow convergence~\cite{sun2021rethinking}. Furthermore, sparse attention~\cite{zhu2021deformable} and scoring network~\cite{wang2021pnp,roh2021sparse} for fore-grounding sampling are conducive to reducing the computational costs and accelerating the convergence of visual Transformers.

(3) For segmentation, the encoder-decoder Transformer models may unify three segmentation sub-tasks into a mask prediction problem via a set of learnable mask embeddings~\cite{wang2021max,strudel2021segmenter,wang2021not}. This box-free approach has achieved the latest SoTA performance on multiple benchmarks~\cite{wang2021not}. Moreover, the specific hybrid task is cascaded with the model~\cite{fang2021queryinst} of the box-based visual Transformers, which have demonstrated a higher performance for instance segmentation.

(4) For 3D visual recognition, the local hierarchical Transformer with a scoring network could efficiently extract features from the point clouds. Instead of the elaborate local design, the global modeling capability enables the Transformer to easily aggregate surface points. In addition, the visual Transformers can handle multi-sensory data in 3D visual recognition, such as multi-view and multi-dimension data.

(5) The mainstream approaches of visual-linguistic pre-training has gradually abandoned the pre-trained detector~\cite{kim2021vilt} and focused on the alignments~\cite{radford2021learning} or similarities~\cite{baevski2022data2vec} among  different data streams in the latent space based on the large-scale noised datasets~\cite{jia2021scaling}. Another concern is to adapt the downstream visual tasks to the pre-training scheme to perform zero-short transferring~\cite{radford2021learning}.

(6) The recent prevailing architecture for multi-sensory data stream fusion is the single-stream method, which spatially concatenates different data streams and performs interaction simultaneously. Based on the single-stream model, numerous recent works devote to finding a latent space to make different data streams semantically consistent.

\subsection{Discussion on Visual Transformers}\label{sec:062}
Despite that the visual Transformer models are evolved significantly, the ``essential'' understanding remains insufficient. Therefore, we will focus on reviewing some key issues for a deep and comprehensive understanding.

\subsubsection{How Transformers Bridge the Gap Between Language and Vision}\label{sec:0621}
Transformers are initially designed for machine translation tasks~\cite{2017Attention}, where each word of a sentence is taken as a basic unit representing the high-level semantics. These words can be embedded into a series of vector representations in the N-dimensional feature space. For visual tasks, each single pixel of an image is unable to carry semantic information, which is not full compliance with the feature embedding as done for the traditional NLP tasks. Therefore, the key for transferring such feature embeddings (i.e., word embedding) to image features and applying Transformer to various vision tasks is to \textit{build an image-to-vector transformation and maintain the image's characteristics effectively}. For example, ViT~\cite{dosovitskiy2021an} transforms an image into patch embeddings with multiple low-level information under strong slackness conditions. And its votarist~\cite{xiao2021early, dai2021coatnet} leverages convolution to extract the low-level features and reduce the redundancy from patches.

\subsubsection{The Relationship between Transformers, Self-Attention and CNNs}\label{sec:0622}
From the perspective of CNNs, its inductive bias is mainly shown as locality, translation invariance, weight sharing, and sparse connection. Such a simple convolutional kernel can perform template matching efficiently in lower-level semantic processing but its upper-bound tend to be lower than Transformers due to the excessive bias. 

 \begin{figure}[htbp]
    \centering
    \includegraphics[width=3in]{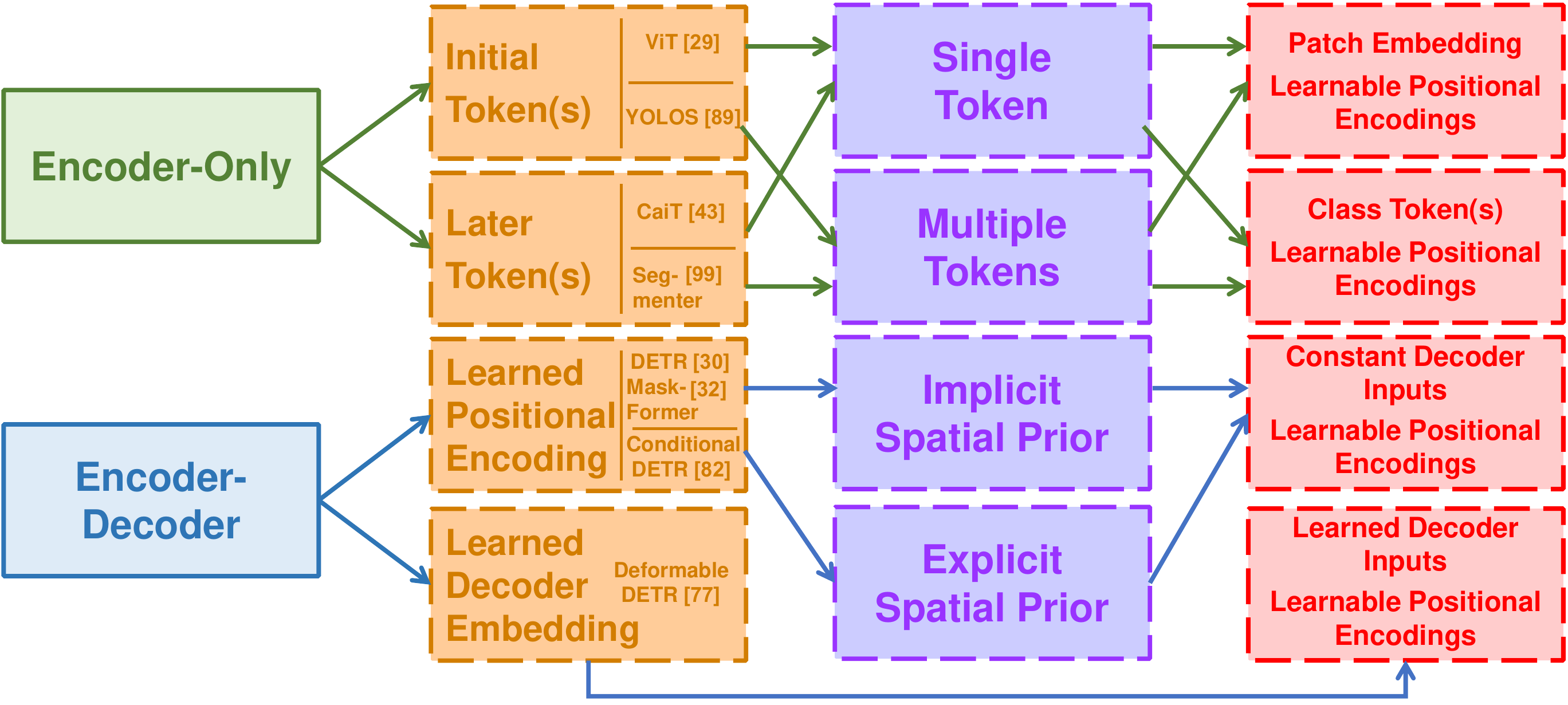}
    \caption{Taxonomy of the learnable embedding.}
    \label{fig:13}
\end{figure}

From the perspective of Transformers, as detailed in Sec.~\ref{sec:032} and Sec.~\ref{sec:034}, \textit{attention layer can theoretically express any convolution when a sufficient number of heads are adopted}~\cite{cordonnier2020on}. Such fully-attentional operation can combine both local-level and global-level attentions, and generate attention weights dynamically according to the feature relationships. Dong et al. demonstrate that the self-attention layer manifests strong inductive bias towards ``token uniformity'' when it is trained on deep layers without short connection or FFNs~\cite{dong2021attention}. Yu et al. also argue that such  an elaborate attention mechanism can be replaced by a pooling operation readily~\cite{yu2022metaformer}. Therefore, it is concluded that \textit{Transformer must consist of two key components: a global token mixer (e.g., self-attention layer) aggregates the relationship of tokens, and a position-wise FFN extracts the features from the inputs}. 

By comparison, the visual Transformer has a powerful global modelling capability, making it efficiently attend to high-level semantic features. CNNs can effectively process the low-level features~\cite{xiao2021early,dai2021coatnet}, enhance the locality of the visual Transformers~\cite{meng2021conditional,d2021convit}, and append the positional features via padding operations~\cite{islam2020much,chu2021conditional,zhang2021rest}.

\subsubsection{Double Edges of Visual Transformers}\label{sec:0624}
We conclude three double-edged properties of visual Transformers as follows. \textit{Global} property enables Transformer to acquire capacious receptive fields and interact easily between various high-level semantic features, while it becomes inefficiency and debility during low-level processing because of quadratic computing and noised low-level features. \textit{Slack bias} offers visual Transformer a higher upper bound than CNNs based on sufficient training data without sophistic assumptions but performs inferiority and slow convergency in small datasets~\cite{park2022vision}. \textit{Low-pass} is also a significant property of visual Transformer showing excellent robustness, whereas it is insensitive to low-level features (e.g., complicated textures and edges) compared with CNN. Accordingly, it is concluded that Transformers at a high-level stage play a vital role in various vision tasks.

\subsubsection{Learnable Embeddings for Different Visual Tasks}\label{sec:0623}
Various learnable embeddings are designed to perform different visual tasks, such as class token, object query, and mask embedding. These learnable tokens are mainly adopted into two different Transformer patterns, i.e., encoder-only and encoder-decoder ones, as illustrated in Fig.~\ref{fig:13}. On the quantity level, the number of learned tokens depends on the target prediction. For example, the visual Transformers~\cite{dosovitskiy2021an,touvron2021training} in the classification task adopt only one class token, and the DETR's votarist in detection~\cite{carion2020end,meng2021conditional} and segmentation~\cite{wang2021not} tasks employ multiple learned queries. On the position level, encoder-only Transformers capitalize on the initial token(s)~\cite{dosovitskiy2021an,fang2021you} and later token(s)~\cite{touvron2021going,strudel2021segmenter}, while the learned positional encoding~\cite{carion2020end,wang2021not,meng2021conditional} and the learned decoder input embedding~\cite{zhu2021deformable} are applied to the encoder-decoder structure. Different from the vanilla ViT with initial class token, CaiT~\cite{touvron2021going} observes that the later class token can reduce FLOPs of Transformer and improve model performance slightly. Segmenter~\cite{strudel2021segmenter} also shows such strategy efficiency for the segmentation tasks. From the viewpoint of the encoder-decoder Transformer, the decoder input token is considered as a special case of the encoder-only Transformer with later token. It standardizes visual Transformers in the fields of detection~\cite{carion2020end} and segmentation~\cite{wang2021not} by using a small set of object queries (mask embeddings). By combing both later tokens and object queries (mask embeddings), the structure like Deformable DETR~\cite{zhu2021deformable}, which takes object queries and the learnable decoder embeddings (equivalent to the later tokens) as the inputs, may unify the learnable embeddings for different tasks into the encoder-decoder Transformer.

\subsection{Future Research Directions}\label{sec:063}
Visual Transformers have achieved significant progresses and obtained promising results. However, some key technologies are still insufficient to cope with complicated challenges in the CV fields. Based on the above analysis, we point out some promising research directions for future investigation.
 
\subsubsection{Set Prediction}\label{sec:0632}
Touvron et al. found that multiple class tokens would converge consistently due to the same gradient from the loss function~\cite{touvron2021training}, whereas it does not emerge in dense prediction tasks~\cite{carion2020end,wang2021not}. We conclude that their marked difference lies in the label assignment and the number of targets. Thus, it is natural to consider a set prediction design for the classification tasks, e.g., multiple class tokens are aligned to mix-patches via set prediction, like the data augmentation training strategy in LV-ViT~\cite{jiang2021token}. Furthermore, the label assignment in the set prediction strategy leads to training instability during the early process, which degrades the accuracy of the final results. Redesigning the label assignments and set prediction losses may be helpful for the detection frameworks.

\subsubsection{Self-Supervised Learning}\label{sec:0633}
Self-supervised pre-training of Transformers has standardized the NLP field and obtained tremendous successes in various applications~\cite{devlin2018bert,radford2018improving}. Because of the popularity of self-supervision paradigms in the CV field, the convolutional Siamese networks employ contrastive learning to perform self-supervised pre-training, which differs from the masked auto-encoders used in the NLP field. Recently, some studies have tried to design  self-supervised visual Transformers to bridge the discrepancy of pre-training methodology between vision and language. Most of them inherit the masked auto-encoders in the NLP field or contrastive learning schemes in the CV field. There is no specific supervised method for the visual Transformers, but it has revolutionized the NLP tasks such as GPT-3. As described in Sec.~\ref{sec:0623}, the encoder-decoder structure may unify the visual tasks by learning the decoder embedding and the positional encoding jointly. Thus it is worth of further investigating the encoder-decoder Transformers for self-supervised learning.

\subsection{Conclusion}\label{sec:064}
Since ViT demonstrated its effectiveness for the CV tasks, the visual Transformers have received considerable attentions and undermined the dominant of CNNs in the CV field. In this paper, we have comprehensively reviewed more than one hundred of visual Transformer models which have been successively applied to various vision tasks (i.e., classification, detection, and segmentation) and data streams (e.g., images, point clouds, image-text pairs, and other multiple data streams). For each vision task and data stream, a specific taxonomy is proposed to organize the recently-developed visual Transformers and their performances are further evaluated over various prevailing benchmarks. From our integrative analysis and systematic comparison of all these existing methods, a summary of remarkable performance improvements is provided in this paper, four essential issues for the visual Transformers are also discussed, and several potential research directions are further suggested for future investment. We do expect that this review paper can help readers have better understandings of various visual Transformers before they decide to perform deep explorations.

\appendices
\section{Overview of Development Trend on Visual Transformers}\label{sup:development}
Transformer backbones sprang up within the last year. When our systematics matches the timeline of these models, we can clearly trace the development tendency of Transformer for image classification (the Fig. 1 in main text). As a type of self-attention mechanism, visual Transformers are mainly redesigned according to either the vanilla structure in NLP (ViT~\cite{dosovitskiy2021an} and iGPT~\cite{chen2020generative}) or attention-based model in CV (VTs~\cite{wu2020visual} and BoTNet~\cite{srinivas2021bottleneck}). 

Then, many approaches start to extend the hierarchical or deep structure of CNN to visual Transformer. T2T-ViT~\cite{yuan2021tokens}, PVT~\cite{wang2021pyramid}, CvT~\cite{wu2021cvt} and PiT~\cite{heo2021rethinking} share a motivation that transferring the hierarchical structure into Transformer but they perform downsampling differently. CaiT~\cite{touvron2021going}, Diverse Patch~\cite{gong2021diverse}, DeepViT~\cite{zhou2021deepvit}, and Refiner~\cite{zhou2021refiner} focus on the problem in deep Transformer. Moreover, some approaches move on to the internal components to further enhance the image processing capability in previous Transformers, i.e., positional encoding~\cite{wu2021rethinking,islam2021position,chu2021conditional}, MHSA~\cite{cordonnier2020on}, and MLP~\cite{dong2021attention}.

The next wave of Transformers is locality paradigm. Most of them introduce locality into Transformer via introducing local attention mechanism~\cite{han2021transformer,liu2021swin,yuan2021volo,chu2021twins} or convolution~\cite{d2021convit, yuan2021incorporating,li2021localvit}. Nowadays, the most recent supervised Transformers are exploring both the structural combination~\cite{xiao2021early,dai2021coatnet} and scaling laws~\cite{zhai2021scaling,riquelme2021scaling}. In addition to supervised Transformers, self-supervised learning accounts for a substantial part of vision Transformers~\cite{caron2021emerging,chen2021empirical,xie2021self,chen2020generative,li2021mst,bao2021beit}. However, it is unclear what tasks and structures are more beneficial to self-supervised Transformer in CV.

\begin{figure}[htbp]
    \centering
    \includegraphics[width=3.2in]{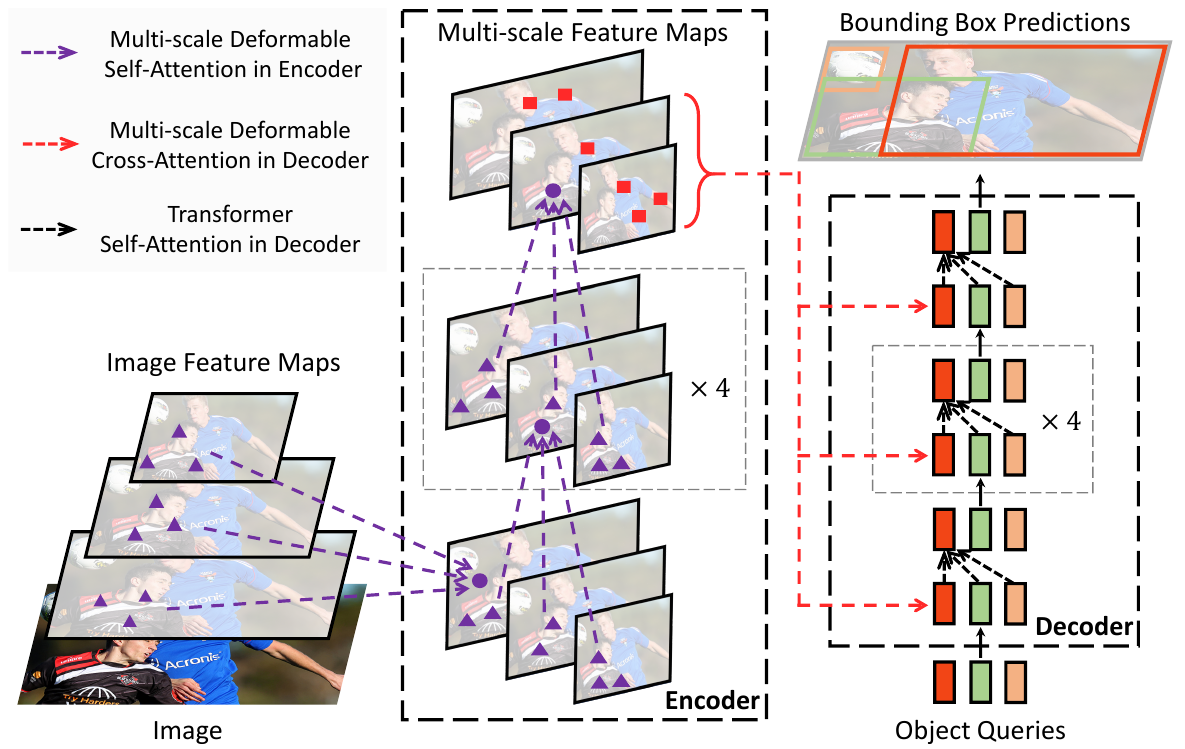}\\
    \caption{Illustration of Deformable DETR. A fixed number of key samples in each scale feature interacting with all queries.(from~\cite{zhu2021deformable}.)}
    \label{fig:s1}
\end{figure}

\section{More detail formula of DETR}
\label{sup:detr}
The bipartite matching loss $\mathcal{L}_{\text{match}}$ is applied between prediction $\hat{y}_{\sigma(i)}$ and ground-truth objects $y_i$ to identify one-to-one label assignment $\hat{\sigma}$ as
\begin{equation}\label{eq:30}
    \begin{array}{l}
        \hat{\sigma }=\mathop{\text{argmin}}\limits_{\sigma \in \mathfrak{S}_N }\sum\limits_{i}^{N}\mathcal{L}_{\text{match}}(y_i,\hat{y}_{\sigma(i)} ) , \vspace{1ex}\\
        \mathcal{L}_{\text{match}}(y_i,\hat{y}_{\sigma(i)}) = -\mathds{1}_{ \{c_i \neq \varnothing\}}\hat{p}_{\sigma(i)}(c_i) \vspace{0.1ex}\\
        \qquad \qquad \qquad \quad \ \ \,   + \mathds{1}_{ \{c_i \neq \varnothing\}}\mathcal{L}_{\text{box}}(b_i,\hat{b}_{\sigma(i)}).\!\!\!
    \end{array}
\end{equation}

In back propagation, the Hungarian loss $\mathcal{L}_{\text{H}}$ includes a negative log-likelihood loss for all label predictions ($i=1\cdots N$) and a box loss for all matched pairs ($c_i \neq \varnothing$) as
\vspace{-0.025cm}
\begin{flalign}\label{eq:31}
\noindent
\hspace{-0.4cm}
    \mathcal{L}_{\text{H}}(\!y_i,\hat{y}_{\sigma(i)}\!) \!=\! \textstyle \sum\limits_{i=1}^N[-\text{log}\hat{p}_{\hat{\sigma}(i)}(\!c_i\!)
    \!+ \!\mathds{1}_{ \!\{c_i \neq \varnothing\}\!}\mathcal{L}_{\text{box}}(\!b_i,\hat{b}_{\sigma(i)}\!)].\!\!\!\!
\end{flalign}

\section{More detailed formula of Deformable DETR}
\label{sup:deformable-detr}
Given $L$ feature maps $X^l \in \mathbb{R}^{H_l\times W_l\times C}$ and a query sequence $\mathbf{z}\in \mathbb{R}^{N_q \times C}$, MSDA samples offsets $\Delta \mathbf{p}\in \mathbb{R}^2$ of each query for $N_k$ sample keys at each layer's head via two linear layers. While it is sampling the features of key points $V_i\in \mathbb{R}^{L\times N_k\times C_v}$, a linear projection layer is applied to the query to generate an attention map $A_i \in \mathbb{R}^{N_q\times L\times N_k}$ for key samples, where $N_q$ and $C$ are query's length and dimension, respectively (see Fig.~\ref{fig:s1}). The process is formulated as
\begin{equation}
\begin{array}{l}\label{eq:32}
    A_{qik}^l\!= \mathbf{z}_{q}W_{ilk}^{A},  \   V_{ik}^l\!=\!X^l(\phi_l(\hat{\mathbf{p}}_q)+\Delta\mathbf{p}_{ilqk})W^{V}_{i},
    \vspace{1ex}\\
    \text{MSDAttn}(A^l_{qik},V^l_{ik})=\sum\limits^{h}_{i=1}(\sum\limits^{L}_{l=1}\sum\limits^{N_k}_{k=1}A_{qik}^lV_{ik}^l)W^O_i,
\end{array}
\end{equation}
where $m$ denotes the attention head, $W_{il}^A \!\!\in\!\! \mathbb{R}^{C \times N_k}\!\!$, $W_{i}^V \!\!\in\!\! \mathbb{R}^{C \times C_v}\!\!$ and $W_i^{O} \!\!\in\!\! \mathbb{R}^{C_v \times C}\!\!$ are linear matrices. $\hat{\mathbf{p}}_q \!\!\in\!\! [0,1]^2\!\!$ is normalized coordinates of each query.

\begin{table*}[htbp]
  \centering
  \caption{Dense prediction results of COCO 2017 val. set based on RetinaNet~\cite{lin2017focal} and Mask R-CNN~\cite{he2017mask}, when trained with 3× schedule and multi-scale inputs (MS). The numbers before and after ``/'' correspond to the parameter of RetinaNet and Mask R-CNN, respectively. (Most of data from~\cite{yang2021focal}.)}
  \setlength \tabcolsep{5.5pt}
    \begin{spacing}{0.7}
    \resizebox{\linewidth}{!}{%
    \begin{tabular}{l|cc|cccccc|cccccc}
    \toprule
    \multicolumn{1}{l}{\multirow{2}[2]{*}{\textbf{Backbone}}} & \multicolumn{1}{c}{\multirow{2}[2]{*}{\textbf{\makecell[c]{\#Params \\ (M)}}}} & \multicolumn{1}{c}{\multirow{2}[2]{*}{\textbf{\makecell[c]{FLOPs \\ (G)}}}} & \multicolumn{6}{c|}{\textbf{RetinaNet 3$\times$ schedule + MS}} & \multicolumn{6}{c}{\textbf{Mask R-CNN 3$\times$ schedule + MS}} \\
    \cmidrule{4-15}
          &      &      & $\textbf{AP}^{box}$ & $\textbf{AP}^{box}_{50}$ & $\textbf{AP}^{box}_{75}$ & $\textbf{AP}^{box}_S$ & $\textbf{AP}^{box}_M$ & $\textbf{AP}^{box}_L$ & $\textbf{AP}^{box}$ & $\textbf{AP}^{box}_{50}$ & $\textbf{AP}^{box}_{75}$ & $\textbf{AP}^{seg}$ & $\textbf{AP}^{seg}_{50}$ & $\textbf{AP}^{seg}_{75}$ \\
    \midrule
    ResNet50~\cite{he2016deep} & 38 / 44 & 239 / 260 & 39.0  & 58.4  & 41.8  & 22.4  & 42.8  & 51.6  & 41.0  & 61.7  & 44.9  & 37.1  & 58.4  & 40.1  \\
    PVTv1-Small~\cite{wang2021pyramid} & 34 / 44 & 226 / 245 & 42.2  & 62.7  & 45.0  & 26.2  & 45.2  & 57.2  & 43.0  & 65.3  & 46.9  & 39.9  & 62.5  & 42.8  \\
    ViL-Small~\cite{zhang2021multi} & 36 / 45 & 252 / 174 & 42.9  & 63.8  & 45.6  & 27.8  & 46.4  & 56.3  & 43.4  & 64.9  & 47.0  & 39.6  & 62.1  & 42.4  \\
    Swin-Tiny~\cite{liu2021swin} & 39 / 48 & 245 / 264 & 45.0  & 65.9  & 48.4  & 29.7  & 48.9  & 58.1  & 46.0  & 68.1  & 50.3  & 41.6  & 65.1  & 44.9  \\
    PVTv2-B2-Li~\cite{wang2021pvtv2} & 32 / 42 & - / - & -     & -     & -     & -     & -     & -     & 46.8  & 68.7  & 51.4  & 42.3  & 65.7  & 45.4  \\
    Focal-Tiny~\cite{yang2021focal} & 39 / 49 & 265 / 291 & 45.5  & 66.3  & 48.8  & 31.2  & 49.2  & 58.7  & 47.2  & 69.4  & 51.9  & 42.7  & 66.5  & 45.9  \\
    PVTv2-B2~\cite{wang2021pvtv2} & 35 / 45 & - / - & -     & -     & -     & -     & -     & -     & 47.8  & 69.7  & 52.6  & 43.1  & 66.8  & 46.7  \\
    
    \specialrule{0.5pt}{0.8pt}{1.7pt}
    ResNet101~\cite{he2016deep} & 57 / 63 & 315 / 336 & 40.9  & 60.1  & 44.0  & 23.7  & 45.0  & 53.8  & 42.8  & 63.2  & 47.1  & 38.5  & 60.1  & 41.3  \\
    ResNeXt101-32x4d~\cite{xie2017aggregated} & 56 / 63 & 319 / 340 & 41.4  & 61.0  & 44.3  & 23.9  & 45.5  & 53.7  & 44.0  & 64.4  & 48.0  & 39.2  & 61.4  & 41.9  \\
    PVTv1-Medium~\cite{wang2021pyramid} & 54 / 64 & 283 / 302 & 43.2  & 63.8  & 46.1  & 27.3  & 46.3  & 58.9  & 44.2  & 66.0  & 48.2  & 40.5  & 63.1  & 43.5  \\
    ViL-Medium~\cite{zhang2021multi} & 51 / 60 & 339 / 261 & 43.7  & 64.6  & 46.4  & 27.9  & 47.1  & 56.9  & 44.6  & 66.3  & 48.5  & 40.7  & 63.8  & 43.7  \\
    Swin-Small~\cite{liu2021swin} & 60 / 69 & 335 / 354 & 46.4  & 67.0  & 50.1  & 31.0  & 50.1  & 60.3  & 48.5  & 70.2  & 53.5  & 43.3  & 67.3  & 46.6  \\
    Focal-Small~\cite{yang2021focal} & 62 / 71 & 367 / 401 & 47.3  & 67.8  & 51.0  & 31.6  & 50.9  & 61.1  & 48.8  & 70.5  & 53.6  & 43.8  & 67.7  & 47.2  \\
     \specialrule{0.5pt}{0.8pt}{1.7pt}
    ResNeXt101-64x4d~\cite{xie2017aggregated} & 96 / 102 & 473 / 493 & 41.8  & 61.5  & 44.4  & 25.2  & 45.4  & 54.6  & 44.4  & 64.9  & 48.8  & 39.7  & 61.9  & 42.6  \\
    PVTv1-Large~\cite{wang2021pyramid} & 71 / 81 & 345 / 364 & 43.4  & 63.6  & 46.1  & 26.1  & 46.0  & 59.5  & 44.5  & 66.0  & 48.3  & 40.7  & 63.4  & 43.7  \\
    ViL-Base~\cite{zhang2021multi} & 67 / 76 & 443 / 365 & 44.7  & 65.5  & 47.6  & 29.9  & 48.0  & 58.1  & 45.7  & 67.2  & 49.9  & 41.3  & 64.4  & 44.5  \\
    Swin-Base~\cite{liu2021swin} & 98 / 107 & 477 / 496 & 45.8  & 66.4  & 49.1  & 29.9  & 49.4  & 60.3  & 48.5  & 69.8  & 53.2  & 43.4  & 66.8  & 46.9  \\
    Focal-Base~\cite{yang2021focal} & 101 / 110 & 514 / 533 & 46.9  & 67.8  & 50.3  & 31.9  & 50.3  & 61.5  & 49.0  & 70.1  & 53.6  & 43.7  & 67.6  & 47.0  \\
    \bottomrule
    \end{tabular}}
   \end{spacing}
   \label{tab:s1}
\end{table*}%

\begin{table*}[htbp]
  \centering
  \caption{Details of visual-linguistic pre-training methods, where $\bullet$ and $\bullet \bullet$ denote single- and dual-stream architecture, respectively, and the zero-shot denotes the method can be zero-shot transfered into down stream tasks. In the pre-training tasks, MRM is Masked Region Modeling, OD is Object Detection,  SMLM and BMLM denote both sequentially and bidirectionally masked language modeling, and MVM is Mased Visual-Token Modeling.}
    \setlength \tabcolsep{4pt}
    \begin{spacing}{0.6}
    \resizebox{\linewidth}{!}{
    \begin{tabular}{lcccccccc}
    \toprule
    \multirow{2}[1]{*}{\textbf{Methods}} & \multirow{2}[1]{*}{\textbf{Arch.}} & \multicolumn{1}{c}{\multirow{2}[1]{*}{\textbf{Visual Token}}} & \multicolumn{3}{c}{\textbf{Pre-training}}  & \multirow{2}[1]{*}{\textbf{\makecell[c]{Zero \\ Shot}}} & \multirow{2}[1]{*}{\textbf{Publiction}}  \\
\cmidrule{4-6}          &       &       & \multicolumn{1}{c}{Main Dataset(s)} & \multicolumn{1}{c}{Data Size}  &  \multicolumn{1}{c}{Tasks}     &       & & \\
    \specialrule{0.5pt}{0.8pt}{1.7pt}
    \multicolumn{9}{l}{\textit{Region-Besed Methods}} \\
    \specialrule{0.5pt}{0.8pt}{1.7pt}
    VideoBERT~\cite{sun2019videobert} &   $\bullet$    & S3D~\cite{xie2018rethinking}/w k-means & YouTube Cooking~\cite{sun2019videobert} & 312K  & \makecell[c]{ITA, \\ MLM, MVM} & $\checkmark$     & ICCV 2019 & \\
    \specialrule{0.5pt}{0.8pt}{1.7pt}
    ViLBERT~\cite{lu2019vilbert} &   \makecell[c]{$\bullet$ \\ $\bullet$}    & RoI~\cite{anderson2018bottom}  & CC3M~\cite{sharma2018cc3m}  & 3.1M  & \makecell[c]{ITA, \\ MLM, MRC-KL } & -     & NeurIPS 2019 & \\
    \specialrule{0.5pt}{0.8pt}{1.7pt}
    LXMERT~\cite{tan2019lxmert} &    \makecell[c]{$\bullet$ \\ $\bullet$}   & RoI~\cite{anderson2018bottom}  & \makecell[c]{ VG-QA \\ VQAv2~\cite{goyal2017vqav2}, VG~\cite{krishna2017vg}, \\ COCO~\cite{lin2014coco}, GQA~\cite{hudson2019gqa}} & 9.2M & \makecell[c]{ITA, MLM, \\  MRM, MRC} & -     & IJCNLP 2019&\\
    \specialrule{0.5pt}{0.8pt}{1.7pt}
    VisualBERT~\cite{li2019visualbert} &    $\bullet$   & Faster RCNN~\cite{ren2017faster} & COCO~\cite{lin2014coco}  & 0.9M  & ITA, MLM & -     & Arxiv 2019& \\
    \specialrule{0.5pt}{0.8pt}{1.7pt}
    VL-BERT~\cite{su2019vl} &   $\bullet$    & RoI~\cite{anderson2018bottom}  & \makecell[c]{CC3M~\cite{sharma2018cc3m},  BooksCorpus \\ \&English Wikipedia} & 11M   & MLM, MRC & -     & ICLR 2020& \\
    \specialrule{0.5pt}{0.8pt}{1.7pt}
    UNITER~\cite{chen2020uniter} &   $\bullet$    & RoI~\cite{anderson2018bottom}  & \makecell[c]{ CC3M~\cite{sharma2018cc3m}, SBU~\cite{ordonez2011sbu},\\ COCO~\cite{lin2014coco}, VG~\cite{krishna2017vg}} & 9.5M & \makecell[c]{ITA, WRA, \\ MLM/MRM } & -     & ECCV 2020& \\
    \specialrule{0.5pt}{0.8pt}{1.7pt}
    Oscar~\cite{li2020oscar} &   $\bullet$    & \makecell[c]{RoI~\cite{anderson2018bottom} \\ +Tags} & \makecell[c]{COCO~\cite{lin2014coco}, GQA~\cite{hudson2019gqa}, \\ CC3M~\cite{sharma2018cc3m}, SBU~\cite{ordonez2011sbu}, \\ VG~\cite{krishna2017vg}, Fliker30K~\cite{young2014flicker30k}}  & 11.4M & ITA, MLM  & -     & ECCV 2020& \\
    \specialrule{0.5pt}{0.8pt}{1.7pt}
    Unified-VLP~\cite{zhou2020unified} &    $\bullet$   & RoI~\cite{anderson2018bottom}  & CC3M~\cite{sharma2018cc3m}  & 3.1M  & SMLM, BMLM & -     & AAAI 2020& \\
    \specialrule{0.5pt}{0.8pt}{1.7pt}
    VinVL(Oscar+)~\cite{zhang2021vinvl} &   $\bullet$    & \makecell[c]{RoI~\cite{anderson2018bottom}/w  NMS \\ +Tags} & \makecell[c]{SBU~\cite{ordonez2011sbu}, VG-Qas~\cite{krishna2017vg}, \\  COCO~\cite{lin2014coco}, CC3M~\cite{sharma2018cc3m}, \\ GQA~\cite{hudson2019gqa}, Fliker30K~\cite{young2014flicker30k}, \\ VQA~\cite{antol2015vqa}, OpenImages~\cite{kuznetsova2020openimages}} & 8.9M  & MLM, ITA & -     & CVPR 2021&\\
    \specialrule{0.5pt}{0.8pt}{1.7pt}
    \multicolumn{8}{l}{\textit{Feature-Based Methods}} \\
    \specialrule{0.5pt}{0.8pt}{1.7pt}
    ViLT~\cite{kim2021vilt}  &    $\bullet$   & Patches from ViT~\cite{dosovitskiy2021an} & \makecell[c]{SBU~\cite{ordonez2011sbu}, CC3M~\cite{sharma2018cc3m}, \\ COCO~\cite{lin2014coco}, VG~\cite{krishna2017vg}} & 10M   & ITM, MLM & $\checkmark$     & ICML 2021& \\
    \specialrule{0.5pt}{0.8pt}{1.7pt}
    UniT~\cite{hu2021unit}  &   $\bullet$    & DETR-ResNet50~\cite{carion2020end} & \makecell[c]{COCO~\cite{lin2014coco}, VG~\cite{krishna2017vg}, \\ VQAv2~\cite{goyal2017vqav2}, SNLI-VE \\ Four LM Datasets} & -     & \makecell[c]{OD, 4LM, \\ 2ILM} & $\checkmark$     & ICCV 2021&\\
    \specialrule{0.5pt}{0.8pt}{1.7pt}
    CLIP~\cite{radford2021learning}  &    \makecell[c]{$\bullet$ \\ $\bullet$}   & ViT~\cite{dosovitskiy2021an}& Internet Pairs~\cite{radford2021learning} & 400M  & Contrasive & $\checkmark$    & ICML 2021& \\
    \specialrule{0.5pt}{0.8pt}{1.7pt}
    DALL-E~\cite{ramesh2021zero} &   $\bullet$    & dVAE  & \makecell[c]{Extension~\cite{ramesh2021zero} \\ from COCO} & 250M  & Contrasive & $\checkmark$     & ICML 2021 &\\
    \specialrule{0.5pt}{0.8pt}{1.7pt}
    ALIGN~\cite{jia2021scaling} &   $\bullet$   & EfficientNet~\cite{tan2019efficientnet} & \makecell[c]{Noise English \\ al-text data~\cite{jia2021scaling}}   & 1.8B  & Contrasive & $\checkmark$     & ICML 2021&\\
    \specialrule{0.5pt}{0.8pt}{1.7pt}
    SimVLM~\cite{wang2021simvlm} &   $\bullet$    & CoAtNet~\cite{dai2021coatnet} & \makecell[c]{Noise English \\ al-text data~\cite{jia2021scaling}}  & 1.8B  & PLM & $\checkmark$     & ICLR 2022&\\
    \specialrule{0.5pt}{0.8pt}{1.7pt}
    Data2Vec~\cite{baevski2022data2vec} &   $\bullet$    & ViT~\cite{dosovitskiy2021an}   &   \makecell[c]{ImageNet-1k \\ LS-960 \\Books Corpus \&\\ English Wikipedia data}    &    \makecell[c]{1k\\960h\\1M}   & Self-Distillation & $\checkmark$    & Arxiv 2022& \\
    \bottomrule
    \end{tabular}}
    \end{spacing}
  \label{tab:s2}
\end{table*}

\section{More Comparison of Visual Transformers for Dense Prediction and Visual-Linguistic Pre-Training}
\label{sup:comparison}
Based on RetinaNet~\cite{lin2017focal} and Mask R-CNN~\cite{he2017mask}, Tab.~\ref{tab:s1} compares several visual Transformer backbones on the COCO datasets for the dense prediction tasks. And Tab.~\ref{tab:s2} collects the visual Transformers as mentioned in visual-linguistic pre-training (Sec. VII-B1). Concretely, we summarize the main datasets, data size, and objective task for each visual Transformer during the pre-training. The architecture (single-/dual-stream) and the type of visual token inputs are also organized for each approach.

{\small
\bibliographystyle{unsrt2authabbrvpp}
\bibliography{ieeetran}
}
\balance
\end{document}